%% file: main.tex
\documentclass[conference]{IEEEtran}
\usepackage{times}

\usepackage[numbers,sort]{natbib}
\usepackage{multicol}
\usepackage{adjustbox}
\usepackage{hyperref}
\usepackage{duckuments} %

\usepackage[table]{xcolor}
\usepackage{booktabs,threeparttable,adjustbox,multirow,makecell,siunitx}
\usepackage[font=small]{caption}

\usepackage{stfloats} %
\usepackage{subcaption}
\usepackage{graphicx}
\usepackage{lipsum}  
\usepackage{makecell}
\usepackage{amsmath}
\usepackage{multirow}
\usepackage{booktabs}
\usepackage{array}
\usepackage{environ}
\usepackage{etoolbox}

\usepackage{wrapfig}
\usepackage{amsfonts}
\let\labelindent\relax %
\usepackage[shortlabels]{enumitem} %
\setlist[enumerate]{topsep=2pt, itemsep=0pt, parsep=0pt}
\usepackage{twemojis}
\usepackage{bbm}
\usepackage{booktabs}
\usepackage{graphicx}

\usepackage{titletoc}

\usepackage{mathtools, algorithm, algpseudocode}
\usepackage{cleveref}

\input{macro}

\definecolor{hiroshige}{HTML}{ffd06f}
\definecolor{algblue}{RGB}{0, 112, 192}
\definecolor{darkBlue}{RGB}{10,50,220}

\definecolor{darkBlue}{RGB}{10,50,220}
\definecolor{customRed}{RGB}{190,110,113}
\definecolor{customGreen}{RGB}{70,170,80}
\definecolor{customPurple}{RGB}{102,2,60}
\definecolor{lightgray}{gray}{0.92}

\hypersetup{
	colorlinks=true,
	linkcolor=customGreen,
	filecolor=magenta,      
	urlcolor=customPurple,
	citecolor=customRed,
}

\title{%
\smash{%
  \raisebox{-2.2ex}{%
    \includegraphics[height=13mm]{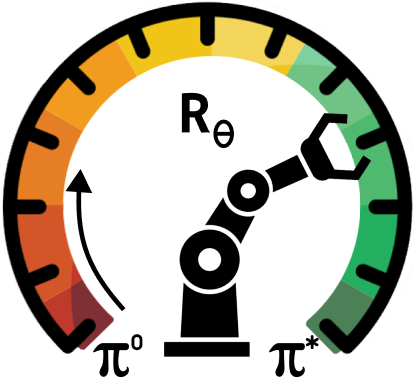}%
  }%
}%
\hspace{0.2em}%
\method: Scaling General-Purpose Robotic\\ \hspace{0.45em}%
Reward Models via Trajectory Comparisons}

\begin{document}

\author{
Anthony Liang$^{\star 1}$,
Yigit Korkmaz$^{\star 1}$,
Jiahui Zhang$^{2}$,
Minyoung Hwang$^{3}$,
Abrar Anwar$^{1}$,
Sidhant Kaushik$^{4}$\\
Aditya Shah$^{5}$,
Alex S. Huang$^{2}$,
Luke Zettlemoyer$^{5}$,
Dieter Fox$^{5,6}$,
Yu Xiang$^{2}$,
Anqi Li$^{7}$\\
Andreea Bobu$^{3}$,
Abhishek Gupta$^{5}$,
Stephen Tu$^{\dag 1}$,
Erdem Bıyık$^{\dag 1}$,
Jesse Zhang$^{\dag 5}$\\[4pt]
{\small
$^{1}$Univ. of Southern California \quad
$^{2}$UT Dallas\quad
$^{3}$MIT \quad
$^{4}$Indep. Researcher \quad
$^{5}$Univ. of Washington \quad
$^{6}$Ai2 \quad
$^{7}$NVIDIA
}\\
{\small
$^\star$Equal contribution \quad
$^\dag$Equal advising \quad
}
}

\makeatletter
\let\@oldmaketitle\@maketitle%
\renewcommand{\@maketitle}{\@oldmaketitle%
    \includegraphics[width=\linewidth]{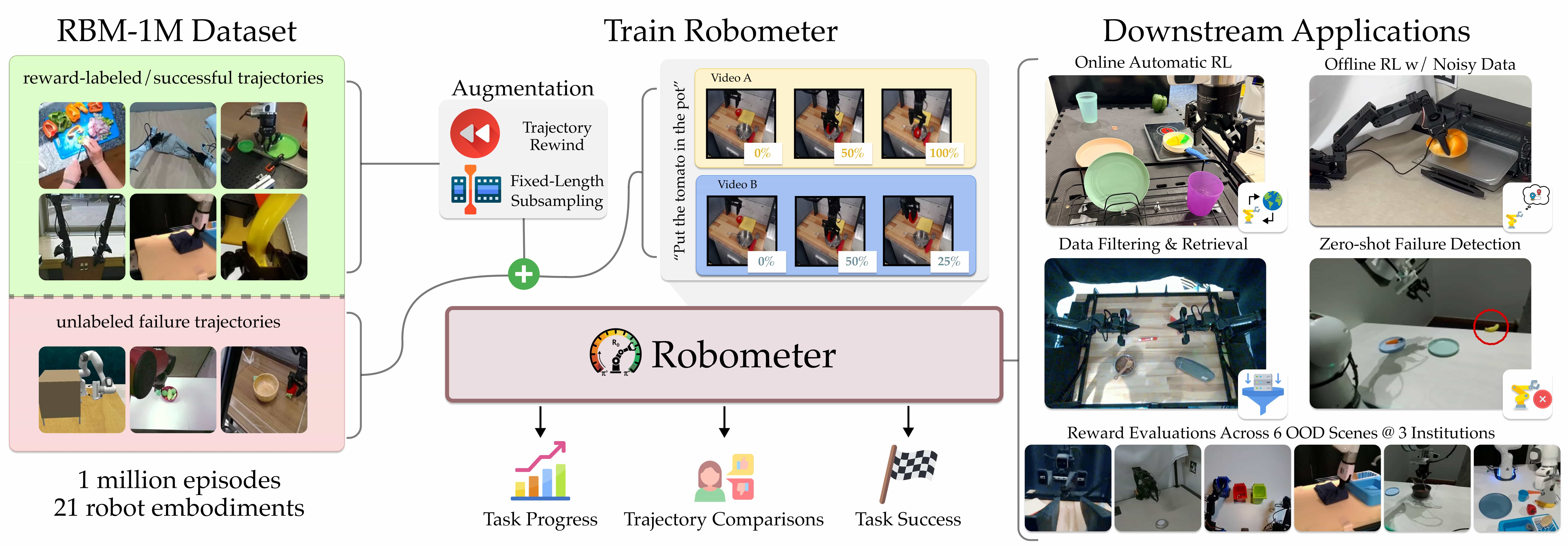} \\[0.25em]
   \refstepcounter{figure}\footnotesize{{Fig. 1:} \textbf{\method\ Overview.}
\method\ is trained on \datasetname, a 1M-trajectory dataset spanning 21 robot embodiments, containing both reward-labeled/expert trajectories and reward-unlabeled, failed trajectories.
The model is supervised with a dual objective: predicting frame-level task progress (reward) and learning trajectory-level preferences from pairwise comparisons. To help with downstream RL, it is also trained to predict per-frame task success.
This training recipe enables scalable reward learning, is validated on reward model evaluations from 6 out-of-distribution scenes collected at 3 institutions, and supports diverse downstream applications such as offline \& online RL, imitation learning data filtering and retrieval, and automated failure detection.}
  \label{fig:teaser} \medskip \vspace{-10pt}}%
\makeatother
\maketitle
\thispagestyle{plain}
\pagestyle{plain}

\bstctlcite{max8authors}

\begin{abstract}
Current general-purpose robot reward models rely on frame-level progress labels from expert demonstrations. This approach scales poorly to large datasets, where suboptimal or failed trajectories are abundant and absolute progress is ambiguous. To address this, we introduce \method{}, a scalable reward modeling framework combining intra-trajectory progress supervision with inter-trajectory preference supervision. \method{} utilizes a dual objective: a frame-level loss that anchors reward magnitude to expert data, and a trajectory-comparison preference loss that imposes global ordering constraints. This enables effective learning from both successful and failed trajectories. To support this formulation at scale, we curate \datasetname{}, a dataset of over one million multi-embodiment trajectories containing extensive suboptimal and failure data. Across benchmarks and real-world evaluations, \method{} learns highly generalizable reward functions and improves downstream robot learning performance.
Code, model weights, and videos at \url{https://robometer.github.io/}.

\end{abstract}

\renewcommand\thefigure{\arabic{figure}}
\setcounter{figure}{1}

\IEEEpeerreviewmaketitle

\input{sections/intro}

\input{sections/related}
\input{sections/method}

\input{sections/experiments}
\section*{Acknowledgments}
We thank Chuning Zhu and Kaiyuan (Kyle) Zheng for testing out \method\ integrated with DreamZero and providing us with starter code for our model-based RL experiment.
Additionally, we thank Harine Ravichandiran for helping supervise an online RL run and Matthew Hong for helping us with task selection for our LIBERO RL ablation experiments.

Anthony L, YK, and EB acknowledge funding by the Airbus Institute for Engineering Research (AIER);
Jesse Z, LZ, and DF acknowledge support by the Cross-Pacific AI Initiative from Amazon;
Jesse Z and AG are partly supported by funding from Amazon FAR through the Amazon Science Hub;
Jesse Z and DF acknowledge support by Amazon and Toyota Research Institute;
MH and AB acknowledge funding from the Tata Group via the MIT Generative AI Impact Consortium (MGAIC) Award; 
Jiahui Z and YX acknowledge funding from the National Science Foundation (NSF) under Grant No. 2520553. 

Additionally, we thank the UW Hyak and Tillicum high-performance computing clusters and USC Center for Advanced Research Computing (CARC) for providing us with compute resources.

\newpage

\bibliographystyle{IEEEtranN}
\bibliography{refs}

\clearpage

\begingroup
\renewcommand{\thesubsectiondis}{\arabic{subsection}.} 
\renewcommand{\thesubsection}{\Alph{section}-\arabic{subsection}}

\crefalias{section}{appendix}
\crefname{appendix}{Appendix}{Appendices}
\Crefname{appendix}{Appendix}{Appendices}

\input{sections/appendix/appendix_main}
\endgroup

\end{document}

%% file: macro.tex
\newcommand{\tok}{\operatorname{Tok}}
\newcommand{\prefhead}{\text{MLP}_\text{pref}}
\newcommand{\proghead}{\text{MLP}_\text{progress}}
\newcommand{\successhead}{\text{MLP}_\text{success}}
\newcommand{\vidstart}{\langle|\text{video\_start}|\rangle}
\newcommand{\prefhidden}{h_{\langle|\text{pref\_token}|\rangle}}

\newcommand{\preftoken}{\langle|\text{pref\_token}|\rangle}

\definecolor{robometer}{HTML}{B20000}

\newcommand{\method}{\textsc{Robometer}}

\newcommand{\datasetname}{\texttt{RBM-1M}}
\newcommand{\evaldatasetid}{\texttt{RBM-EVAL-ID}}
\newcommand{\evaldatasetood}{\texttt{RBM-EVAL-OOD}}

\newcommand{\usc}{USC}
\newcommand{\mituniv}{MIT}
\newcommand{\utd}{UTD}

\newcommand{\splittoken}{\langle|\text{split\_token}|\rangle}
\newcommand{\videostarttoken}{\langle|\text{video\_start}|\rangle}
\newcommand{\progtoken}{\langle|\text{prog\_token}|\rangle}

\definecolor{vlac}{HTML}{76A3A0}      %
\definecolor{gvl}{HTML}{95A06A}       %
\definecolor{rewind}{HTML}{BE8D96}    %
\definecolor{roboreward}{HTML}{9C92C0}%
\definecolor{robodopamine}{HTML}{6FD999} %
\definecolor{topreward}{HTML}{db8712} %

\newcommand{\vlaccolor}[1]{\textcolor{vlac}{#1}}
\newcommand{\gvlcolor}[1]{\textcolor{gvl}{#1}}
\newcommand{\rewindcolor}[1]{\textcolor{rewind}{#1}}
\newcommand{\roborewardcolor}[1]{\textcolor{roboreward}{#1}}
\newcommand{\robometercolor}[1]{\textcolor{robometer}{#1}}
\newcommand{\robodopaminecolor}[1]{\textcolor{robodopamine}{#1}}
\newcommand{\toprewardcolor}[1]{\textcolor{topreward}{#1}}

%% file: sections/intro.tex
\section{Introduction}
In human cognition, comparative judgments are a core mechanism for internalizing calibrated scales~\citep{Laming1984TheRO, Stewart2005AbsoluteIB, sharif2016relativememory}, enabling reasoning about relative progress and outcomes rather than isolated states. Analogously, the supervision signals used to train robotic reward models determine how well they internalize notions of task progress, enabling downstream applications such as online reinforcement learning (RL)~\citep{yang2024rank, zhang2025rewind}, imitation learning (IL) from noisy data~\citep{ma2024generative, chen2025sarm}, automated failure detection~\citep{gu2025safe}, and offline RL~\citep{venkataraman2025rlmvlmoffline}. Current general-purpose robot manipulation reward models rely exclusively on absolute progress labels derived from expert or reward-labeled demonstrations, providing pointwise, \emph{trajectory-local} supervision~\citep{zhang2025rewind, zhai2025VLAC, lee2026roboreward, tan2025robodopamine}.

Such labels are easy to obtain for expert trajectories—for example, by linearly interpolating progress from 0 to 1—but become ill-defined and costly to annotate for failed attempts, where progress may fluctuate over time. As a result, large amounts of suboptimal data—ubiquitous in real-world robot learning—cannot be effectively leveraged~\citep{tian2026position}. This reliance on trajectory-\emph{local} progress supervision limits both scalability and generalization. In this work, we address this limitation by training reward models with an additional \emph{global} supervision signal that improves generalization across embodiments, scenes, and varying trajectory quality.

Our key insight is that \emph{preference prediction} over trajectory pairs provides complementary supervision. While progress labels anchor reward values along individual trajectories, pairwise comparisons impose ordering constraints across diverse trajectories, tasks, robots, and viewpoints. This formulation enables learning from previously unusable suboptimal data by requiring only relative comparisons—curated without additional human annotation—rather than absolute scores. Specifically, trajectory comparison supervision (1) enforces consistent ordering across trajectories, providing global grounding beyond individual rollouts, and (2) scales naturally to unlabeled failed trajectories where absolute progress is ambiguous, resulting in better-calibrated rewards.

To instantiate this insight, we propose a scalable training recipe for reward modeling based on a dual reward-prediction objective: a frame-level progress loss on expert demonstrations and a preference-prediction loss over trajectory comparisons. Using this recipe, we train \method, a manipulation-centric reward model. 
Our training recipe reveals a strong mutual reinforcement effect between the two supervision signals. Even when trained only on expert demonstrations, preference supervision improves \method's ability to distinguish successful from suboptimal trajectories, suggesting that global comparative constraints induce a more structured internal reward representation. Moreover, as additional unlabeled suboptimal data is introduced, \method\ naturally scales to further improve performance.

To train \method, we curate \datasetname, a large-scale reward-learning dataset for robot manipulation containing over one million trajectories collected from 21 robot platforms, including single-arm, bimanual, and mobile manipulators, as well as human demonstrations (see \Cref{fig:teaser}). Importantly, \datasetname\ intentionally includes a substantial number of suboptimal and failed trajectories that naturally arise during real-world data collection but are difficult to leverage with purely absolute progress-based supervision. The scale and diversity of \datasetname\ are therefore critical for learning globally consistent preference relations across embodiments, tasks, and viewpoints while fully exploiting failure data that would otherwise be discarded. Beyond real-world failures, we further construct preference pairs using a suite of augmentations---including video rewinding~\citep{zhang2025rewind} and cross-task comparisons---that expose the model to diverse successful and suboptimal behaviors. Across external benchmarks and our own evaluation trajectories collected from six out-of-distribution scenes spanning three institutions, \method\ outperforms state-of-the-art baselines by an average of 14\% in reward rank correlation and achieves a 32\% relative improvement in distinguishing successful from suboptimal trajectories.

Finally, we show that \method\ outperforms relevant baselines in real-world robot learning applications across a diverse set of downstream applications: (1) automatic online RL, (2) offline RL with noisy and expert trajectories, (3) dataset filtering for imitation learning, and (4) zero-shot failure detection across multiple robot embodiments and institutions. 
Overall, policy learning experiments with \method\ demonstrate $\mathbf{{2.4}}-\mathbf{{4.5\times}}$ higher success rates than the best baseline in each category.
We publicly release \method, \datasetname, and the training code at \url{https://robometer.github.io}, enabling the community to train their own models using our recipe or to directly use \method\ for robot manipulation applications.

%% file: sections/related.tex
\section{Related Work}
\label{sec:related}
Learning reward functions is a central problem in reinforcement learning and robotics, and prior work has explored a wide range of approaches that differ in both the form of supervision and the scope of generalization.  

\textbf{Reward Learning from Demonstrations.} Learning reward functions from human supervision is a long-studied topic in inverse RL (IRL), where reward functions are inferred from human demonstrations~\citep{ng2000IRL, abbeell2004IRL, ziebart2008maxentirl, finn2016gcl, bobu2021feature} or from expert and goal-state distributions~\citep{ho2016generative, leiImageGAIL2018, fu2017learning, fu2018VICE, fu2024robot, jain2025sailor}. 
However, most IRL methods require task-specific expert demonstrations, necessitating the collection of new demonstrations whenever the task or reward specification changes.
In contrast, we propose a framework for training reward models that can generalize effectively to new tasks.

\textbf{Preference-Based Reward Learning.}
Prior work in psychology has observed that humans often use relative comparisons over absolute numerical scales when making judgments~\citep{Laming1984TheRO, Stewart2005AbsoluteIB, sharif2016relativememory}. 
This insight has been central to modern reinforcement learning from human feedback (RLHF), where preference supervision from humans is used to learn reward models that are only identifiable up to monotone transformations, yet sufficient for effective policy optimization~\citep{christiano2017rlhf, sadigh2017active, bajcsy2018learning, biyik2020active, lee2021pebble, hejna2022fewshot, myers2021learning, yang2024trajectory, korkmaz2025mile}. 
In contrast to these works, which treat preferences as the primary supervision signal to train domain-specific reward models, we generate preference supervision from both synthetic and real trajectories for which assigning progress labels is difficult. 
We use this preference signal as an \emph{auxiliary} objective that complements direct progress prediction, enabling \method\ to learn from large, heterogeneous datasets without additional human preference supervision.
Closest in spirit, \citet{kwok2025robomonkey} generate synthetic preference labels from action mean-squared-error to train a robot action verifier, and \citet{wangrlvlmf2024,venkataraman2025rlmvlmoffline} use vision-language models (VLMs) to generate frame-level preferences as the primary supervision signal for task-specific reward models.
In contrast, we generate preference supervision by comparing entire trajectories and use it as an auxiliary signal to learn a general-purpose reward function across tasks and embodiments.

\textbf{Rewards from Foundation Models.}
Finally, recent work has sought to construct more general reward functions that operate directly on images, videos, and language. 
Large language models (LLMs) and VLMs have been applied to reward design, for example by generating executable reward code or shaping functions from natural language descriptions~\citep{ma2023eureka, yu2023language, xie2024textreward}, directly producing reward functions~\citep{kwon2023reward}, or guiding reward computation via language-conditioned state masking~\citep{hwang2025maskedirl}.
However, most of these approaches assume access to privileged state information, which is often unavailable or difficult to obtain in real-world robotic deployment settings. 

To overcome this challenge, some approaches use task progress as a proxy reward, either by applying pre-trained VLMs as zero-shot progress or success estimators~\citep{cui2022can, mahmoudieh2022zero, roboclip, du2023vision, ma2023vipuniversalvisualreward, adeniji2023languagerewardmodulationpretraining, furl, guan2024tasksuccessenoughinvestigating, ma2024generative, rocamonde2024visionlanguage, chen2026topreward}, or by training domain-specific models with progress-prediction objectives~\citep{cui2022can, fan2022minedojo, DECKARD2023, nam2023liftunsupervisedreinforcementlearning, ma2023liv, yang2024rank, hung2025victor, kim2025reds, zhang2025rewind, chen2025sarm}. 
However, directly using pre-trained VLMs for zero-shot video-language reward prediction often yields noisy or inconsistent signals~\citep{zhang2025rewind, lee2026roboreward, budzianowski2025opengvl}, while smaller per-task models tend to overfit to domain-specific visual and semantic features, limiting generalization.

Recent methods address these limitations by fine-tuning VLMs or VLAs, enabling progress prediction to leverage visual–semantic representations learned from diverse data~\citep{zhai2025VLAC, ghasemipour2025selfimproving, intelligence2025pi06vla, lee2026roboreward, tan2025robodopamine, zhang2026progresslmprogressreasoningvisionlanguage} and demonstrating improved sample efficiency for RL.
VLAC~\citep{zhai2025VLAC}, for example, co-trains a VLA with relative progress difference targets, while $\pi_{0.6}^*$~\citep{intelligence2025pi06vla} and \citet{ghasemipour2025selfimproving} train distance-to-goal value functions that encode task progress.
RoboReward~\citep{lee2026roboreward} fine-tunes a VLM to predict discretized (1-5) progress labels generated via counterfactual instruction labeling by closed-source VLMs, and RoboDopamine~\citep{tan2025robodopamine} similarly fine-tunes a VLM for progress prediction but requires a goal image in addition to the language instruction and task-specific one-shot fine-tuning.
In contrast, we introduce a more scalable training framework which uses an explicit auxiliary \emph{preference prediction} objective, enabling us to scale to datasets containing suboptimal or failed trajectories without reward labels. Using such failed trajectories, even as just in-context learning examples for pre-trained VLMs, has been demonstrated to help reward models align more closely with human-specified rewards~\citep{tian2026position}.

%% file: sections/method.tex
\begin{figure*}
    \centering
    \includegraphics[width=\linewidth]{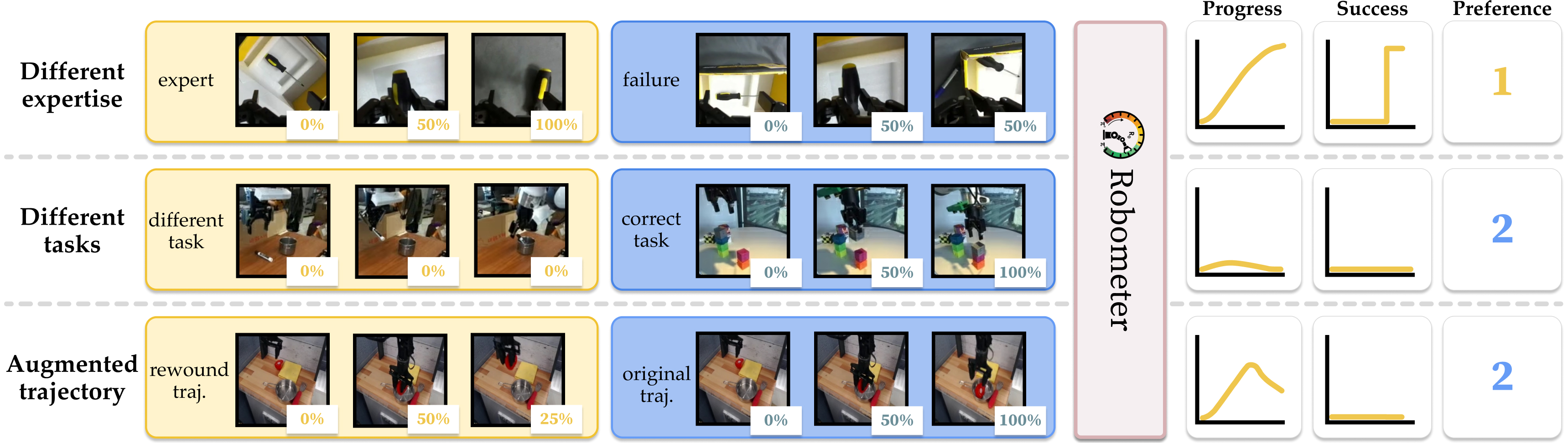}
    \caption{\textbf{\method} is a VLM-based reward model that predicts dense, per-frame progress-based rewards and success labels for the first of two video trajectories. To scalably train with failure data, we also predict which of the two video trajectories better completes the task (preference). We use three strategies for curating training examples from our given datasets, which are further detailed in Section~\ref{sec:augmentation} with model architecture shown in Appendix \Cref{fig:architecture}. At inference, the model can input either up to two trajectories: with a single trajectory, it outputs per-frame progress \& success predictions; with two, it additionally predicts a preference for which trajectory best performs the task.} 
    \label{fig:training_objectives}
\end{figure*}
\section{Scalable Reward Model Training}
We propose a \emph{scalable recipe} for training reward models, along with \robometercolor{\method}, an instantiation of this recipe that provides dense rewards for robot manipulation. Our approach rests on three pillars: a diverse dataset which includes unlabeled failure trajectories, a pre-trained VLM backbone for generalization, and a training objective that combines \textbf{per-frame rewards} with \textbf{trajectory preferences}.

\subsection{\datasetname\ Dataset}
\label{sec:dataset}
\textbf{Notation.} We define the dataset $\mathcal{D} = \{\tau_i\}$ of trajectories, where each $\tau = \{o_{1:T}, l, p\}$ contains image observations $o$, a language instruction $l$, and a scalar progress label $p \in \{\text{None}, [0, 1]\}$ corresponding to the progress at the end of the trajectory. For expert demos, $p=1.0$; for datasets with partial progress labels (e.g., RoboArena~\citep{atreya2025roboarena}), we use the provided score. For unlabeled failed trajectories, we set $p = \texttt{None}$.

\textbf{Data Composition.} Rather than maximizing trajectory quantity, \datasetname\ focuses on viewpoint, scene, and embodiment diversity. We aggregate 1 million trajectories from: (1) \textbf{Expert robot data} from diverse, multi-robot sources such as Open-X~\citep{open_x_embodiment_rt_x_2023} and subsets of high-quality, single-robot datasets such as AGIBotWorld~\citep{contributors2024agibotworldrepo}; (2) \textbf{Human videos} from datasets such as Epic-Kitchens~\citep{Damen2022EpicKitchens} for scene diversity or human-robot paired datasets like RH20T~\citep{fang2023rh20t} to promote embodiment-invariant representations; (3) \textbf{Simulation} data from sources like LIBERO~\citep{liu2023libero}; and (4) \textbf{Failed trajectories} from automated policy rollouts~\citep{zhou2024autonomous} and failure-detection datasets~\citep{lin2025failsafe}.

Our dataset overall includes 21 robot embodiments and over 1 million trajectories, hence \datasetname{}.
We also construct two evaluation datasets,  \evaldatasetid\ and \evaldatasetood, detailed in \Cref{sec:app:data:eval_datasets}.
For further details on dataset filtering and aggregation, see \Cref{sec:app:data}. We also list all dataset categories and trajectory counts in Appendix \Cref{fig:appdx:pie_chart}, \Cref{tab:appdx:training_datasets}.

\subsection{\method\ Architecture and Tokenization}
\label{sec:vlm model}
\method\ buids on a causally masked VLM, \textsc{Qwen3-VL-4B-Instruct}, to process either one video (for reward inference) or a pair of videos (for preference training). 

\textbf{Hidden Embedding Extraction.} To extract rewards without disrupting the VLM's pre-trained internal representations, we insert new, learned tokens into the sequence. We interleave \textbf{progress tokens} ($\progtoken$) within the first video sequence and a single \textbf{preference token} ($\preftoken$) at the end of the multi-video prompt:
\begin{equation}
\resizebox{0.91\columnwidth}{!}{$
    \begin{aligned}
    \tok(l, o^1, o^2) \rightarrow & \tok(l) \videostarttoken \left[ \tok(o^1_t) \progtoken \right]_{t=1}^{T} \\
    & \splittoken \left[ \tok(o^2_t) \right]_{t=1}^T \preftoken,
    \end{aligned}
$}
\label{eq:tokenized_prompt}
\end{equation}
where $\videostarttoken$ is the model's default image-start delimiter and $\splittoken$ is a separator. 
The causal mask ensures that $\progtoken$ tokens attend only to current/previous frames of $o^1$, producing dense, frame-level progress estimates for online reward inference, while $\preftoken$ attends to both trajectories to make a relative judgment.
We fix both trajectories to length $T=8$ to avoid preference predictions that rely on trajectory length as a proxy for quality. Progress tokens are inserted only for $o^1$ since at inference time, progress is predicted for a single trajectory; furthermore, if we insert progress tokens between $o^2$ frames, they would attend to $o^1$.

\subsection{Training Objectives}
\label{sec:objective}
We optimize \method\ using a composite loss: $\mathcal{L} = \mathcal{L}_{\text{pref}} + \mathcal{L}_{\text{prog}} + \mathcal{L}_{\text{succ}}$. This allows the model to anchor rewards to absolute progress while learning to distinguish subtle quality differences through trajectory comparisons across the dataset.

\textbf{Preference Prediction.} We train a binary classifier, $\prefhead$, on the hidden state $\prefhidden$ of the $\preftoken$ to predict which trajectory better satisfies $l$:
\begin{equation}
\resizebox{0.91\columnwidth}{!}{$
\begin{aligned}
\mathcal{L}_{\mathrm{pref}}
= & - \Big[
\mathbb{I}_{y = 1}\,\log \sigma\!\left(\prefhead(\prefhidden)\right) \\
& \quad + \mathbb{I}_{y = 2}\,\log\!\left(1 - \sigma\!\left(\prefhead(\prefhidden)\right)\right)
\Big],
\end{aligned}
$}
\label{eq:preference}
\end{equation}
where $y$ is the ground-truth preferred trajectory.

\textbf{Progress and Success.} For the first trajectory $o^1$, we attach an MLP head to each $h_{\progtoken, t}$ to predict continuous progress $p_t$ and binary success $s_t$.
Similar to prior work~\citep{zhang2025rewind, zhai2025VLAC, chen2025sarm}, we define per-frame progress targets $p_{1:T}$ for expert demonstration data, where the final target progress $p=1$.
To better model multi-modal reward/progress distributions than continuous progress prediction, we discretize progress into $N=10$ uniformly spaced bins over $[0, 1]$ and model progress prediction as a categorical distribution~\citep{bellemare2017distributional, farebrother2024stop, intelligence2025pi06vla}.
For a trajectory of length $T$, the ground-truth continuous progress target at frame $t$ is defined as $p_t = t / T$ for $t \in \{1,\ldots,T\}$.
This scalar target is projected onto a categorical distribution over $N$ bins using linear interpolation between the 2 neighboring bin centers.
The progress head $\proghead$ outputs a categorical distribution $\hat{p}_t \in \Delta^{N}$, and the progress loss is computed using cross-entropy:
\begin{equation*}
\mathcal{L}_{\text{prog}} =
\frac{1}{T} \sum_{t=1}^{T}
\text{CE}\!\left(\text{Project}(p_t),\, \proghead(h_{\progtoken, t})\right).
\end{equation*}

At inference time, a continuous progress estimate is recovered by taking the expectation over the bin centers, $\hat{p}_t = \sum_{i=1}^{N} z_i \, \hat{p}_{t,i}$, where $\{z_i\}_{i=1}^{N}$ denote the fixed bin centers.
Per-frame success targets are defined such that $s_t = 0$ for $t < T$ and $s_t = 1$ for $t = T$.
In some datasets, the human operator stops recording a trajectory several frames after the task has already been completed; to improve the accuracy of both progress and success prediction, we sample 10 trajectories per data source to determine the point at which the task actually ends (Appendix~\Cref{sec:app:data:datasets}).
We train success prediction with binary cross-entropy on $s$, with balanced class weights adjusted per-batch to account for negative sample imbalance:
\begin{equation*}
    \resizebox{0.9\columnwidth}{!}{$\mathcal{L}_{\text{succ}} = \text{BalancedBCE}(s_{1:T}, [\successhead(h_{\progtoken, t})]_{1:T})$}.
\end{equation*}

\subsection{Data Sampling and Augmentation}
\label{sec:augmentation}
Given these losses, the ideal training regime for \method\ would rely on large-scale, preference-labeled robot trajectory datasets containing explicit progress-labeled failures. Such failures are particularly important because, at deployment time, reward models are frequently queried on out-of-distribution trajectories---e.g., failures induced by online RL exploration, compounding execution errors, or noisy data collection---that deviate substantially from the training distribution. 
In practice, however, preference annotations over robot trajectories are limited, and dense per-frame progress labels for failed executions are expensive and difficult to obtain.
We address this limitation by constructing training inputs $(l, o^1, o^2)$ and targets $y$ dynamically from $\datasetname$ using three complementary strategies displayed in \Cref{fig:training_objectives}:

\begin{figure*}[ht]
    \centering
    \includegraphics[width=\linewidth]{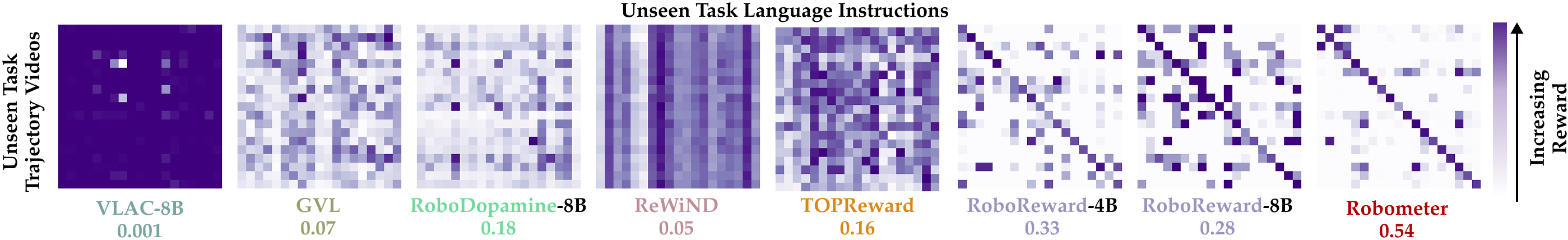}
    \caption{\textbf{Video-Language Reward Confusion Matrix.} For each task sampled at random from \emph{self-collected, unseen} data 
    from \evaldatasetood, we compute rewards for all combinations of demonstration videos and language descriptions. \robometercolor{\method} produces the most diagonal-heavy confusion matrix, indicating strong alignment between unseen demos and instructions. We also report 
    the column-normalized diagonal mean under each model, which represents the fraction of the model’s total reward for aligned task and video pairs.}
    \label{fig:exp:reward:confusion}
\end{figure*}

\begin{enumerate}
    \item \textbf{Progress-Based Comparisons (Different Expertise).} 
    To teach the model to distinguish execution quality, we sample two trajectories $\tau_1, \tau_2$ sharing an instruction $l$ but differing in outcome (e.g., an expert demonstration $p{=}1$ vs. an unlabeled failure $p{=}\texttt{None}$) or progress ($p^1{\neq}p^2$). We set the preference target $y{=}1$ if $p^1 {>} p^2$ (or if $\tau_1$ is the expert), and $y=2$ otherwise. This allows \method\ to leverage unlabeled failures by contrasting them against successful demonstrations.

    \item \textbf{Instruction Negatives (Different Tasks).} 
    To ensure rewards are grounded in the language command, we sample $\tau_1$ and $\tau_2$ with distinct instructions $l^1{\neq}l^2$. We randomly select one instruction as the conditioning text $l$, set the preference label $y$ to the trajectory corresponding to the selected instruction, and set the progress target $p_t{=}0$ for the other, enforcing that correct behavior for the wrong task yields no reward.

    \item \textbf{Video Rewind (Augmented Failures).} 
    To explicitly model ``undoing'' progress---a common failure mode in RL---we generate synthetic preferences from a single expert trajectory $\tau$ by reversing a segment of time. Prior work denotes this type of augmentation as \emph{video rewind}~\citep{zhang2025rewind, chen2025sarm, huang2025adapowerspecializingworldfoundation}. We sample indices $1 \le t_1 < t_2 < t_3 \le T$ to form a \emph{Chosen} forward sequence $o^c = o_{t_1:t_3}$ and a \emph{Rejected} rewound sequence, either $o^r = [o_{t_1:t_3}, o_{t_3-1:t_2}]$ or $[o_{t_3:t_1}]$. 
    The Chosen and Rejected sequences are randomly assigned to $o^1$ and $o^2$ to construct the preference label.
    While $o^c$ targets linear forward progress, we explicitly penalize the reversal in $o^r$ by assigning decreasing progress targets matched to their frame indices.
\end{enumerate}

\textbf{Fixed-length Subsampling.}
To avoid biasing toward longer trajectories, we construct fixed-length inputs by randomly selecting start and end indices in the trajectory, and uniformly sampling $T$ frames between them. \footnote{\method\ is still a \emph{dense}, per-frame reward model: during inference, we query over $o_{1:t} \; \forall t \in [1, ..., T]$ for a trajectory of length $T$ to produce $T$ rewards. Our subsampling always includes the first and current frame.} 

\textbf{Summary.}
\method\ builds upon a base VLM, \textsc{Qwen3-VL-4B-Instruct}, and inserts additional learnable tokens 
to predict preference, progress, and success.
We train \method\ on \datasetname, a dataset consisting of both progress-labeled and progress-unlabeled data, by sampling trajectory comparisons across the entire dataset. 
These comparisons come from failure trajectories, comparisons across different tasks, and rewound videos.
At inference, \method\ can operate on either one or two trajectories: with a single trajectory, it outputs per-frame progress and success predictions; with two trajectories, it additionally predicts a preference while still producing per-frame predictions for the first trajectory.
For additional model implementation and training details, see \Cref{sec:app:model}.

%% file: sections/experiments.tex
\begin{table*}[htb]
  \centering
  \small
  \renewcommand{\arraystretch}{1.2}
  \begin{adjustbox}{max width=\linewidth}
  \begin{tabular}{ll cccc ccc cc}
  \toprule
  & & \multicolumn{4}{c}{w/ Varied Training Data} & \multicolumn{3}{c}{w/ RoboReward Training Data} & \multicolumn{2}{c}{w/ our \datasetname\ data} \\
  \cmidrule(lr){3-6} \cmidrule(lr){7-9} \cmidrule(lr){10-11}
  & & \textbf{\gvlcolor{GVL}} & \textbf{\vlaccolor{VLAC}} & \textbf{\toprewardcolor{TOPReward}} & \textbf{\robodopaminecolor{RoboDopamine}-8B} & \textbf{\roborewardcolor{RoboReward}-4B} & \textbf{\roborewardcolor{RoboReward}-8B} &
  \textbf{\robometercolor{\method{}}} & \textbf{\rewindcolor{ReWiND}} & \textbf{\robometercolor{\method}} \\
  \midrule
  \multirow{2}{*}{\textbf{(a) VOC $r \uparrow$}}
  & RBM-EVAL-ID
  & 0.16 & 0.16 & 0.42 & 0.79 & 0.77 & 0.82 & 0.84 & 0.46 & \textbf{0.92} \\

  & RBM-EVAL-OOD
  & 0.21 & 0.17 & 0.40 & 0.80 & 0.88 & 0.88 & 0.93 & 0.51 & \textbf{0.94} \\

  \midrule

  \multirow{1}{*}{\textbf{(b) Kendall $\tau_a \uparrow$}}
  & RBM-EVAL-OOD
  & 0.19 & 0.08 & 0.13 & 0.45 & 0.50 & 0.47 & 0.55 & 0.01 & \textbf{0.64} \\
  \bottomrule
  \end{tabular}
  \end{adjustbox}
  \caption{(a) Reward alignment (VOC Pearson $r$) and (b) trajectory ranking (Kendall $\tau_a$) on RBM-EVAL datasets.
  Baselines are split into categories based on training data: \gvlcolor{GVL} and \toprewardcolor{TOPReward} training data is unknown, \vlaccolor{VLAC} is trained on a 300k-trajectory dataset, and \robodopaminecolor{RoboDopamine} is trained on a
  100k-trajectory dataset.
  We compare \robometercolor{\method} against \roborewardcolor{RoboReward-4B/8B} with their own training data, and we also evaluate \rewindcolor{ReWiND} and \robometercolor{\method} trained with the full \datasetname\ dataset. Kendall $\tau_a$
  is not calculated for \evaldatasetid\ due to it only having simulation failure data.}
  \label{tab:exp:reward}
  \end{table*}

\begingroup %
\renewcommand{\thesubsectiondis}{Q\arabic{subsection}:} 
\renewcommand{\thesubsection}{\Roman{section}-Q\arabic{subsection}} 
\section{Experiments}
\label{sec:experiments}
Our experiments aim to study \method's effectiveness in producing rewards for robot learning. Specifically, we organize our experiments to answer the following questions:
\begin{enumerate}[label=(\textbf{Q\arabic*}), nosep, leftmargin=0pt, labelindent=0pt, itemindent=*]
    \item \label{q1}\textbf{Reward Evaluation}: How well do \method\ rewards reflect task progress on \emph{unseen} tasks and embodiments?
    \item \label{q2}\textbf{Ablation + Analysis}: How much does each component of \method\ contribute to reward performance?
    \item \label{q3}\textbf{Policy Learning}: How does \method\ compare against baselines in enabling downstream robot learning? 
\end{enumerate}

\textbf{Baselines.}
We compare \robometercolor{\method} against the strongest set of video-language input, zero-shot-capable, and open-sourced or API-accessible reward baselines described in~\Cref{sec:related}. 
We list a dataset size comparison table for baselines and related papers in Appendix \Cref{tab:traj_comparison}.
\begin{itemize}[leftmargin=*]
    \item \textbf{\vlaccolor{VLAC-8B}} \citep{zhai2025VLAC} trains a VLA that predicts actions and rewards on a dataset of 300k human and robot trajectories. We compare against their larger 8B parameter checkpoint.
    \item \textbf{\gvlcolor{GVL}} \citep{ma2024generative} prompts a pre-trained closed-source LLM with shuffled video frames to predict task progress for subsampled frames across the video sequence. We use GPT-5 mini as it is the best-performing closed-source model on the RoboRewardBench reward evaluation benchmark~\citep{lee2026roboreward}.
    \item \textbf{\toprewardcolor{TOPReward}} \citep{chen2026topreward} prompts a pre-trained VLM (Qwen-3-VL 8B) with ``does the trajectory complete the task?'' and uses the log-likelihood of the ``true'' token as the reward. 
    \item \textbf{\rewindcolor{ReWiND}}~\citep{zhang2025rewind} trains a small transformer-based network with a direct progress prediction objective along with video rewinding to simulate failed policy rollouts.
    We train ReWiND with \datasetname\ to maximize its zero-shot capability.
    \item \textbf{\robodopaminecolor{RoboDopamine-8B}} \citep{tan2025robodopamine} fine-tunes a VLM for reward prediction via ``frame hops'' comparing forward and rewound frames. While designed for reward prediction conditioned on a \emph{goal image} and instruction, we evaluate it in our zero-shot setting without a goal image for fair comparison. We use the latest, 8B parameter, RoboDopamine-2.0 checkpoint.
    \item \textbf{\roborewardcolor{RoboReward-4B/8B}} \citep{lee2026roboreward}: Fine-tunes a Qwen-3-VL 4B/8B VLM for discrete (1-5) progress prediction on a dataset consisting of data from OXE~\citep{open_x_embodiment_rt_x_2023} and RoboArena evaluations~\citep{atreya2025roboarena}. Generates \emph{counterfactual} instructions via closed-source VLMs to simulate failed trajectories.
\end{itemize}

\textbf{Custom Evaluation Datasets.} We train \method, and certain baselines when applicable, on the aforementioned \datasetname\ dataset.
Prior large-scale reward modeling baselines mainly evaluate on validation or test set versions of the datasets they train on~\citep{zhai2025VLAC, zhang2025rewind, lee2026roboreward, tan2025robodopamine}, which contain in-distribution arms, camera angles, or scenes.
Instead, we collect our own evaluation dataset, \evaldatasetood, consisting of 976 trajectories collected from 3 academic institutions that are guaranteed not to be in the training data of \emph{any} baseline, consisting of 6 embodiments (including human hands), 3 of which are not in \datasetname, collected across diverse camera angles.
We also aggregate an in-distribution test split of unseen trajectories collected from datasets in \datasetname, denoted \evaldatasetid.
See more evaluation dataset details in \Cref{sec:app:data:eval_datasets}.

\subsection{Reward Evaluation}

\begin{figure*}
    \centering
    \includegraphics[width=\linewidth]{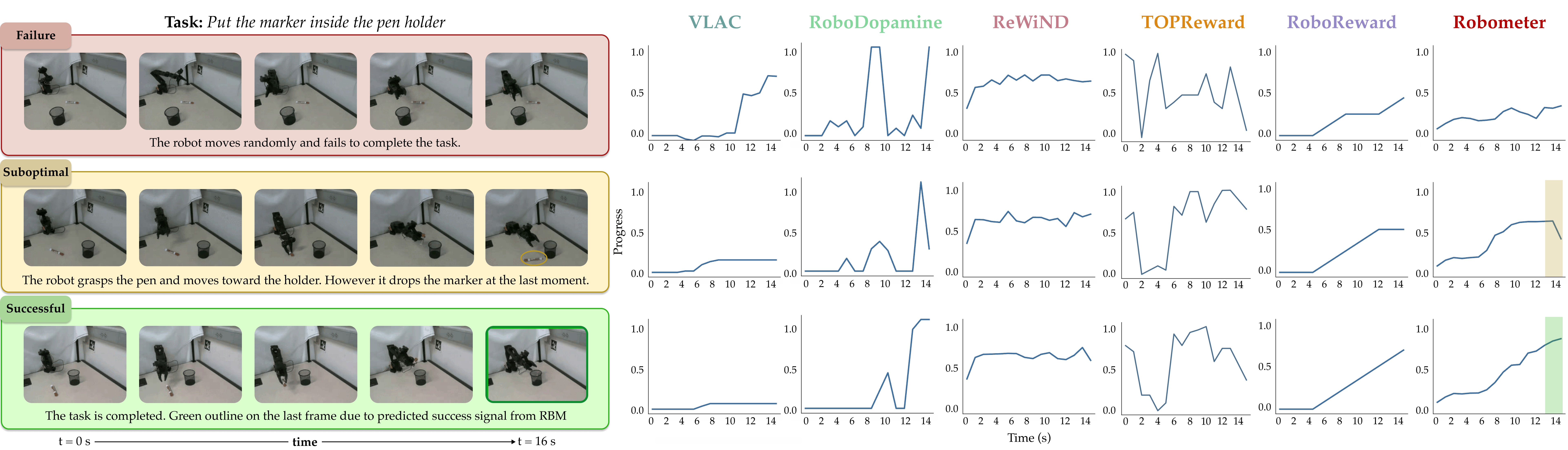}
    \caption{\textbf{Qualitative Analysis of Failure, Suboptimal and Successful Trajectories.} We visualize the progress predictions for three trajectories of different quality for the same task.
    Notably, for the suboptimal trajectory, \robometercolor{\method} predicts steadily increasing progress as the robot approaches the pen holder, but sharply reduces its progress estimate when the marker is dropped, correctly reflecting regression in task completion. 
    In contrast, \roborewardcolor{RoboReward} continues to assign high progress despite the task failure. 
    Finally, \robometercolor{\method} is the only model that correctly predicts task success for the successful trajectory
    (i.e., high final progress value and explicit success prediction).     
    }
    \label{fig:exp:reward:qualitative reward comparison}
\end{figure*}

As detailed in \Cref{sec:vlm model}, our focus is on training a reward model which: 
(1) generalizes to new tasks, embodiments, and domains while (2) providing reward feedback useful for \emph{policy learning}. We structure this subsection to highlight \method's strong performance across both criteria.

\textbf{Trajectory Task Alignment.} 
Our first main result demonstrates that \method\ accurately distinguishes between different tasks in \evaldatasetood, which directly reflects its ability to assign rewards that align with task semantics, even across unseen robot embodiments, camera viewpoints, and scenes.
We plot confusion matrices comparing unseen, successful trajectory videos versus their language instructions in \Cref{fig:exp:reward:confusion}.
Ideally, a \textcolor[HTML]{572b92}{\textbf{purple diagonal}} indicates correct video-instruction pairs, with low (white) values elsewhere. 
\method\ clearly produces the strongest disparity between the diagonal and off-diagonal elements, highlighting its superior ability to reward a robot for performing the \emph{correct} task, which is especially important in cluttered, multi-task settings.
This ability is in part due to how we sample \emph{different-task} negative preference and progress examples across the entire dataset (cf.~\Cref{sec:augmentation}).

\textbf{Reward Alignment.} Quantitatively, we evaluate the ability of baselines to predict increasing progress for \emph{successful} robot videos from both \evaldatasetood\ and \evaldatasetid\ in \Cref{tab:exp:reward}(a).
We report Value Order Correlation (VOC)~\citep{ma2024generative} $\in [-1, 1]$, which calculates the Pearson correlation of predicted rewards for each trajectory video frame against their ground-truth timestep value. 
Overall, \method\ performs the best across the board on both test sets, especially on \evaldatasetood.
We break down per-dataset and per-subset results for both test sets in \Cref{sec:app:reward_eval}, and evaluate on an external RoboRewardBench~\citep{lee2026roboreward} benchmark in \Cref{sec:app:reward_eval:roborewardbench}.

\textbf{Relative Trajectory Rankings for Mixed Expertise Data.} Next, we quantitatively demonstrate that \method\ is more effective than baselines at providing rewards useful for \emph{policy learning}.
For a robot policy to learn with rewards, the rewards should not only be high when performing the correct task, but also be low for incorrect execution.
We measure this using the Kendall-$\tau_a$ coefficient~\citep{muslimani2025towards}, an ordering metric $\in [-1, 1]$ robust to ties. 
We calculate the alignment between model-assigned final rewards and the ground-truth ordering between failed, suboptimal, and successful trajectories for the same task. 
A higher $\tau_a$ value demonstrates that the reward model more accurately distinguishes between levels of policy performance and thus can provide proper reward signals to the policy for both low- and high-quality behaviors.

We report results in \Cref{tab:exp:reward}(b). On \evaldatasetood, \method\ achieves a Kendall-$\tau_a$ of 0.66, substantially outperforming RoboReward-4B (0.50) and RoboReward-8B (0.47), indicating that \method\ more reliably recovers the correct ordering among failed, suboptimal, and successful trajectories. 
Notably, even when trained on the same data as RoboReward, \method\ outperforms RoboReward in both policy ranking and reward alignment, highlighting the effectiveness of our data augmentation strategies.

To further illustrate this behavior, we visualize reward predictions over time for failed, suboptimal, and successful trajectories in \Cref{fig:exp:reward:qualitative reward comparison}. 
\method\ exhibits sharper separation in rewards between different levels of execution and accurately reflects regression in task progress.
Additional results in \Cref{sec:app:reward_eval} evaluating preference prediction accuracy.

\textbf{Reward Fine-tuning.} Finally, we demonstrate that \method\ serves as a good initialization for domain-specific fine-tuning. We fine-tune on RoboFAC~\citep{lu2025robofac}, a dataset of robotic failures and corrections spanning 16 tasks and 53 scenes (11k trajectories), including both simulated and real-world successes and failures. 
We adapt \method\ via LoRA~\citep{hu2022lora} adapters and full fine-tuning (FFT).
We compare against fine-tuning the base VLM \texttt{Qwen/Qwen3-VL-4B-Instruct} in the same way.

In addition to the previous VOC $r$ and Ranking Kendall $\tau_a$, we also compare a \emph{success - fail} metric measuring the difference in final reward between successful and failed trajectories of the same task.
Qwen3 fine-tuned still performs worse than \method\ zero-shot on 2 out of 3 metrics;
fine-tuning from \method\ yields substantially better reward evaluation results than training Qwen3-VL from scratch across all of our ranking metrics (\Cref{tab:robofac_finetune_results}). 
Importantly, LoRA and FFT perform similarly, demonstrating that \method\ can be effectively fine-tuned with just 1 GPU.
See further experiment details in \Cref{sec:appdx:robofac_lora_finetune}.

\begin{table}[t]
\vspace{6pt}
\centering
\small
\setlength{\tabcolsep}{8pt}
\begin{adjustbox}{max width=\linewidth}
\begin{tabular}{lccc}
\toprule
\textbf{Method} & \textbf{VOC} $r$ $\uparrow$ & \textbf{Kendall} $\tau$ $\uparrow$ & \textbf{Succ-Fail Diff} $\uparrow$ \\
\midrule
\robometercolor{\method-4B} (Zero-shot) & 0.652 & 0.436 & 0.141 \\
\midrule
Qwen3-VL (LoRA)    & 0.701 & 0.067 & 0.005 \\
Qwen3-VL  (Full FT)    & 0.727 & 0.102 & 0.008 \\
\robometercolor{\method-4B} (LoRA) & 0.875 & 0.786 & 0.271 \\
\robometercolor{\method-4B} (FFT) & \textbf{0.884} & \textbf{0.802} & \textbf{0.302} \\
\bottomrule
\end{tabular}
\end{adjustbox}
\vspace{2pt}
\caption{
\textbf{Finetuning \method\ on RoboFAC dataset.} Zero-shot performance is strong, and fine-tuning the base Qwen3 VLM performs worse on Kendall $\tau$ and success - fail difference compared to zero-shot \robometercolor{\method}. Meanwhile, \robometercolor{\method} fine-tuned performs best, either with LoRA or FFT.}
\label{tab:robofac_finetune_results}
\end{table}

Overall, our results demonstrate that \textbf{\method\ outperforms baselines} in both \textbf{generalization} and \textbf{distinguishing} successful / failed trajectories, and that it also serves as a \textbf{strong initialization for fine-tuning}.
We next analyze \emph{why}. 

\subsection{Ablations: Why does \method\ Perform so Well?}
\label{sec:exp:reward_model_ablations}
Here, we investigate individual components of \method\ to evaluate specific hypotheses about reward model training and its effects on downstream RL performance.
\begin{enumerate}[label=\textbf{H\arabic*}, nosep]
    \item \label{h1} Predicting \textbf{preferences} (\Cref{eq:preference}), even without paired failure trajectories, improves reward performance.
    \item \label{h2} Scaling preference prediction with additional \textbf{failure data} leads to improved reward model performance.
    \item \label{h3} Fine-tuning from \textbf{pre-trained VLMs} helps with reward predictions on unseen tasks.
\end{enumerate}

Our main analysis is performed via a controlled setting with data from the LIBERO~\citep{liu2023libero} robot manipulation simulated benchmark.
We train models with 1,709 successful demos from \texttt{LIBERO-\{10, Object, Goal, Spatial\}} and evaluate performance on a sample of the 8,262 unseen, paired, successful and failed trajectories from \texttt{LIBERO-90}.

\begin{table}[tb]
\vspace{6pt}
\centering
\small
\renewcommand{\arraystretch}{1.2}
\begin{adjustbox}{max width=\linewidth}
\begin{tabular}{l c c c c c c}
\toprule
& \multicolumn{3}{c}{\textbf{(a) LIBERO-90}} & \multicolumn{3}{c}{\textbf{(b) \evaldatasetood}} \\
\cmidrule(lr){2-4} \cmidrule(lr){5-7}
Ablation 
& \textbf{VOC $r$} 
& \textbf{Kendall $\tau$} 
& \textbf{Suc $-$ Fail}
& \textbf{VOC $r$} 
& \textbf{Kendall $\tau$} 
& \textbf{Suc $-$ Fail} \\
\midrule
\ref{h1} \textbf{Prog. Only} 
& 0.96 & 0.63 & 0.11 
& 0.93 & 0.31 & 0.08 \\

\ref{h1} \textbf{+Preference} 
& 0.90 & 0.74 & 0.22
& \textbf{0.95} & 0.54 & 0.24 \\
\midrule
\ref{h2} \textbf{+Failed Data} 
& \textbf{0.98} & \textbf{0.92} & \textbf{0.46}
& 0.94 & \textbf{0.64} & \textbf{0.32} \\

\ref{h3} \textbf{ReWiND Arch.} 
& 0.48 & -0.14 & -0.02
& 0.51 & 0.01 & 0.02 \\
\bottomrule
\end{tabular}
\end{adjustbox}
\caption{Reward alignment (VOC Pearson $r$), policy ranking (Kendall $\tau$), and average reward difference between successful and failed trajectories on LIBERO-90 and \evaldatasetood.}
\label{tab:exp:ablations}
\end{table}
\begin{figure}[tb]
   \centering
   \includegraphics[width=\linewidth]{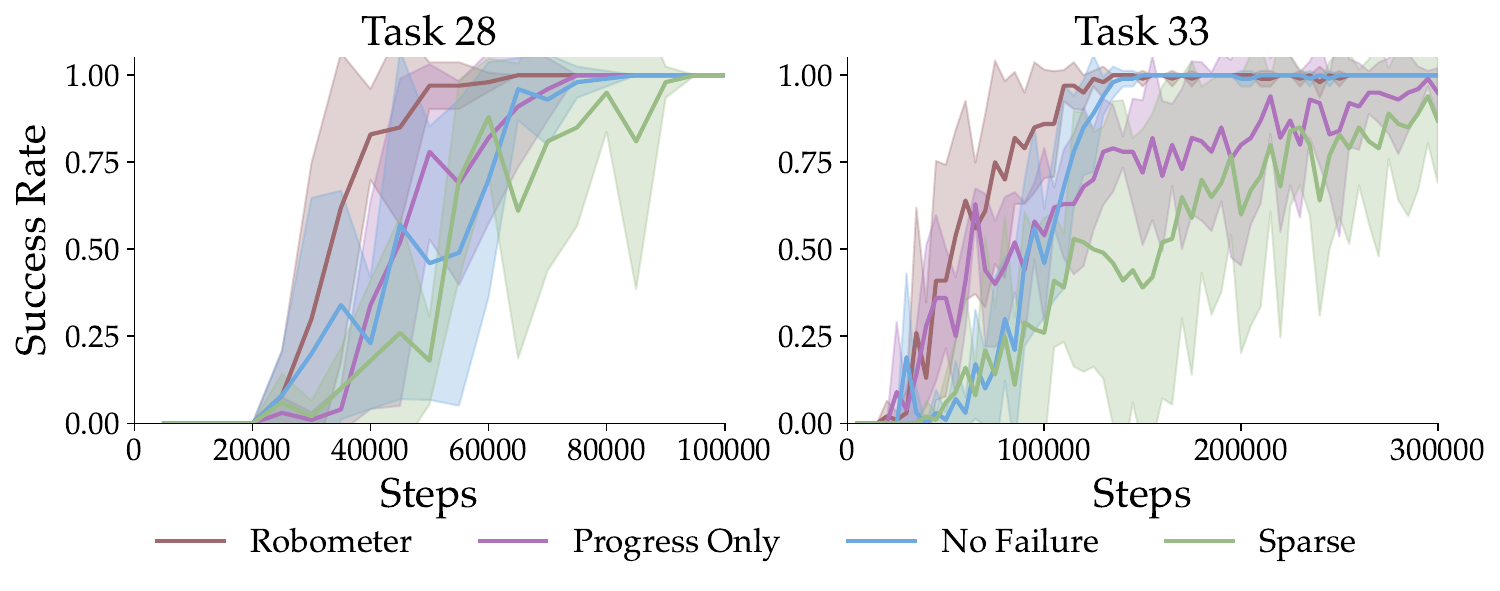}
   \caption{\textbf{RL w/ Ablation Models} in LIBERO-90 tasks from scratch, corresponding to ablations trained only on LIBERO-10/Object/Goal/Spatial data from \Cref{tab:exp:ablations}. We report the average success rate $\pm$ standard deviation across 5 seeds.}
   \label{fig:exp:rl_ablation}
\end{figure}
\textbf{Reward Model Ablations.}
To measure \ref{h1}, we train \method\ with \emph{only} progress prediction and also \method\ with both progress and preference prediction on the 1,709 demo dataset containing no failed trajectories.
We then measure \ref{h2}---scaling with failure data for preference training---by adding in 1,929 generated, failed LIBERO trajectories and train \method\ with the full \method\ training objective of progress and preference prediction on the larger dataset.
These LIBERO ablations are trained with LoRA~\citep{hu2022lora} due to the small dataset size.
Finally, we verify the importance of a pre-trained VLM (\ref{h3}) by training a larger, 500M-parameter version of ReWiND's transformer model (originally designed for low-data regimes) with our preference and progress objectives on the paired-failure LIBERO dataset.

We depict results on LIBERO, and separately, trained on the full \datasetname\ and evaluated on \evaldatasetood\ (with failed data removed for +Preference and +Failed Data), in \Cref{tab:exp:ablations}.
First, comparing \textbf{H1 Prog. Only} to \textbf{H1 +Preference}, adding preference supervision consistently improves policy ranking performance, increasing Kendall-$\tau_{a}$ on both LIBERO-90 and \evaldatasetood.
Second, \textbf{incorporating failed trajectories} for preference training (\textbf{H2 +Failed Data}) yields the \textbf{largest gains} across all ranking-based metrics. 
On LIBERO-90, Kendall-$\tau$ improves to 0.92 and the average difference in final rewards between success–failure increases more than 4$\times$ relative to progress-only training.
Similar trends hold on \evaldatasetood, where Kendall-$\tau$ and success–failure separation improve significantly.
Finally, replacing the pretrained VLM backbone with a scaled variant of ReWiND’s architecture (\textbf{H3 ReWiND Arch.}) results in a severe degradation across all metrics, confirming that large-scale multimodal pretrained backbone is essential for learning generalizable reward representations.
See \Cref{sec:app:ablations} for additional ablations.

\textbf{Ablation Results on RL Performance.}
Before moving to our full policy learning experiments, we verify that the reward evaluation trends observed in our ablation studies hold for policy learning. We train policies via online RL using ablated reward models on two tasks from the unseen LIBERO-90 suite. 
These tasks were selected because sparse-reward RL eventually achieves a near-100\% success rate, allowing us to directly compare \emph{sample efficiency} rather than success rates.

Results in \Cref{fig:exp:rl_ablation} demonstrate that improvements in reward evaluation metrics consistently transfer to policy success rates. For both tasks, policies trained using \method{} (the \ref{h2} LIBERO model) demonstrate better sample efficiency than ablations (\ref{h1} LIBERO models) and sparse reward, highlighting the importance of dense, well-calibrated reward signals for efficient policy optimization. 
Overall, \method{} trained only on LIBERO achieves \textbf{$\mathbf{2-4\times}$ better sample efficiency} than sparse reward on these unseen tasks.
Additional details of the RL training setup are provided in \Cref{sec:app:policy_learning}.

\begin{figure*}[tb]
   \centering
   \includegraphics[width=\linewidth]{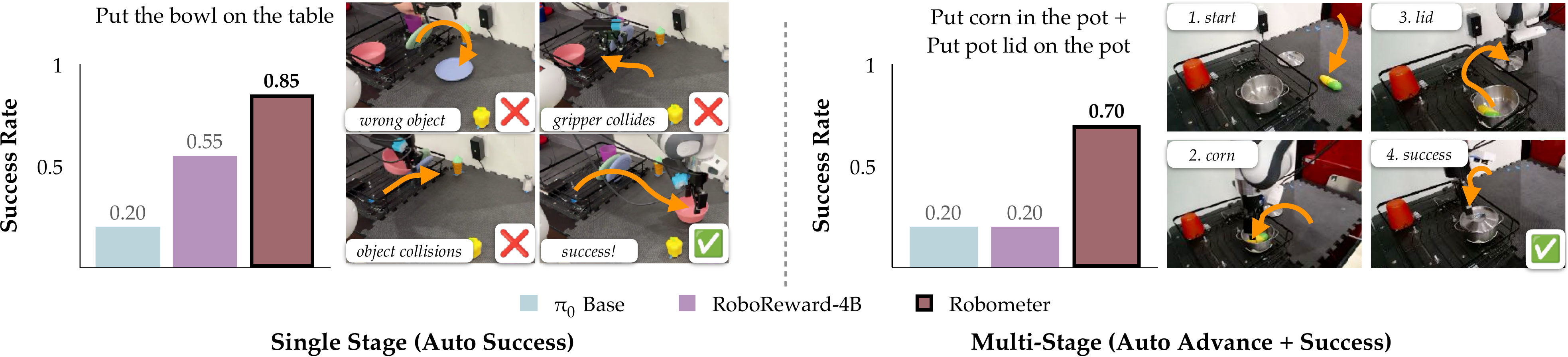}
   \caption{\textbf{Automatic online RL} with DSRL~\citep{wagenmaker2025steering} on a DROID~\citep{khazatsky2024droid} setup with \robometercolor{\method{}} improves $\pi_0$ from 20\% to 85\% on a single-stage task and 20\% to 70\% on a two-stage task, outperforming \roborewardcolor{RoboReward}'s overall success rate by $2.5\times$. DSRL with \robometercolor{\method{}} learns to avoid base $\pi_0$ errors such as collisions or moving the wrong object. The setup is deemed ``automatic'' because success detection and stage advancement are handled automatically by the reward model, requiring human intervention only for physical scene resets.}
   \label{fig:exp:dsrl_fig}
\end{figure*}

\subsection{Accelerating Robot Learning with Generalizable Rewards}
\label{sec:exp:marquee_exps}
We evaluate whether \method's dense and generalizable reward signals can be used \textbf{zero-shot} into improved downstream robot learning across four settings, including both prehensile and non-prehensile tasks: (1) automatic online RL, (2) offline RL with mixed-expertise data, (3) data filtering and retrieval for policy improvement, and (4) out-of-distribution failure detection. 
Across all experiments, we compare against RoboReward-4B---the strongest baseline reward model in our offline evaluations---to assess how \method's dense, instruction-aligned rewards affect learning stability, robustness, and sample efficiency. 
We also compare against strong, relevant, non-reward-model baselines for each setting where applicable.
All policy learning results are averaged over 20 evaluation trials.
Additional details and finer-grained results on each experiment can be found in \Cref{sec:app:policy_learning}.

\begin{figure}[t]
   \centering
   \includegraphics[width=\linewidth]{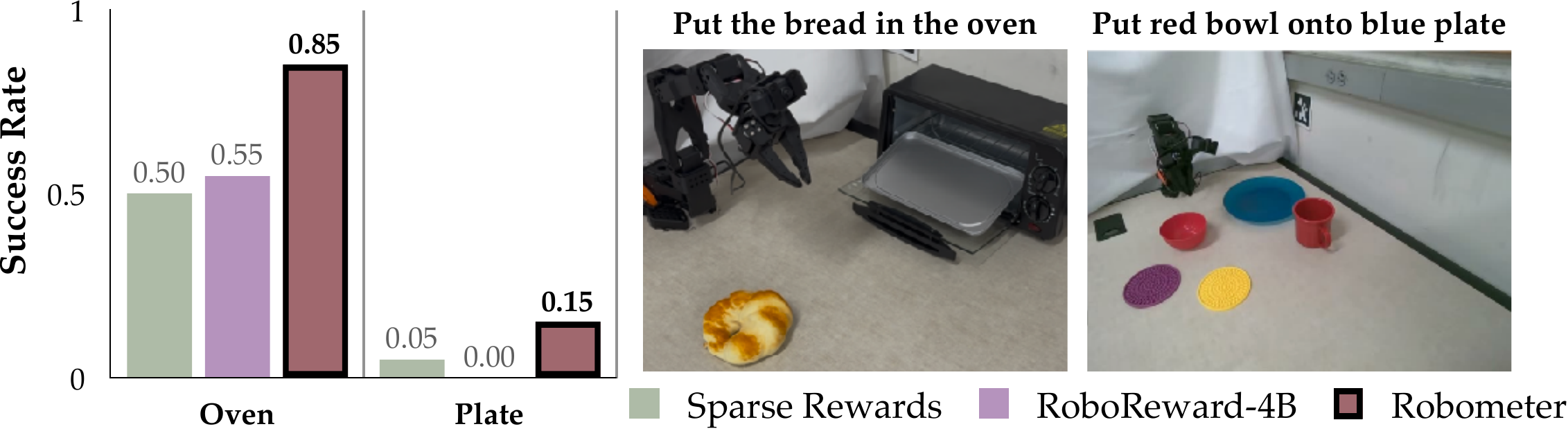}
   \caption{\textbf{Offline RL} results using IQL %
   on a mixture of Noisy and Expert trajectories. %
   \robometercolor{\method{}} rewards consistently outperform both \roborewardcolor{RoboReward} and sparse rewards: $2.4\times$ average success rate improvement over the best baseline for each task.
   }
   \label{fig:exp:offline_rl}
\end{figure}
\textbf{Automatic Online RL.}
First, we evaluate \method{} in an \emph{automated} online RL setting by training DSRL~\cite{wagenmaker2025steering} from scratch on a $\pi_0$ base policy~\citep{black2024pi0visionlanguageactionflowmodel} pre-trained on DROID~\citep{khazatsky2024droid}. \method{} enables autonomous RL by providing dense rewards and explicit \emph{success predictions}, which we use to automate episode termination; manual human intervention is required only for physical scene resets. For comparison, RoboReward’s discrete scores are also used for both reward shaping and success detection.
As shown in \Cref{fig:exp:dsrl_fig} (left), DSRL+\method\ improves success from $20\%$ to $85\%$ in \textbf{$\le 45$ minutes} (10k timesteps), outperforming RoboReward’s $55\%$. This gap arises from a key failure mode of RoboReward: it frequently assigns maximum rewards for unrelated tasks (e.g., picking up the wrong object), leading to premature resets and reinforcing incorrect behaviors. In contrast, \method{} provides a more reliable learning signal.

Next, we evaluate a \textbf{longer-horizon RL} setting in \Cref{fig:exp:dsrl_fig} (right), where \method{}'s success predictions trigger progression between stages of a multi-stage task. Unlike methods that explicitly train with multi-stage rewards and thus require stage labels~\citep{kim2025reds, chen2025sarm}, we simply decompose tasks into stages at inference time using a pre-trained VLM and use \method{} to advance stages automatically. In this setting, DSRL+\method\ improves $\pi_0$'s success from $20\%$ to $70\%$ over 10k timesteps, outperforming RoboReward’s 20\%, which suffers from inaccurate rewards and unreliable stage transitions.
Across both setups, \method\ outperforms RoboReward's overall success rate by an average of $\mathbf{2.5\times}$.

Finally, we perform an additional online RL experiment---\emph{model-based RL} integrating \method\ into DreamZero~\citep{ye2026dreamzero}---where \method\ improves success rate from 20\% to 70\%. See details and results in \Cref{sec:appdx:modelbasedrl}.

\textbf{Combining Noisy and Expert Data via Offline RL.}
We consider an offline RL setting with mixed-expertise data for two tasks on an SO-101 robot (SO-101 is not in \datasetname), combining expert and noisy, suboptimal demos, as shown in \Cref{fig:exp:offline_rl}.
We train policies with Implicit Q-Learning (IQL)~\citep{kostrikov2021offline} to study how dense rewards from \method\ improve learning stability and policy extraction in offline RL.

Accurate, dense reward signals can provide informative intermediate feedback, reducing reliance on long-horizon credit assignment and enabling trajectory ``stitching'' with smaller discount factors $\gamma$, thereby reducing value function variance. 
For each of sparse reward, RoboReward, and \method{}, we sweep $\gamma \in {0.90, 0.95, 0.99}$ and report the best-performing checkpoint over 30,000 offline training steps.
We observe that \method{}, which provides dense, temporally aligned rewards, performs best at a lower discount factor $\gamma=0.9$ and outperforms both baselines across both tasks with a $\mathbf{2.4\times}$ success rate improvement over the best baseline in each.
RoboReward performs similarly to sparse rewards across the $\gamma$ sweep, as its categorical (1–5) outputs provide less dense guidance and often assign high rewards to suboptimal trajectories, leading to noisy training signals.

\textbf{Data Filtering \& Retrieval.}
We next evaluate \method{} as a mechanism for unsupervised data filtering and retrieval. Using a bimanual ``play'' dataset~\citep{lynch2019play} of unannotated, multi-task trajectories collected on a Trossen AI setup (not in \datasetname), we retrieve the top 100 subtrajectories for a given task instruction. We compare retrieval relevance against RoboReward, pre-trained SigLIP~\citep{zhai2023siglip}, and a retrieval-specific baseline, STRAP~\citep{memmel2025strap}.
For \method\ we retrieve subtrajectories using (i) the preference objective via pairwise trajectory comparisons, or (ii) the progress objective by computing each trajectory’s value–order correlation. 
As shown in \Cref{fig:exp:retrieval_rate}(a), \method{} consistently achieves higher retrieval relevance across five tasks. 
Finally, we LoRA-finetune $\pi_{0.5}$~\citep{hu2022lora, liu2024tail, intelligence2025pi05} on these retrieved segments.
Policies trained on \method{}-filtered data vastly outperform those using baseline-retrieved data on \texttt{Stir the Pot} and \texttt{Open the Red Drawer} (\Cref{fig:exp:retrieval_rate}(b)), demonstrating its efficacy for targeted imitation learning. 
Low baseline success rates despite high retrieval rates stem from their retrieval of more failed and suboptimal, yet task-relevant, subtrajectories.
Overall, \method-retrieval averages a $\mathbf{4.5\times}$ higher success rate than the best baseline.

\
\begin{figure*}[!ht]
   \centering
   \includegraphics[width=\linewidth]{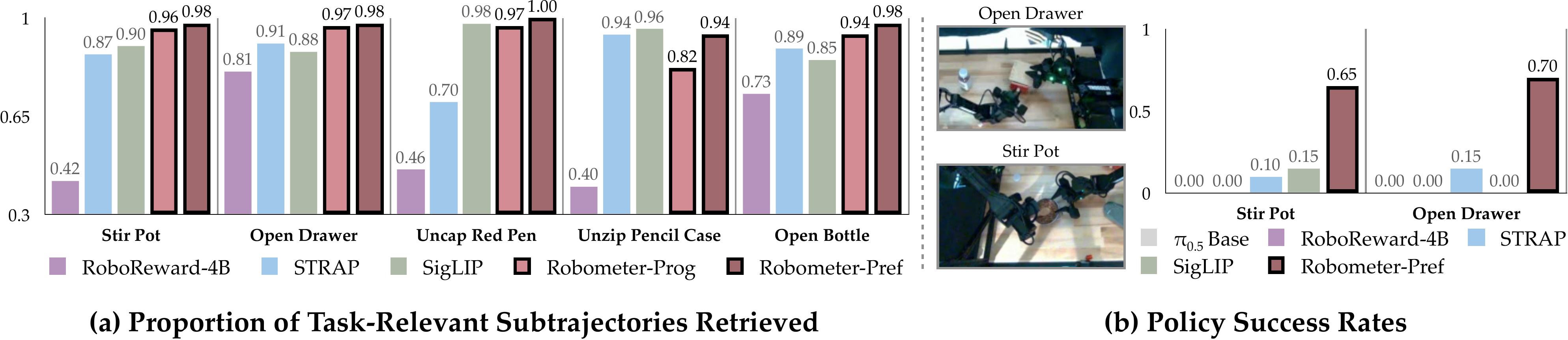}
   \caption{\textbf{(a): Proportion of task-relevant subtrajectories} out of 100 retrieval queries. Our method consistently retrieves a high number of relevant subtrajectories using either the preference or progress objective.
   \textbf{(b): Success rates} of LoRA-finetuned $\pi_{0.5}$ policies using the retrieved trajectories from each method. Small amounts of suboptimal \& unrelated data retrieved by other baselines degrade policy-learning performance: \robometercolor{\method}-retrieval attains an average $4.5\times$ success rate improvement over the best baseline.
   }
   \label{fig:exp:retrieval_rate}
\end{figure*}

\begin{table}[!bt]
\vspace{6pt}
\centering
\small
\begin{adjustbox}{max width=\linewidth}
\begin{tabular}{l ccccc}
\toprule
\textbf{Task} 
& \textbf{T.U.} 
& \textbf{\vlaccolor{VLAC}} 
& \textbf{GPT-5-mini} 
& \textbf{\roborewardcolor{RoboReward}-4B} 
& \textbf{\robometercolor{\method}} \\
\midrule
move banana  & 0.53 & 0.45 & 0.48 & 0.91 & \textbf{0.94} \\
move mouse   & 0.50 & 0.00 & 0.89 & 0.80 & \textbf{0.91} \\
pour pebble  & 0.32 & 0.00 & 0.25 & 0.73 & \textbf{0.83} \\
fold towel   & \textbf{0.58} & 0.16 & 0.27 & 0.40 & \textbf{0.58} \\
pull tissue  & 0.43 & 0.00 & 0.00 & 0.57 & \textbf{0.76} \\
put spoon    & 0.22 & 0.00 & 0.25 & \textbf{0.73} & \textbf{0.73} \\
stir pot     & 0.47 & 0.00 & 0.17 & \textbf{0.95} & 0.90 \\
\midrule
\textbf{Average} 
& 0.48 & 0.16 & 0.33 & 0.74 & \textbf{0.81} \\
\bottomrule
\end{tabular}
\end{adjustbox}
\caption{
\textbf{Failure detection performance}. \robometercolor{\method} achieves the highest average F1 score. T.U. is the token-uncertainty baseline.
}
\label{tab:ood_failure_detection}
\end{table}

\textbf{Failure Detection.}
Detecting failures during online deployment is critical for safe deployment. Thus, we evaluate \method{}’s zero-shot failure detection on 100 manipulation trajectories from a Franka Panda DROID robot (30 successful, 70 failed) spanning 7 tasks collected in scenes not in \datasetname. Failures are evenly split between irreversible failures (e.g., drops or spills) and insufficient-progress failures, where execution stalls or terminates prematurely. We compare our method against: the token-uncertainty~\citep{gu2025safe} of $\pi_0$-FAST-DROID~\citep{pertsch2025fast} as proposed by ~\citet{gu2025safe} for zero-shot failure detection; VLAC, which reports failure detection results; GPT-5-mini; and RoboReward-4B.
Failures are detected via temporal inconsistencies in predicted per-frame rewards. 

As shown in Table~\ref{tab:ood_failure_detection}, \method{} achieves the highest average F1 score, effectively balancing true positive and true negative rates (TPR and TNR); the full breakdown with TPR and TNR is provided in Appendix Table~\ref{tab:ood_failure_detection_full}.
VLAC frequently flags trajectories as failures, achieving high TPR but low TNR, resulting in lower F1 scores.
RoboReward-4B performs competitively but underperforms \method{}, particularly on tasks with subtle failure modes such as \emph{fold towel} and \emph{pull tissue}. 
As detailed and visualized in \Cref{sec:app:policy_learning:ood-failure-detection}, \method{} robustly detects irreversible, insufficient-progress, and non-terminal failures (e.g., hovering, oscillation, or partial completion), fully zero-shot across tasks and environments—unlike prior methods that require task-specific thresholds, calibration, or test-time interaction~\citep{gu2025safe, xu2025detect, agia2025unpacking}.
\Cref{fig:failure_insufficient} shows two examples of insufficient-progress failures.

\begin{figure*}[t]
    \centering
    \captionsetup{font=small}

    \begin{subfigure}[t]{0.49\textwidth}
        \centering
        \includegraphics[width=\linewidth]{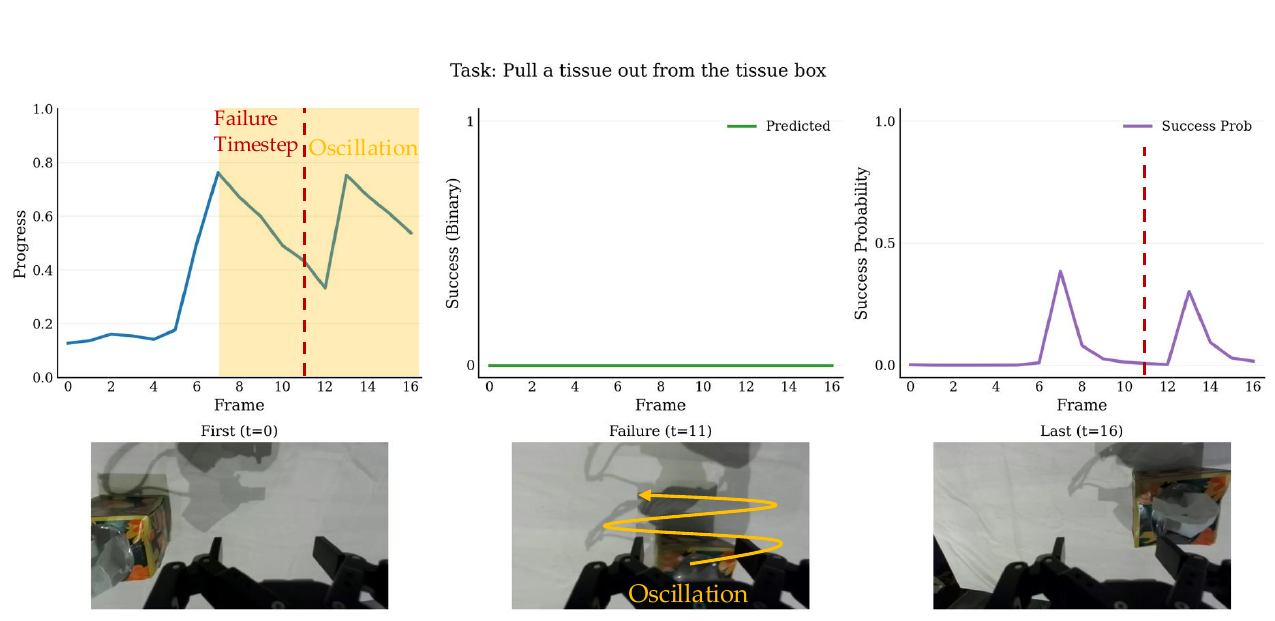}
        \caption{Oscillatory behavior without task completion.}
    \end{subfigure}
    \hfill
    \begin{subfigure}[t]{0.49\textwidth}
        \centering
        \includegraphics[width=\linewidth]{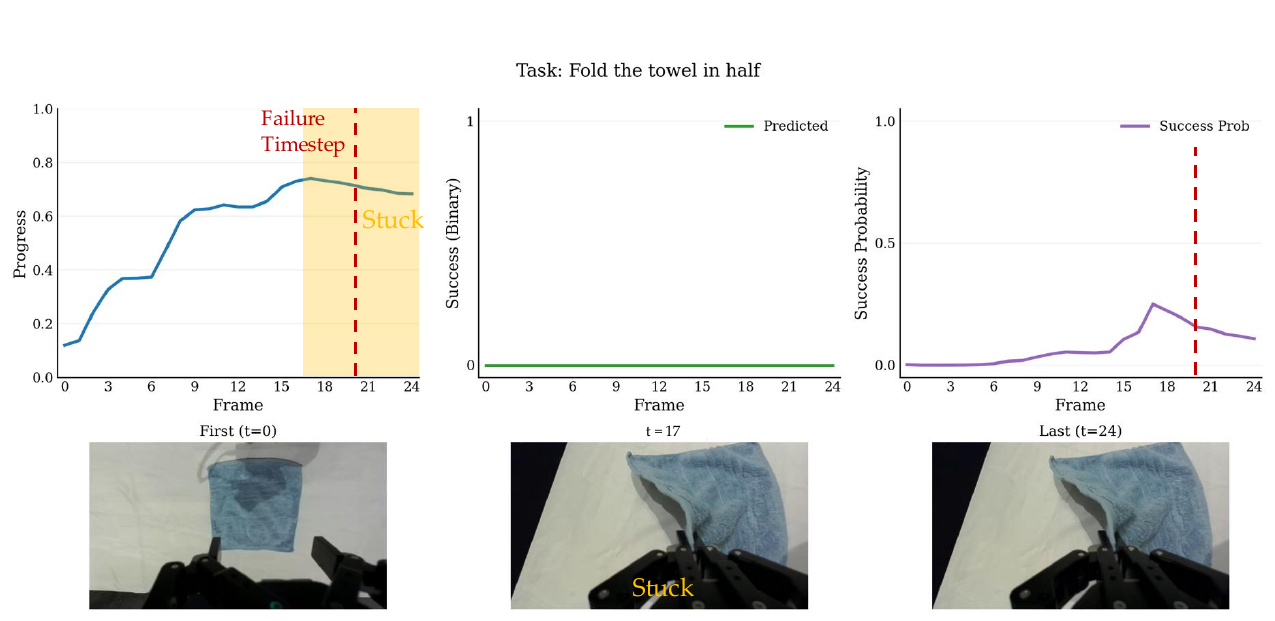}
        \caption{Execution stalls with plateaued progress.}
    \end{subfigure}

    \caption{\textbf{Insufficient-progress failures.}
    Although the robot continues to move, predicted progress stagnates or oscillates without converging to success, enabling detection of failures caused by stalling or premature termination.}
    \label{fig:failure_insufficient}
\end{figure*}

\endgroup

\section{Limitations and Future Work}

\method{} operates as a dense, per-frame reward model over subsampled video inputs, enabling scalable training but limiting its ability to memorize specific past key events for very long-horizon or partially observed tasks.
In addition, real-world robot executions exhibit a wide diversity of failure modes, many of which are rare, subtle, or task-specific, and the current training data may not fully capture this breadth. 
As a result, \method{} may fail to recognize or correctly reward certain failure cases that fall outside the dominant patterns seen during training. 
We hope that our training recipe enables future work to build upon larger reward modeling datasets.

As a vision-language-based model, \method{}  also lacks direct access to latent physical state such as contact forces, grasp stability, or compliance, and may fail to recognize or correctly reward failure cases driven by these factors until they become visually observable.
Future work could address these limitations by incorporating denser temporal modeling, VQA-style supervision to reason about task structure and completion criteria, and off-domain data for better generalization~\citep{hamster2025, zhang2025peek}, as well as by developing more systematically curated failure datasets that better reflect the diversity of real-world failure modes~\citep{tian2026position}.
Finally, real-world executions may also contain safety-critical states (e.g., in a constrained MDP formulation) that \method\ is not pre-trained to predict. Future work can train \method\ with an additional safety head, similar to the success prediction head, on provided safety labels.

%% file: sections/appendix/appendix_main.tex
\appendices

\hypersetup{linkcolor=customPurple}
\section*{Appendix Table of Contents}

\startcontents[appendix]

\printcontents[appendix]{l}{1}{\setcounter{tocdepth}{2}}
\vspace{1cm}

\hypersetup{linkcolor=customGreen}

\input{sections/appendix/data}
\input{sections/appendix/model}

\input{sections/appendix/addtl_reward_exps}
\input{sections/appendix/addtl_ablations}
\input{sections/appendix/policy_learning_exps}

%% file: sections/appendix/data.tex
\section{Dataset Details}
\label{sec:app:data}

\begin{table}[htb]
\centering
\small
\begin{adjustbox}{max width=\linewidth}
\begin{tabular}{l r}
\toprule
\textbf{Paper} & \textbf{\# Trajectories} \\
\midrule
RoboReward~\citep{lee2026roboreward} & 45k \\
RoboDopamine~\citep{tan2025robodopamine} & 100k \\
VLAC~\citep{zhai2025VLAC} & 300k \\
\datasetname\ (ours) & 1M \\
\bottomrule
\end{tabular}
\end{adjustbox}
\caption{Comparison of known approximate trajectory counts across recent general-purpose reward modeling papers.}
\label{tab:traj_comparison}
\end{table}

\begin{figure}[htb]
    \centering
    \includegraphics[width=0.7\linewidth]{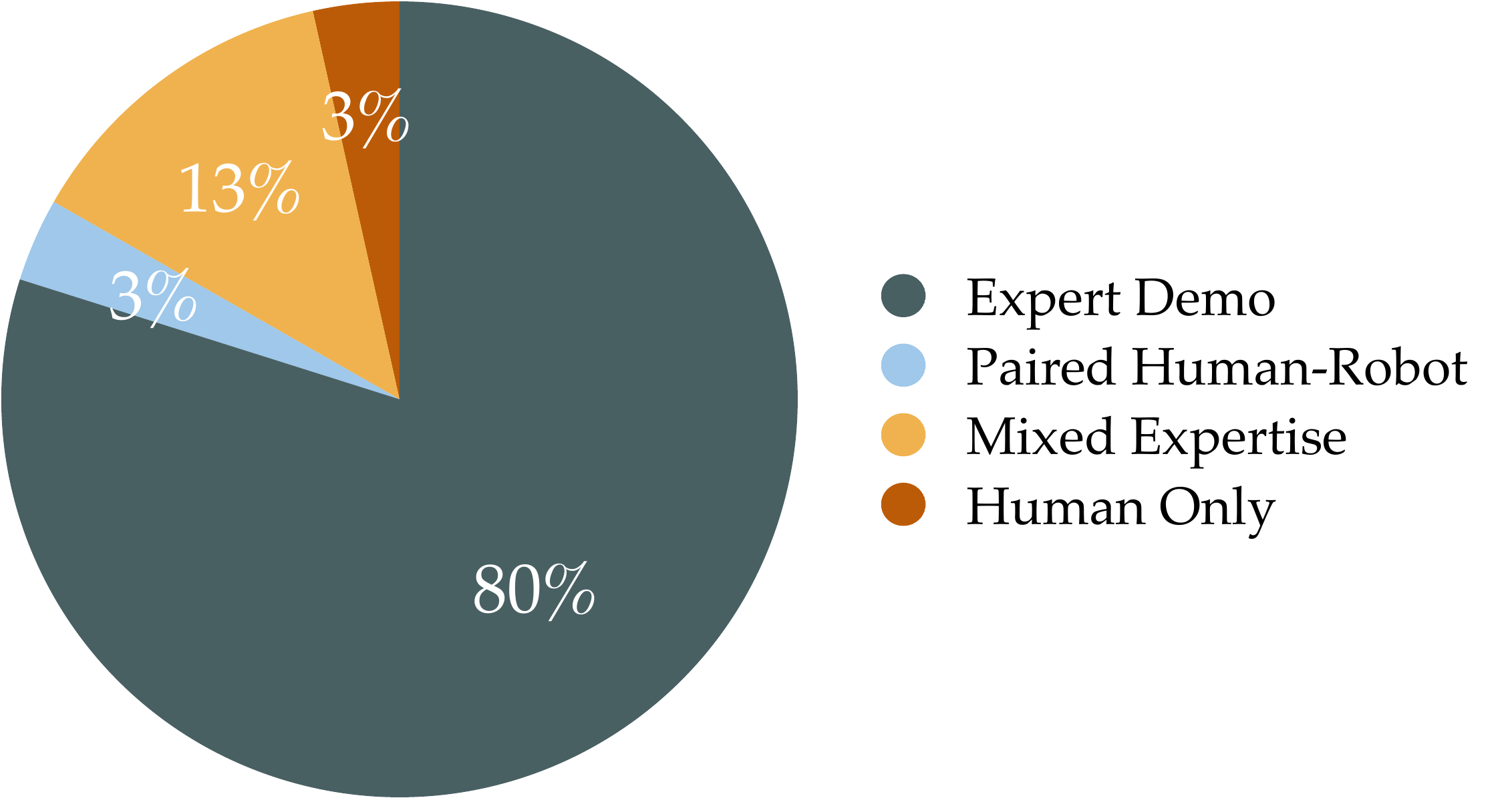}
    \caption{Pie chart of \datasetname\ dataset types. Full table with individual dataset details in \Cref{tab:appdx:training_datasets}.}
    \label{fig:appdx:pie_chart}
\end{figure}

\begin{table*}[t]
  \centering
  \begin{tabular}{l l r l}
  \toprule
  \textbf{Dataset} & \textbf{Embodiment} & \textbf{\# Trajectories} & \textbf{Citation} \\
  \midrule
  OXE Eval Suite & Franka, Google Robot, Jaco 2, WidowX & 14,398 & \cite{open_x_embodiment_rt_x_2023} \\
  RACER (Eval) & Franka Panda (simulation) & 7,227 & \cite{dai2025racer} \\
  Metaworld Eval & Sawyer & 151 & \cite{yu2021metaworldbenchmarkevaluationmultitask} \\
  LIBERO Eval & Franka Panda (simulation) & 3,950 & -- \\
  \midrule
  \textbf{Total} & 5 unique robot embodiments & 25,726 & -- \\
  \bottomrule
  \end{tabular}
\caption{\textbf{\evaldatasetid} - In distribution evaluation datasets overview. \# Trajectories is computed by aggregating all evaluation splits per dataset, counting unique annotated trajectories.}
\label{tab:appdx:evaluation_datasets_id}
\end{table*}

\begin{table*}[t]
\centering
\begin{tabular}{l l r l}
\toprule
\textbf{Dataset} & \textbf{Embodiment} & \textbf{\# Trajectories} \\
\midrule
\usc\ Franka & Franka Panda & 24 \\
\usc\ Koch & Bimanual Kochv1.1 & 150 \\
\usc\ Trossen & Trossen & 27 \\
\usc\ xArm & xArm & 36 \\
\mituniv\ Franka & Franka Panda & 304 \\
\utd\ SO101 & SO101 & 30 \\
\midrule

\textbf{Total} & 5 unique robot embodiments & 571 \\
\bottomrule
\end{tabular}

\caption{\textbf{\evaldatasetood} - Out-of-distribution evaluation datasets overview. \# Trajectories is computed by aggregating all evaluation splits per dataset, counting unique annotated trajectories. }
\label{tab:appdx:evaluation_datasets_ood}
\end{table*}

\begin{table*}[t]
\centering
\begin{tabular}{l l r l}
\toprule
\textbf{Dataset} & \textbf{Embodiment} & \textbf{\# Trajectories} & \textbf{Citation} \\
\midrule
\multicolumn{4}{l}{\textbf{Expert Demonstration Datasets}} \\
OXE Mix & 11 mixed embodiments & 449,475 & \cite{open_x_embodiment_rt_x_2023} \\
AGIBotWorld-Alpha Subset & AgiBot G1 Bimanual Mobile Manipulator & 216,911 & \cite{contributors2024agibotworldrepo} \\
Galaxea Open World & Galaxea R1 Lite Bimanual Mobile Manipulator & 108,118 & \cite{jiang2025galaxea} \\
RoboReward OXE Mix & 13 mixed embodiments & 45,072 & \cite{lee2026roboreward} \\
MolmoACT & Franka Panda & 15,546 & \cite{lee2025molmoact} \\
Humanoid Everyday & Unitree G1 \& H1 Bimanual Humanoids & 9,208 & \cite{zhao2025humanoid} \\
LIBERO & Franka Panda (simulation) & 1,709 & \cite{liu2023libero} \\
MetaWorld (ReWiND) & Sawyer (simulation) & 100 & \cite{yu2021metaworldbenchmarkevaluationmultitask, zhang2025rewind} \\

\midrule
\multicolumn{4}{l}{\textbf{Paired Human–Robot Datasets}} \\
MotIF & Stretch Mobile Manipulator & 83 & \cite{hwang2025motif} \\
RH-20T & 4 embodiments & 29,969 & \cite{fang2023rh20t} \\
PH2D & H1 (partly in MuJoCo simulation) & 3,596 & \cite{qiu2025humanoid} \\
H2R & XArm & 2,254 & \cite{xie2025human2robot} \\

\midrule
\multicolumn{4}{l}{\textbf{Mixed Expertise Datasets}} \\
RoboArena Pairwise Comparisons & Franka Panda & 12,379 & \cite{atreya2025roboarena} \\
SOAR Paired Success/Fail & WidowX & 16,812 & \cite{zhou2024autonomous} \\
FAILSafe & Franka Panda (ManiSkill~\cite{taomaniskill3} simulation) & 71,614 & \cite{lin2025failsafe} \\
RACER Failure Dataset & Franka Panda (RLBench~\cite{james2020rlbench} simulation) & 29,115 & \cite{dai2025racer} \\
AutoEval Failed Trajectories & WidowX250 & 8,677 & \cite{zhou2025autoeval} \\
LIBERO Failure Dataset & Franka Panda (simulation) & 1,473 & \cite{liu2023libero} \\
Fino-Net Paired Failure (Baxter) & Baxter & 229 & \cite{inceoglu2020fino} \\

\midrule
\multicolumn{4}{l}{\textbf{Human-Only Datasets}} \\
Epic-Kitchens & Human & 37,030 & \cite{Damen2022EpicKitchens} \\

\midrule
\textbf{Total} & 21 unique robot embodiments  & 1,059,370 & -- \\
\bottomrule
\end{tabular}
\caption{\textbf{\datasetname} Training datasets overview. \# Trajectories is determined by counting every unique video-language annotation (possibly across multiple views when available). }
\label{tab:appdx:training_datasets}
\end{table*}

\subsection{Individual \datasetname\ Training Dataset Details}
\label{sec:app:data:datasets}
Here, we list dataset curation details for every source in \datasetname. 

\textbf{Preference Only Data.} Some data sources vary significantly in when the teleoperator marks the end of the trajectory.
As such, using these data sources for progress prediction and end-state success prediction negatively affects prediction accuracy.
To avoid this issue, progress-only generalist robotic reward models would have to either manually label end states or simply forgo using these data sources.
However, due to \method's trajectory comparison-based preference prediction objective, we can still use these noisier datasets for preference prediction; we list which sources are used only for preference in each subsection.

\textbf{Image Resolution and Frame Downsampling.}
We first perform data pre-processing before using the data for training.
For storage efficiency, we downsample all trajectories to a maximum of 32 frames. 
Furthermore, we downsize image resolution such that the shortest edge (either height or width) of each image is 240 pixels.
This procedure allows us to maintain the aspect ratio while ensuring the dataset does not consume excessive storage space.
In total, the dataset size is around 6 TB.

\textbf{Task End Thresholds.} For predicting both progress and success (\Cref{sec:objective}), it's important that the end-frame of the trajectory corresponds to the timestep at which the task is actually finished.
However, in most real-world datasets, this is not the case due to human teleoperator delay or different notions of when tasks are finished.
We address this issue by manually setting this ``task finished'' threshold for each data source, which takes about 2 minutes per data source. 
See \Cref{sec:app:data:filtering} for further details.

\subsubsection{Expert Demonstration Datasets}
\label{sec:app:data:expert_demo}
We start with expert demonstration datasets, i.e., datasets that contain only successful trajectories demonstrated by competent teleoperators.
For all of these datasets, the progress target we use in \Cref{sec:objective} is $1.0$.

\paragraph{Open-X Embodiment Mix}
We select a subset of datasets from the Open-X Embodiment (OXE) Dataset~\citep{open_x_embodiment_rt_x_2023}. 
This subset was chosen from those selected for reward learning from prior work~\citep{zhang2025rewind}. 
The subset includes DROID~\citep{khazatsky2024droid}, Fractal~\citep{brohan2022rt1}, BRIDGE-v2~\citep{walke2023bridgedata}, Language Table~\citep{lynch2023interactive}, BC-Z~\citep{jang2021bcz}, RoboSet~\citep{bharadhwaj2023roboagent}, FurnitureBench~\citep{heo2023furniturebench}, UT Austin Mutex~\citep{shah2023mutex}, Berkeley Cable Routing~\citep{luo2023multistage}, CLVR Jaco Play~\citep{dass2023jacoplay}, Berkeley RPT~\citep{Radosavovic2023}, Toto~\citep{zhou2023train}, CMU Franka Pick-Insert~\citep{saxena2023multiresolution}, Stanford Hydra~\citep{belkhale2023hydra}, Berkeley MVP~\citep{Radosavovic2022}, Berkeley Fanuc Manipulation~\citep{fanuc_manipulation2023}, Mobile ALOHA~\citep{fu2024mobile}, Imperial College Sawyer Wrist Cam, UCSD Kitchen~\citep{ucsd_kitchens}, Austin BUDS~\citep{austinbuds}, DLR Edan~\cite{vogel_edan_2020, quere_shared_2020}, UTokyo LSMO~\citep{OsaLSMO2022}, and NYU Rot~\citep{haldar2023watch}.

We use wrist camera and external camera viewpoints from DROID due to the wide-angle of the DROID wrist cameras. For most other datasets, we only use external cameras unless wrist is the only viewpoint available.

Some of these datasets have validation or test-set splits; we use these splits for \evaldatasetid.

For BC-Z and DLR Edan, we use these datasets only for preference prediction due to highly varied trajectory termination times relative to when the task was actually completed.

\paragraph{AGIBot World}
The AgiBotWorld-Alpha dataset~\cite{contributors2024agibotworldrepo} consists of 100k+ long-horizon trajectories on the AgiBot G1 bimanual mobile manipulator.
We randomly select a 34,098 trajectory subset of the 100k trajectories to form our dataset.
Each long-horizon trajectory is annotated with each shorter-horizon subskill that is required to accomplish the task. In total, this leads 216,911 long-horizon and short-horizon trajectories. 
These short and long horizon skills include higher-level tasks such as ``put all the oranges in the basket" and the lower-level skills for each ``pick orange" and ``place orange."
Although the dataset includes wrist cameras and multiple fisheye camera angles, we only include the egocentric head camera in our dataset.

\paragraph{Galaxea Open-World}
The Galaxea Open-World Dataset~\cite{jiang2025galaxea} is a large-scale humanoid dataset with 108,118 trajectories across 150 task categories. The dataset includes diverse tasks from pick-and-place to whole-body manipulation on a Galaxea R1 Lite bimanual mobile manipulator. Due to highly varied trajectory termination times relative to when the task was actually completed, we use this dataset only for preference prediction.

\paragraph{RoboReward OXE + Roboarena Mix} In addition to our own OXE mix, we incorporate the 45k trajectory OXE + Roboarena~\citep{atreya2025roboarena} training subset from RoboReward. 
This dataset is labeled using RoboReward's VLM-based counterfactual instruction labeling technique to generate pseudo-failure instructions for successful trajectories.
This technique complements \method's training objectives — by directly incorporating all data into \method's training mix and objectives.
Final rewards for each trajectory are discrete numbers ranging from 1 to 5. 
We normalize their rewards to the range of $[0, 1]$, making this data suitable for all of our prediction objectives.
We refer readers to \citet{lee2026roboreward} for further details.
Similar to the OXE dataset mix, the DLR Edan and BC-Z subsets of the RoboReward data is also only used for preference prediction due to highly varied trajectory termination times relative to when the task was actually completed.

\paragraph{MolmoACT} We use all external-view data from MolmoACT~\citep{lee2025molmoact}, collected on a Franka Panda arm, excluding trajectories with corrupted videos. 
We do not use wrist-cam data from MolmoACT because the camera angle is too narrow and often does not show the object being manipulated.
We selected this dataset because it includes a diverse set of trajectories collected in clutter with good camera visibility.

\paragraph{Humanoid Everyday} We use all data from ~\citep{zhao2025humanoid}, which contains Unitree G1 \& H1 bimanual humanoid data from an egocentric viewpoint. This data was selected for diversity as it includes bimanual mobile data.

\paragraph{LIBERO} %
LIBERO~\citep{liu2023libero} provides a diverse set of simulated household manipulation tasks across 5 task suites. 
We use \texttt{LIBERO-\{10, Object, Spatial, Goal\}} for \datasetname\ and \texttt{LIBERO-90} for evaluation in \evaldatasetid.
We follow OpenVLA's~\citep{kim2024openvla} dataset re-generation scheme by re-rendering at 256x256, removing no-ops, and removing demonstrations which are not successful upon replay.
We selected LIBERO for its use as a popular VLA benchmark and for the ease with which we can generate our own failed trajectories across a diverse set of tasks.
Thus, we also include a corresponding set of failed trajectories for all 5 task suites, constructed by replaying demonstration trajectories with added Gaussian noise to each action, mimicking policy execution error that results in failure.

\paragraph{Metaworld} MetaWorld~\citep{yu2021metaworldbenchmarkevaluationmultitask} is a multi-task simulated manipulation benchmark with a Sawyer arm.
We use the 20-task training split consisting of 5 demonstrations each from \citet{zhang2025rewind} for \datasetname.
Correspondingly, we use the 17-task evaluation dataset from \citet{zhang2025rewind} for \evaldatasetid.
MetaWorld was selected early on for basic testing and ensuring that we can reproduce the results from ReWiND~\citep{zhang2025rewind} in our own implementation of it. 
We kept it in \datasetname\ for visual feature diversity.

\subsubsection{Paired Human-Robot Datasets}

\paragraph{MotIF} A human-robot paired dataset with a Stretch Mobile Manipulator containing tasks involving motion-counting such as ``stir 3 times'' or shaking boba tea~\citep{hwang2025motif}. This dataset was selected for these counting tasks to encourage learning to track repetitive motions.
\paragraph{RH-20T} A paired human-robot dataset for tabletop manipulation spanning four robot embodiments including Flexiv, UR5, Franka Panda, and Kuka robots~\citep{fang2023rh20t}. This dataset was selected for its diversity in tasks and embodiments. The RH20T dataset consists of 7 configurations each with its own robot, table and camera setup. For each configuration, we select 1--2 camera views which both 1) capture the full scene and robot motion and 2) are consistent with the language instruction in terms of spatial relationships, e.g., left, right, top, bottom, etc. We removed null robot demonstrations without any arm movements, as well as demonstrations which seem to be the concatenation of demonstrations for multiple tasks. 
\paragraph{PH2D} A human-robot, real and simulation dataset containing Unitree H1 trajectories collected from an egocentric viewpoint~\citep{qiu2025humanoid}. This dataset was selected because it pairs simulation, real, and human trajectories; it contains many pouring tasks and many very detailed task descriptions.

\paragraph{H2R} A human-robot paired dataset with a UFACTORY XARM robot~\citep{xie2025human2robot}. It contains pick-and-place and pushing tasks. We selected it because many tasks have multiple objects in the scene (e.g., a lighter tray and a darker tray to place a cube on), helping the reward model learn to better distinguish \emph{correctly} manipulated objects.

\subsubsection{Mixed Expertise Datasets}
The data here contains paired successful and failed trajectories. We incorporate these mixed expertise data to encourage \method\ to effectively reward failed trajectories, which can help in a variety of domains (e.g., all downstream experiments in \Cref{sec:exp:marquee_exps}).
\paragraph{RoboArena} Roboarena~\citep{atreya2025roboarena} data is from a set of human-performed evaluations of various generalist VLA policies on the DROID setup with a Franka Panda arm. Each evaluation has a partial progress score $\in [0, 1]$ with the vast majority of evaluations being failures. 
We use these progress scores solely to construct trajectory comparisons to predict over as trajectory termination times are highly varied and can essentially undo progress made in the middle of the trajectory, for which the human gave a partial progress score. We save videos from all available camera viewpoints, including wrist camera.

\paragraph{SOAR} SOAR~\citep{zhou2024autonomous} data comes from autonomous policy rollouts guided by a VLM on a WidowX250 robot. Success/fail labels, generated by a VLM, are also provided.
Due to automatic task generation and success/fail labeling, the dataset labels are quite noisy, and many trajectories contain tasks that are not possible in the scene, tasks that have already started from the first frame, and incorrect success/fail classifications.
Therefore, we use a pre-trained Qwen-3-VL-4B~\citep{Qwen3-VL} to filter out incorrectly labeled samples, including ones that were infeasible or unrelated to the task description. We do this by using the first, middle, and last frame to establish a general flow of the trajectory and ask the model to critique the positioning of relevant items. This is separated into a stage prompt that empirically improved filtering quality using a small, manually verified set and ultimately filters out 45\% of trajectories from the original dataset.

We save all filtered successful trajectories, and we also save all failed trajectories that have the same language description as at least one successful trajectory.
Because this data does not contain progress labels and because trajectory end frames are highly variable with respect to when the task was actually completed in successful trajectories, we use the entire dataset only for generating trajectory comparisons for preference prediction.

\paragraph{FAILSafe} FAILSafe~\citep{lin2025failsafe} contains successful and failed trajectories from a Franka Panda collected in the Maniskill simulator~\citep{taomaniskill3}. Each task has many example failures collected from both wrist and external cameras. Tasks are also segmented into sub-tasks, e.g., reaching a cube $\rightarrow$ grasping the cube $\rightarrow$ ...

\paragraph{RACER} RACER contains paired failed and successful trajectories on a Franka Panda in the RLBench simulator~\citep{james2020rlbench, dai2025racer}. This dataset was picked for its non-prehensile tasks, such as opening/closing drawers and sweeping.

\paragraph{AutoEval} AutoEval contains data from automatic policy evaluations collected on 2 different WidowX250 setups~\citep{zhou2025autoeval}. While task diversity is limited, we used this dataset because it contains diverse strategies coming from the evaluation of arbitrary policies.

This dataset is used only for preference prediction due to some noisy automatic success/fail detection.

\paragraph{LIBERO}
We self-generated a failure dataset in LIBERO~\citep{liu2023libero} by adding Gaussian noise to successful demonstration trajectories to re-generate paired failure trajectories for every task.
This data is paired with the original success-only LIBERO dataset mentioned in \Cref{sec:app:data:expert_demo}.

\paragraph{Fino-Net} Fino-Net data contains egocentric, paired success/fail data from a Baxter robot~\citep{inceoglu2020fino}. The data consists mainly of pick-and-place tasks, but it was selected for its use of a unique robot not present in the other datasets.

\subsubsection{Human only Datasets}
Finally, we also include human-only data. Our final dataset contains just one dataset, Epic-Kitchens, but early on we also experimented with EgoDex~\citep{egodex}. We found that training solely on Epic-Kitchens helped predict rewards for robot data, but this was not the case with EgoDex, perhaps due to Epic-Kitchen's background scene diversity and clutter.

\paragraph{Epic-Kitchens}
We include a subset of data from Epic-Kitchens~\citep{Damen2022EpicKitchens}. Due to some difficulties we encountered in downloading the entire dataset, we picked a subset of Epic-Kitchens 100 uploaded to HuggingFace Datasets.\footnote{\url{https://huggingface.co/datasets/awsaf49/epic_kitchens_100}}

Furthermore, because language annotations and trajectory end times are quite noisy, we use the Epic-Kitchens dataset only for preference prediction.

\subsection{Dataset Filtering and Task End-State Adjustment}
\label{sec:app:data:filtering}
To determine task-finished thresholds for each dataset, we designed a lightweight UI to visualize randomly sampled trajectories from each data source.
With this visualizer, we sample 10 trajectories from each data source and manually mark the frame at which we deem the task to be complete.
We define this threshold as the point at which the task description is satisfied. 
Then, we use the 90th percentile of end-frame thresholds (i.e., when 90\% of the visualized trajectories are complete) as our threshold.
Most teleoperators collecting data in \datasetname\ define the trajectory as complete when the robot has performed a partial or full reset to neutral.
As such, our end-state thresholds are typically set to around 80-95\% of the trajectory length.

This process takes no more than 2 minutes per data source, and the thresholds are used to appropriately adjust target progress and success thresholds for training, detailed further in \Cref{sec:app:model:objectives}.

\subsection{Individual Evaluation Dataset Details}
\label{sec:app:data:eval_datasets}

We summarize the number of trajectories in each of our in-distribution evaluation dataset in \Cref{tab:appdx:evaluation_datasets_id} and the out-of-distribution evaluation dataset in \Cref{tab:appdx:evaluation_datasets_ood} and describe them in detail below.

\paragraph{\usc\ Franka} \usc\ Franka is a dataset of mixed expertise trajectories collected between four different tabletop tasks such as ``fold towel'' and ``put the plate on the sink.'' For each quality label, we collected at least two trajectories each.

\paragraph{\usc\ Trossen} \usc\ Trossen comprises of mixed expertise trajectories collected between five different tabletop tasks using a bimanual Trossen. Some tasks are more articulated and dexterous such as "unzip the pencil case" and "stir the pot". For each quality label, we collected at least two trajectories each.

\paragraph{\usc\ Koch Arms} \usc\ Koch Arms replicates the real-world data collected in ReWiND \cite{zhang2025rewind} using bimanual Koch Arms. This dataset consists of 10 demos per-task over 20 tasks. We also collect suboptimal and failure examples for each task. 

\paragraph{\mituniv\ Franka} \mituniv\ Franka is a dataset composed of diverse tasks, including pick and place (``pick up the banana from X and place it on Y'') and dexterous tasks such as ``fold the towel in half'', ``pick up the spatula and stir the beans in the pot''. We also include a task ``pick up the mouse and place it on X while avoiding Y'' that requires semantic scene understanding. We collect trajectories of different levels of expertise for each task.

\paragraph{\utd\ SO-101}  Univ3 SO-101 is a real-world dataset of manipulation trajectories collected using a single-arm SO-101 robot. The dataset comprises two mixed-quality settings, each containing both successful and failure trajectories. The first setting is a clean pick-and-place environment centered on a single household task, “put the bread in the oven.” This setting includes 30 successful demonstrations and 45 failure trajectories. The second setting is a cluttered multitask environment consisting of three pick-and-place tasks, such as “put the red bowl on the blue plate.” For each task in this setting, we collect 20 successful demonstrations and 15 failure trajectories.

%% file: sections/appendix/model.tex
\section{Model Details}
\label{sec:app:model}

\subsection{Model Architecture and Training Parameters}
\label{sec:app:model:architecture}

\begin{table}[tb]
\centering
\label{tab:our_hyperparams}
\begin{tabular}{@{} l l @{}}
\toprule
\textbf{Parameter} & \textbf{Value} \\
\midrule
Base Model & \texttt{Qwen/Qwen3-VL-4B-Instruct} \\
Number of frames & 8 \\
Per-device batch size & 16 \\
Learning rate & $2\times10^{-5}$ \\
Weight decay & 0.01 \\
Total training steps & 6500 \\
Max sequence length & 1024 \\
LR scheduler & Cosine \\
Warmup ratio & 0.1 \\
Min frames per trajectory & 5 \\
Progress loss type & Discrete \\
Number of discrete bins & 10 \\
MLP head num hidden layers & 1 \\
MLP head dropout & 0.1 \\
MLP head hidden dim & 2048 \\
\bottomrule
\end{tabular}
\caption{\textbf{Training Configuration for \method}}
\end{table}

\begin{figure*}[t]
    \centering
    \includegraphics[width=0.75\linewidth]{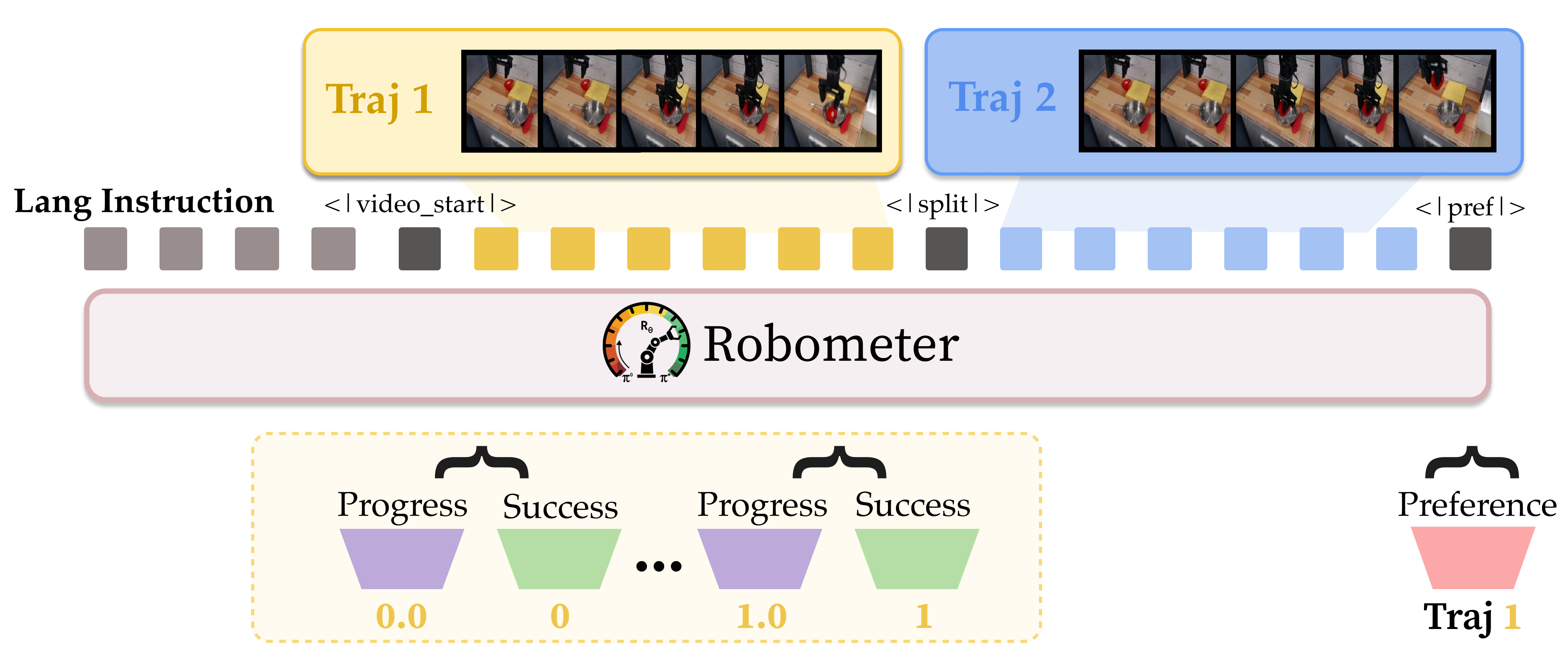}
    \caption{\textbf{\method\ model architecture.} A VLM backbone processes a language instruction with one or two trajectories. 
    For both single and paired inputs, per-frame hidden states of the first trajectory are passed to MLP heads predicting dense task progress and binary success. For paired trajectories, a preference token is used with a preference head to select the trajectory that better satisfies the instruction. The model is trained with joint progress, success, and preference objectives.}
    \label{fig:architecture}
\end{figure*}

\paragraph{Architecture}
We illustrate the overall architecture of \method\ in \Cref{fig:architecture}. \method\ instantiates a causally masked vision–language model (VLM) backbone, \textsc{Qwen3-VL-4B-Instruct}, a unified transformer that processes interleaved text and visual tokens using a single autoregressive decoder. Natural language instructions are tokenized using Qwen’s SentencePiece-based tokenizer.

Each video trajectory is first subsampled into a fixed number of frames. Each frame is independently encoded by a ViT-style visual encoder and projected into a sequence of visual tokens. A special $\vidstart$ token marks the beginning of visual input, after which visual tokens are appended sequentially and assigned unique positional indices in the unified token sequence. The decoder jointly attends over language tokens, visual tokens, and special tokens using a single causal attention mask, supporting unified multimodal reasoning.

To enable dense, frame-level reward estimation, a learned progress token $\progtoken$ is inserted after each frame $o^1_t$ in the first trajectory. Under causal masking, the hidden state of $\progtoken$ can attend only to the instruction and visual tokens from frames $o^1_{1:t}$.

We introduce a dedicated separator token $\splittoken$ to demarcate the boundary between two video trajectories. The second trajectory $o^2$ is appended after $\splittoken$ and processed jointly with $o^1$ in the same causal sequence. A learned preference token $\preftoken$ is appended at the end of the prompt; its hidden state aggregates information from the instruction and both trajectories and is used to predict pairwise preference.

We attach lightweight MLP heads to the shared VLM backbone. Specifically, the progress, preference, and success heads each consist of a two-layer MLP followed by LayerNorm, GELU activation, and dropout, and a final linear projection that outputs a scalar logit. The progress head is applied to the hidden states of the interleaved $\progtoken$ tokens to produce frame-level progress logits, while the success head operates on the corresponding per-frame hidden states to predict frame-level success logits. The preference head is applied to the hidden state of $\preftoken$ to produce a single logit indicating whether the first trajectory is preferred over the second.

For preference supervision, we construct a single multimodal prompt that contains two trajectories, serialized into a single causal sequence and separated by a split token.
This is a more detailed, expanded version of \Cref{eq:tokenized_prompt}.
\begin{equation*}
\resizebox{1.0\columnwidth}{!}{$
\begin{aligned}
\tok(l, o^A, o^B) \rightarrow \; &
\tok(l)\; \text{This is Trajectory A.''}\; \videostarttoken \; \tok(o^A_{1:T}) \\ & \splittoken \; \text{This is Trajectory B.''} \\
&\videostarttoken \;  \tok(o^B_{1:T}) \;
\preftoken.
\end{aligned}
$}
\label{eq:preference_prompt}
\end{equation*}

\noindent\textbf{Prompt.}
Since we train the model on all 3 objectives (progress, success, preference) simultaneously, we always sample a preference prompt.
Thus, we always condition the model on the following natural-language prompt.
\newif\ifprog
\progtrue %
\begin{center}
\fbox{%
\parbox{0.95\linewidth}{%
Given these two trajectories for the task ``\texttt{\{task\}}'', evaluate which one makes more progress towards the task.
Return A for the first trajectory and B for the second trajectory.
\ifprog
Additionally, predict the task progress at each frame of the first trajectory as a float between 0 and 1, where 0 corresponds to the initial state and 1 corresponds to task completion. If the robot is not performing the specified task, predict 0 progress.
\fi
}
}
\end{center}

\paragraph{Model and Training Params}
We list overall hyperparameters in \Cref{tab:our_hyperparams}.
We did not extensively sweep these hyperparameters---we followed best practices and parameters from prior work~\citep{bai2025qwen2, Qwen3-VL, lee2026roboreward}.
For the preference, success, and progress prediction MLPs, we heuristically select a hidden dimension of 2048, which is half the input size (Qwen's hidden embedding size) of 4096.

\subsection{Training Objectives}
\label{sec:app:model:objectives}
\paragraph{Trajectory Cutoffs and Success Supervision}
In several teleoperated datasets, episode termination does not coincide with task completion. 
Operators typically complete the task and then manually stop the recording after a short delay, resulting in trailing frames in which the robot remains static or performs incidental motions unrelated to task execution. 
These frames do not reflect additional task progress and introduce noise when used for frame-level progress supervision.
To mitigate this issue, we manually annotate a dataset-specific success cutoff corresponding to the frame at which the task objective is first achieved and apply this cutoff uniformly to all trajectories within the dataset (see \Cref{tab:success_cutoffs}).

For frames occurring after the dataset-specific success cutoff, we assign target progress and success labels of $1.0$, reflecting that the task has already been completed. 
Success supervision is applied selectively to avoid ambiguous or conflicting learning signals. 
Specifically, we predict success only for frames whose target progress is either strictly below a minimum success threshold $\tau_{\text{succ}}$, corresponding to clearly pre-completion states, or exactly equal to $1.0$, corresponding to completed states. 
Frames with intermediate progress values near completion are excluded from success supervision, as they often correspond to visually ambiguous transitional or stabilization phases.

\begin{table}[t]
\centering
\small
\begin{tabular}{l c}
\toprule
\textbf{Dataset} & \textbf{Success Cutoff} \\
\midrule

\multicolumn{2}{l}{\textit{Open-X Embodiment (OXE)}} \\
Aloha Mobile & 0.95 \\
Austin BUDS & 0.95 \\
Berkeley Cable Routing & 0.95 \\
Berkeley FANUC Manipulation & 0.98 \\
Bridgev2 & 0.95 \\
DLR EDAN Shared Control & 0.95 \\
CMU Pickup Insert & 0.95 \\
UCSD Kitchen & 0.95 \\
UT Austin Mutex & 0.95 \\
BC-Z & 0.95 \\
Berkeley MVP & 1.00 \\
Berkeley RPT & 0.76 \\
Fractal & 1.00 \\
Furniture Bench & 1.00 \\
DROID & 0.95 \\
Imperial College Sawyer Wrist Cam & 0.90 \\
Language Table & 1.00 \\
NYU ROT & 0.70 \\
RoboSet & 0.85 \\
Stanford HYDRA & 1.00 \\
Tokyo-U LSMO & 1.00 \\
TOTO & 1.00 \\

\midrule
\multicolumn{2}{l}{\textit{Other Datasets}} \\
MolmoAct Household & 0.94 \\
MolmoAct Tabletop & 0.94 \\
AgibotWorld & 0.95 \\
PH2D & 0.95 \\
RH20T (Human) & 0.92 \\
RH20T (Robot) & 0.94 \\
RoboArena & 0.90 \\
H2R & 0.90 \\
SOAR & 0.95 \\
AutoEval& 0.94 \\
Galaxea & 0.80 \\
FINO-Net & 0.75 \\
Humanoid Everyday & 0.80 \\
Motif & 0.95 \\

\bottomrule
\end{tabular}
\caption{\textbf{Dataset-specific success cutoffs} used to correct for delayed episode termination in teleoperated data. Datasets not contained here use a default success cutoff of 1.0.}
\label{tab:success_cutoffs}
\end{table}

\paragraph{Progress Supervision in Preference Samples}
For preference samples, we apply progress supervision only to the first trajectory (Trajectory~A), reflecting the fact that progress prediction is used at inference time for a single video. 
To avoid introducing noisy or ill-defined targets, we compute progress loss for Trajectory~A only when it corresponds to a successful trajectory.
When Trajectory~A corresponds to a failed or suboptimal execution, we do not apply progress supervision.

For datasets that provide continuous partial completion annotations, such as RoboArena, we supervise progress on the final frame of Trajectory~A using the ground-truth \texttt{partial\_success} value, even when the trajectory is suboptimal. 
This allows the model to learn calibrated progress estimates from human-annotated partial completion labels.

\paragraph{Data Source and Strategy Sampling}
Detailing \Cref{sec:augmentation} further, we construct pairs of trajectory comparisons for preference supervision (and also progress/success supervision with the first trajectory input to the model) by first sampling a \emph{preference sampling strategy} from \emph{different-task}, \emph{rewind augmentation}, or \emph{different-expertise}. 
Conditioned on the selected strategy, we restrict sampling to datasets from which the corresponding type of preference pair can be constructed.

\begin{itemize}
    \item 
For \emph{different-expertise} strategy, samples are constructed from mixed expertise datasets, such as RACER or FINO-Net, where trajectories are labeled as successful, suboptimal, or failed. 
We use these annotations to form preference pairs that rank trajectories by execution quality.
In datasets such as RoboArena that additionally provide continuous \texttt{partial\_success} annotations, we sample two trajectories from the same task and assign the trajectory with higher partial success as the preferred one.

\item \emph{Rewind augmentation} can be applied to trajectories from any dataset and does not require additional annotations.

\item For \emph{different-task pairing}, we sample a trajectory that is successful for one task and pair it with a trajectory executing a different task. 
When constructing different-task pairs, we control whether both trajectories are drawn from the same data source or from different sources. 
With probability $\rho_{\text{same}}$, we sample different-task pairs from the same dataset to discourage reliance on dataset-specific visual cues, while with probability $1-\rho_{\text{same}}$ the trajectories are drawn from different datasets to encourage robustness to domain shift.
We set $\rho_{\text{same}} = 0.5$, yielding an equal mix of same-source and cross-source different-task comparisons.
\end{itemize}

\paragraph{Preference Prediction Loss: Bradley-Terry vs BCE}
Our preference prediction loss in \Cref{eq:preference} uses a binary cross-entropy loss coming from an MLP on top of the $\preftoken$ embedding in \Cref{eq:tokenized_prompt}.
Using this token with \emph{two} input videos allows \method\ to simultaneously attend to \emph{both} videos to predict this loss, resulting in one forward and backward pass to train \method\ on preference prediction.

A common alternative in both reward modeling from pseudo-preferences~\citep{yang2024rank} and general RLHF reward function training on real human preferences~\citep{christiano2017rlhf} is to instead use the Bradley-Terry loss~\citep{Bradley1952RankAO, christiano2017rlhf}, where a single preference score is computed over an entire individual trajectory at a time, and then the loss is backpropagated across batch comparisons.

Ignoring computational efficiency differences, our main reason for not implementing the Bradley--Terry loss into \method\ is that it does not leverage the pre-trained attention mechanism as effectively: our two-video formulation allows tokens from one trajectory to explicitly attend to tokens from the other, enabling direct cross-video comparison during preference prediction. This strategy is common for comparative reasoning tasks in language reference games~\cite{monroe2017colors, mitra2024one, bao2022learning}. In contrast, Bradley--Terry computes independent scalar scores per trajectory and only couples them through the loss, which can make learning more sensitive to score calibration across batches. We compare these objectives in \Cref{sec:app:ablations}. 
As shown in \Cref{tab:abl:pref_loss_bt_vs_bce}, predicting a preference label from a dedicated $\preftoken$ given both videos (BCE) improves reward alignment (from 0.862 to 0.948) and trajectory ranking over Bradley--Terry (from 0.325 to 0.655).

\begin{table}[t]
\centering
\small
\renewcommand{\arraystretch}{1.15}
\begin{tabular}{l c c c}
\toprule
\textbf{Pref. Loss}
& \textbf{VOC $r \uparrow$}
& \textbf{Kendall $\tau \uparrow$}
& \textbf{Succ--Fail Diff.\ $\uparrow$} \\
\midrule
BT
& 0.862 & 0.325 & 0.242 \\
BCE
& \textbf{0.945} & \textbf{0.643} & \textbf{0.320} \\
\bottomrule
\end{tabular}
\caption{
\textbf{Bradley--Terry (BT) versus preference label} from a dedicated $\preftoken$ (BCE) given both videos in a single forward pass, evaluated on \evaldatasetood. 
}
\label{tab:abl:pref_loss_bt_vs_bce}
\end{table}

\subsection{Computational Resources}
We train \method\ using a per-GPU batch size of 16 across 4 GPUs, resulting in an effective batch size of 64. 
All experiments are run on a server with jobs requesting 4 NVIDIA H200 GPUs and 32 CPU threads for 6.5k training steps, corresponding to approximately 2 days of wall-clock training time. 
Unless otherwise specified, we use the same training configuration across all experiments.

%% file: sections/appendix/addtl_reward_exps.tex
\begin{table*}[htb]
  \centering
  \small
  \renewcommand{\arraystretch}{1.2}
  \begin{adjustbox}{max width=\linewidth}
  \begin{tabular}{l l cccc ccc cc}
  \toprule
  & & \multicolumn{4}{c}{Baselines} & \multicolumn{3}{c}{w/ RoboReward Training Data} & \multicolumn{2}{c}{w/ our \datasetname\ data} \\
  \cmidrule(lr){3-6} \cmidrule(lr){7-9} \cmidrule(lr){10-11}
  Split & Dataset
  & \textbf{\gvlcolor{GVL}}
  & \textbf{\vlaccolor{VLAC}-8B}
  & \textbf{\robodopaminecolor{RoboDopamine}-8B}
  & \textbf{\toprewardcolor{TOPReward}}
  & \textbf{\roborewardcolor{RoboReward}-4B}
  & \textbf{\roborewardcolor{RoboReward}-8B}
  & \textbf{\robometercolor{\method{}}}
  & \textbf{\rewindcolor{ReWiND}}
  & \textbf{\robometercolor{\method}} \\
  \midrule

  \multirow{10}{*}{\evaldatasetid}
  & RACER (Val) & 0.131 & 0.156 & 0.752 & 0.239 & 0.491 & 0.652 & 0.937 & 0.561 & \textbf{0.950} \\
  & OXE (BC-Z Eval) & 0.142 & -0.150 & 0.767 & 0.413 & 0.643 & \textbf{0.809} & 0.683 & 0.442 & 0.785 \\
  & OXE (Berkeley Cable Routing Eval) & 0.075 & -0.425 & 0.728 & 0.151 & 0.657 & 0.801 & 0.900 & 0.491 & \textbf{0.915} \\
  & OXE (Bridge V2 Eval) & 0.143 & -0.875 & 0.830 & 0.427 & 0.891 & \textbf{0.898} & 0.633 & 0.526 & 0.878 \\
  & OXE (Jaco Play Eval) & 0.114 & -0.151 & 0.873 & 0.499 & 0.793 & 0.785 & 0.861 & 0.550 & \textbf{0.894} \\
  & OXE (Toto Eval) & 0.254 & -0.416 & 0.732 & 0.734 & 0.886 & 0.939 & 0.947 & 0.340 & \textbf{0.973} \\
  & OXE (Viola Eval) & 0.264 & 0.479 & 0.750 & 0.463 & 0.915 & 0.896 & 0.967 & -0.014 & \textbf{0.988} \\
  & Metaworld (Eval) & 0.134 & 0.211 & 0.773 & 0.238 & 0.746 & 0.779 & 0.737 & 0.630 & \textbf{0.929} \\
  & Libero (90) & 0.254 & 0.217 & 0.876 & 0.586 & 0.874 & 0.846 & 0.912 & 0.592 & \textbf{0.962} \\
  \cmidrule(lr){2-11}
  & Average & 0.168 & 0.089 & 0.787 & 0.417 & 0.766 & 0.823 & 0.842 & 0.458 & \textbf{0.919} \\

  \midrule

  \multirow{6}{*}{\evaldatasetood}
  & \usc\ Franka & 0.102 & 0.356 & 0.842 & 0.480 & 0.909 & 0.909 & \textbf{0.959} & 0.772 & 0.949 \\
  & \usc\ Koch & 0.176 & 0.074 & 0.847 & 0.384 & 0.866 & 0.916 & \textbf{0.969} & 0.585 & 0.967 \\
  & \usc\ Trossen & 0.542 & 0.256 & 0.647 & 0.390 & 0.781 & 0.726 & \textbf{0.925} & 0.226 & 0.910 \\
  & \usc\ xArm & 0.282 & 0.454 & 0.840 & 0.160 & 0.896 & 0.914 & 0.951 & 0.435 & \textbf{0.980} \\
  & \mituniv\ Franka & 0.268 & 0.601 & 0.733 & 0.471 & 0.896 & 0.899 & 0.868 & 0.531 & \textbf{0.945} \\
  & \utd\ SO101 & 0.203 & 0.511 & 0.868 & 0.497 & 0.926 & \textbf{0.930} & 0.888 & 0.497 & 0.916 \\
  \cmidrule(lr){2-11}
  & Average & 0.262 & 0.375 & 0.796 & 0.397 & 0.879 & 0.882 & 0.927 & 0.508 & \textbf{0.945} \\

  \bottomrule
  \end{tabular}
  \end{adjustbox}
  \caption{
  \textbf{Per-dataset VOC reward alignment results.} Pearson correlation results (VOC) across individual datasets for \evaldatasetid\ and \evaldatasetood.
  }
  \label{tab:exp:reward:pearson_ood}
  \end{table*}

\begin{table*}[htb]
  \centering
  \small
  \renewcommand{\arraystretch}{1.2}
  \begin{adjustbox}{max width=\linewidth}
  \begin{tabular}{l ccccc ccc cc}
  \toprule
  & \multicolumn{5}{c}{Baselines} & \multicolumn{3}{c}{w/ RoboReward Training Data} & \multicolumn{2}{c}{w/ our \datasetname\ data} \\
  \cmidrule(lr){2-6} \cmidrule(lr){7-9} \cmidrule(lr){10-11}
  Dataset
  & \textbf{\gvlcolor{GVL}}
  & \textbf{\vlaccolor{VLAC}-2B}
  & \textbf{\vlaccolor{VLAC}-8B}
  & \textbf{\robodopaminecolor{RoboDopamine}-8B}
  & \textbf{\toprewardcolor{TOPReward}}
  & \textbf{\roborewardcolor{RoboReward}-4B}
  & \textbf{\roborewardcolor{RoboReward}-8B}
  & \textbf{\robometercolor{\method{}}}
  & \textbf{\rewindcolor{ReWiND}}
  & \textbf{\robometercolor{\method}} \\
  \midrule

  \usc\ Franka
  & 0.250 & 0.292 & 0.271 & 0.479 & 0.167 & 0.625 & 0.625 & 0.583 & -0.125 & \textbf{0.625} \\

  \usc\ Koch
  & -0.008 & 0.167 & 0.064 & 0.442 & 0.183 & 0.332 & 0.264 & \textbf{0.533} & 0.336 & 0.477 \\

  \usc\ Trossen
  & 0.292 & -0.111 & -0.417 & 0.333 & -0.292 & 0.333 & 0.389 & \textbf{0.646} & 0.028 & 0.569 \\

  \usc\ xArm
  & 0.056 & 0.167 & 0.139 & 0.431 & -0.111 & 0.528 & 0.347 & 0.403 & -0.167 & \textbf{0.750} \\

  \mituniv\ Franka
  & 0.306 & -0.017 & 0.072 & 0.333 & 0.286 & 0.494 & 0.396 & 0.479 & 0.080 & \textbf{0.572} \\

  \utd\ SO101
  & 0.300 & -0.033 & 0.167 & 0.700 & 0.533 & 0.700 & 0.767 & 0.667 & -0.067 & \textbf{0.867} \\

  \midrule
  \textbf{Average}
  & 0.199 & 0.077 & 0.049 & 0.453 & 0.128 & 0.502 & 0.465 & 0.552 & 0.014 & \textbf{0.643} \\
  \bottomrule
  \end{tabular}
  \end{adjustbox}
  \caption{\textbf{Per-dataset trajectory ranking results.} Trajectory ranking results on individual \evaldatasetood\ datasets.}
  \label{tab:exp:trajectory_ranking_ood}
  \end{table*}

\section{Additional Reward Evaluation Results}
\label{sec:app:reward_eval}
We first list baseline implementation details for main paper baselines before discussing additional reward evaluation results.

\paragraph{VLAC} 
VLAC released 2 models built on the InternVL~\citep{chen2024internvl} VLM, with 2B and 8B parameters.
VLAC takes 2 frames as input and predicts the relative increase in progress between the first and second frames; thus, its output range is $[-1, 1]$.
We directly perform inference using the publicly available GitHub code and test both the 2B and 8B models---we found the 8B model to perform slightly better on our evaluation results and thus use VLAC-8B as our baseline. The pretrained VLAC checkpoints are obtained from the authors’ public release.\footnote{\url{https://huggingface.co/InternRobotics/VLAC}}
For relevant reward comparisons, we normalize its outputs to make them directly comparable to those of all other models that output only positive rewards.

\paragraph{Robo-Dopamine} trains a step-aware \emph{general process reward model} (GRM) to estimate fine-grained manipulation progress from \emph{multi-view} observations by predicting discretized progress ``hops'' between states and aggregating these signals into a dense reward. We use the authors' released inference code and 8B checkpoint.\footnote{\url{https://github.com/FlagOpen/Robo-Dopamine}}\footnote{\url{https://huggingface.co/tanhuajie2001/Robo-Dopamine-GRM-2.0-8B-Preview}} Robo-Dopamine relies on richer task-specific context at inference time (e.g. such as multi-view and goal images), which may not be consistently available across datasets or deployment settings.

\paragraph{GVL}
We compare against GVL, which queries a closed-source vision-language model to estimate task progress directly from video frames. 
GVL does not involve model fine-tuning; instead, progress is inferred solely through prompting at inference time. 
In our experiments, we use \texttt{GPT-5-mini},\footnote{\texttt{GPT-5-mini-2025-08-27} the latest version as of writing.} which is the strongest-performing closed-source model reported in the RoboRewardBench~\citep{lee2026roboreward} benchmark.
To discourage the model from exploiting trivial temporal cues and to reduce correlations with frame index, GVL shuffles video frames during inference and prompts the model to predict progress values that are subsequently reordered chronologically.

\paragraph{TOPReward} prompts a pre-trained VLM with the question ``does the trajectory complete the task?'' for a given task instruction and robot trajectory. 
Then, it extracts the log-likelihood of the ``true'' token as the reward before normalization to bound the reward between $0$ and $1$.
They experiment on: Gemini-2.5-Pro~\citep{geminiteam2024geminifamilyhighlycapable}, Molmo2~\citep{molmo2openweightsdata}, and Qwen-3-VL-Instruct-8B~\citep{Qwen3-VL}, finding that Qwen-3 works the best overall on VOC scores. 
Thus, we use TOPReward with Qwen-3-VL-8B in our experiments and for cross-trajectory comparisons (such as Kendall-$\tau_a$) we do not normalize intra-trajectory rewards (otherwise most trajectories will end at near-1.0 reward).

\paragraph{ReWiND}
We implement ReWiND~\cite{zhang2025rewind}, which operates on precomputed visual and language embeddings rather than raw pixels. 
Each video frame is encoded using a frozen DINOv2 vision encoder~\cite{oquab2024dinov2learningrobustvisual}, and task instructions are embedded using the frozen Sentence-Transformers MiniLM-L6-v2 model (\texttt{all-MiniLM-L6-v2}~\citep{reimers-2019-sentence-bert}). 
The resulting visual and text embeddings are linearly projected into a shared latent space and processed by a Transformer encoder.

To enable a direct architectural comparison with \method, we predict per-frame progress and success directly from the hidden states corresponding to visual frame embeddings using lightweight MLP heads, and train progress using a discrete bin formulation. 
In our ablation study in \Cref{tab:exp:ablations}, we further extend the ReWiND baseline with a preference objective by concatenating two trajectories into a single sequence and predicting a binary preference label via a learned preference token, and we scale up its original architecture (4 layers, 8 attention heads) to a hidden dimension of 1024, 32 transformer layers, and 16 attention heads. 

\paragraph{RoboReward} is trained to predict a discrete 1-5 progress target on OXE and RoboArena. 
RoboReward is trained using trajectory-level supervision and does not model intermediate task progress within a trajectory. 
As a result, the model is primarily designed to provide a sparse, terminal reward.
For fair evaluation and comparison with our dense reward formulation, we obtain frame-level rewards from RoboReward by running inference independently on each frame of a trajectory and treating the resulting predictions as per-frame rewards. 
We use the publicly released RoboReward-8B pretrained checkpoint provided by the authors.\footnote{\url{https://huggingface.co/teetone/RoboReward-4B}}\footnote{\url{https://huggingface.co/teetone/RoboReward-8B}}

\subsection{Preference Prediction}
\label{see:app:reward_eval:preference}

\paragraph{RL-VLM-F}
RL-VLM-F~\citep{wangrlvlmf2024} predicts trajectory preferences by prompting a
closed-source vision--language model to compare the \emph{final frame} of two trajectories
conditioned on a task description. In our experiments, we instantiate RL-VLM-F using the
same OpenAI \texttt{GPT-5-mini} model as for GVL. 
RL-VLM-F uses the following fixed prompt for preference prediction.

\begin{center}
\fbox{%
\begin{minipage}{0.95\linewidth}
\small
Each frame comes from a robot trajectory.
(Think causally and use image comparison to verify any confusion
between the base of the robot and the end effector.)

\begin{enumerate}
  \item What is shown in the first image (Image A)?
  \item What is shown in the second image (Image B)?
  \item For this question, here is the Goal Text: \texttt{GOAL\_TEXT}
\end{enumerate}

Is the goal being better achieved in Image A or Image B?

Reply with a single line containing \texttt{0} if the goal is better achieved in Image A,
or \texttt{1} if the goal is better achieved in Image B.
Reply \texttt{-1} if there is no discernible difference or progress.
\end{minipage}}
\end{center}

For RL-VLM-F, preferences are inferred using only the final frame
of each trajectory, whereas \method\ applies a learned preference head that
directly compares full video sequences.
Preference accuracy is computed as the fraction of pairwise comparisons in
which the predicted ordering matches the ground truth.
To evaluate preference quality, we construct trajectory pairs from the
\evaldatasetood\ split along two axes: (i) differing task instructions and
(ii) differing trajectory quality labels.
For each dataset, we randomly sample 500 pairwise trajectory comparisons
and evaluate predicted preferences against ground-truth ordering labels.
As shown in \Cref{tab:quality_preference_breakdown,tab:task_preference_breakdown}, \method\ consistently outperforms RL-VLM-F, improving average preference accuracy by 27.0\% on different-quality pairs and by 32.4\% on different-task pairs.

\begin{table}[t]
\centering
\small
\renewcommand{\arraystretch}{1.2}
\begin{tabular}{l c c}
\toprule
\textbf{Dataset} & \textbf{RL-VLM-F (\%)} & \textbf{\method\ (\%)} \\
\midrule
\usc\  Franka 
& 52.1 & \textbf{75.0} \\

\usc\ Koch 
& 54.4 & \textbf{79.4} \\

\usc\  Trossen 
& 66.7 & \textbf{76.2} \\

\usc\ xArm 
& 48.6 & \textbf{88.9} \\

\mituniv\ Franka 
& 54.4 & \textbf{85.4} \\

\utd\ SO-101 
& 56.7 & \textbf{90.0} \\
\midrule
\textbf{Average} 
& 55.5 & \textbf{82.5} \\
\bottomrule
\end{tabular}
\caption{\textbf{RL-VLM-F vs \method\ on Trajectory Quality}. Different quality trajectory pairwise preference accuracy on 500 comparisons for each of the \evaldatasetood\ datasets.}
\label{tab:quality_preference_breakdown}
\end{table}

\begin{table}[t]
\centering
\small
\renewcommand{\arraystretch}{1.2}
\begin{tabular}{l c c}
\toprule
\textbf{Dataset} & \textbf{RL-VLM-F (\%)} & \textbf{\method\ (\%)} \\
\midrule
\usc\  Franka 
& 70.7 & \textbf{100.0} \\

\usc\ Koch 
& 54.7 & \textbf{89.8} \\

\usc\  Trossen 
& 64.0 & \textbf{99.0} \\

\usc\ xArm 
& 73.3 & \textbf{98.2} \\

\mituniv\ Franka 
& 55.3 & \textbf{98.4} \\

\utd\ SO-101 
& 72.7 & \textbf{100.0} \\

\midrule
\textbf{Average} 
& 65.1 & \textbf{97.6} \\
\bottomrule
\end{tabular}
\caption{\textbf{RL-VLM-F vs \method\ on Different Tasks}. Different task trajectory pairwise preference accuracy on 500 comparisons for each of the \evaldatasetood\ datasets.}
\label{tab:task_preference_breakdown}
\end{table}

\begin{table*}[htb]
\centering
\small
\renewcommand{\arraystretch}{1.2}
\begin{adjustbox}{max width=\linewidth}
\begin{tabular}{l c c c c c c c c c}
\toprule
& \multicolumn{3}{c}{\textbf{VOC $r \uparrow$}} 
& \multicolumn{3}{c}{\textbf{Kendall $\tau \uparrow$}} 
& \multicolumn{3}{c}{\textbf{Succ--Fail Diff. $\uparrow$}} \\
\cmidrule(lr){2-4} \cmidrule(lr){5-7} \cmidrule(lr){8-10}
\textbf{Dataset} 
& \textbf{Prog. Only} 
& \textbf{+Preference} 
& \textbf{\method} 
& \textbf{Prog. Only} 
& \textbf{+Preference} 
& \textbf{\method} 
& \textbf{Prog. Only} 
& \textbf{+Preference} 
& \textbf{\method} \\
\midrule

\usc\ Franka
& 0.913 & \textbf{0.974} & 0.949
& 0.083 & 0.542 & \textbf{0.625}
& 0.039 & \textbf{0.428} & 0.326 \\

\usc\ Koch Arm
& 0.933 & 0.932 & \textbf{0.967}
& 0.231 & 0.357 & \textbf{0.477}
& 0.081 & 0.142 & \textbf{0.191} \\

\usc\ Trossen
& 0.199 & 0.902 & \textbf{0.910}
& 0.333 & 0.423 & \textbf{0.569}
& 0.052 & 0.231 & \textbf{0.312} \\

\usc\ xArm
& 0.890 & 0.973 & \textbf{0.980}
& 0.389 & 0.597 & \textbf{0.750}
& 0.079 & 0.154 & \textbf{0.345} \\

\mituniv\ Franka
& 0.936 & 0.942 & \textbf{0.945} 
& 0.183 & 0.458 & \textbf{0.572}
& 0.063 & 0.223 & \textbf{0.310} \\

\utd\ SO101
& \textbf{0.964} & 0.899 & 0.916
& 0.533 & 0.667 & \textbf{0.867}
& 0.134 & 0.244 & \textbf{0.438} \\

\midrule
\textbf{Average}
& 0.806 & 0.939 & \textbf{0.945}
& 0.292 & 0.507 & \textbf{0.643}
& 0.075 & 0.237 & \textbf{0.320} \\
\bottomrule
\end{tabular}
\end{adjustbox}
\caption{
\textbf{Per-dataset model ablation results.} Reward alignment, trajectory ranking, and final reward difference between successful and failed trajectories on \evaldatasetood
}
\label{tab:exp:rbm_ablations}
\end{table*}

\subsection{\method{} in Long Horizon Tasks}
\label{see:app:reward_eval:long_horizon}

\begin{figure}
    \centering
    \includegraphics[width=\linewidth]{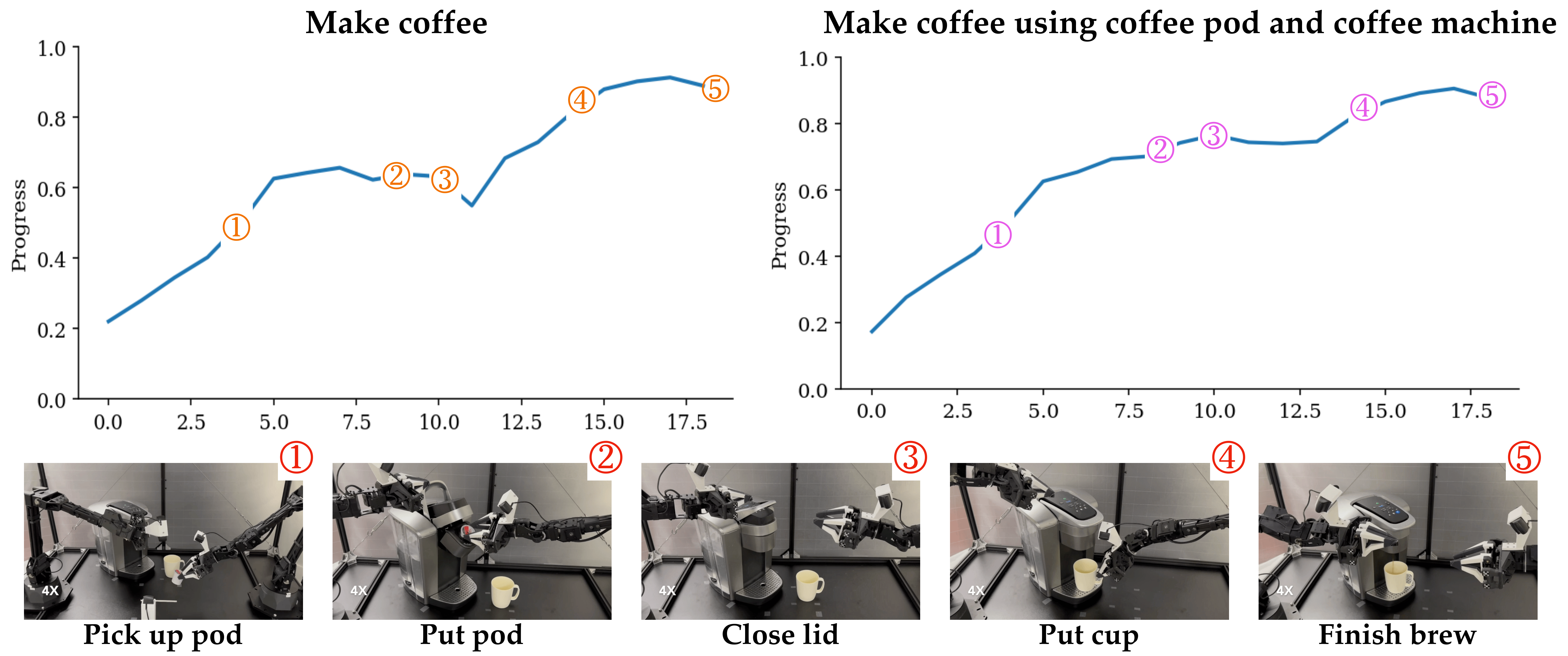}
    \caption{\textbf{Qualitative analysis of \method\ on a long-horizon task.} With sufficiently detailed task instructions, \method\ correctly assigns increasing rewards across multiple stages of the task.}
    \label{fig:appdx:long_horizon}
\end{figure}

\method{}’s progress predictions are conditioned on the \emph{instruction}, rather than simple geometric proximity; as long as the instruction is sufficiently detailed, it can assign increasing progress even when the robot temporarily moves away from intermediate objects to complete the task. We analyze this behavior in \Cref{fig:appdx:long_horizon} on the \emph{“Make coffee”} task, which consists of multiple steps such as picking up the pod, closing the lid, and placing the cup correctly.

When the task instruction is brief or underspecified, \method{}’s predicted progress tends to stagnate or occasionally decrease. In contrast, with more detailed instructions, the model assigns consistently increasing rewards as the task progresses, shown on the right side of the figure. An alternative approach to handling long-horizon tasks is to decompose them into multiple stages, as demonstrated in our multi-stage Automatic Online RL experiments (\Cref{sec:exp:marquee_exps}).

\subsection{RoboRewardBench Evaluation}
\label{sec:app:reward_eval:roborewardbench}
We further evaluate on the \emph{external} RoboRewardBench benchmark~\citep{lee2026roboreward}, reporting Mean Absolute Error (MAE) on rewards discretized into 1--5 scores. Baseline results under the same protocol are shown in \Cref{tab:exp:reward:roborewardbench} with a \method\ variant using only RoboReward data with matching 5-bin outputs.
This RoboReward-only \method\ variant attains an MAE of 0.75, outperforming RoboReward-4B. 
Finally, note that in our OOD evaluations in \Cref{tab:exp:reward}, RoboReward-8B and 4B perform similarly, and GPT-5-mini with GVL performs markedly worse. We therefore attribute the stronger results of larger models on RoboRewardBench to its 5 discrete labels and its final-frame-only evaluation protocol.

\begin{table}[htb]
\centering
\setlength{\tabcolsep}{2pt}
\renewcommand{\arraystretch}{0.95}
\begin{tabular}{p{1.5cm} l r}
\toprule
\textbf{Model Type} & \textbf{Model} &
\makecell[r]{\textbf{RoboRewardBench}\\\textbf{MAE ($\downarrow$)}} \\
\midrule
\multirow{4}{*}{\makecell[c]{Qwen3-4B \\ Models}} 
& \robometercolor{\method} (only RoboReward data) & \textbf{0.75} \\
& \roborewardcolor{RoboReward}-4B & 0.85 \\
& Qwen3-VL-4B-Instr. & 1.03 \\
\midrule
\multirow{4}{*}{\makecell[c]{Closed / \\Larger}} &
\roborewardcolor{RoboReward}-8B & \textbf{0.67} \\
& GPT-5-mini & 0.69 \\
& Qwen3-VL-8B-Instr. & 0.89 \\
& Gemini-2.5-pro & 0.90 \\
\bottomrule
\end{tabular}
\caption{Evaluation on the RoboRewardBench benchmark~\citep{lee2026roboreward}.}
\label{tab:exp:reward:roborewardbench}
\end{table}

%% file: sections/appendix/addtl_ablations.tex
\section{Additional Ablations}
\label{sec:app:ablations}

\begin{table*}[htb]
\centering
\small
\renewcommand{\arraystretch}{1.2}
\begin{adjustbox}{max width=\linewidth}
\begin{tabular}{l c c c c c c}
\toprule
& \multicolumn{2}{c}{\textbf{VOC $r \uparrow$}} 
& \multicolumn{2}{c}{\textbf{Kendall $\tau \uparrow$}} 
& \multicolumn{2}{c}{\textbf{Succ--Fail Diff. $\uparrow$}} \\
\cmidrule(lr){2-3} \cmidrule(lr){4-5} \cmidrule(lr){6-7}
\textbf{Model}
& \textbf{\evaldatasetood} & \textbf{LIBERO (90)}
& \textbf{\evaldatasetood} & \textbf{LIBERO (90)}
& \textbf{\evaldatasetood} & \textbf{LIBERO (90)} \\
\midrule

No Different Task
& 0.930 & 0.966
& 0.560 & 0.903
& 0.260 & 0.362 \\

No Rewind
& 0.860 & 0.818
& 0.480 & 0.815
& 0.150 & 0.241 \\

No Subopt
& 0.915 & 0.910
& 0.585 & 0.890
& 0.235 & 0.413 \\

Ours
& \textbf{0.945} & \textbf{0.976}
& \textbf{0.643} & \textbf{0.919}
& \textbf{0.320} & \textbf{0.455} \\
\bottomrule
\end{tabular}
\end{adjustbox}
\caption{\textbf{Data sampling strategy ablation results.}
Reward alignment, trajectory ranking, and final reward difference between successful and failure trajectories on \evaldatasetood{} and LIBERO-90.
Models are trained without using \evaldatasetood{} scenes; \evaldatasetood{} and LIBERO-90 are held out for evaluation.}
\label{tab:exp:rbm_ood_and_libero_ablations_policy_ranking}
\end{table*}

\begin{table*}[htb]
\centering
\small
\renewcommand{\arraystretch}{1.2}
\begin{tabular}{l c c c c c c}
\toprule
& \multicolumn{2}{c}{\textbf{VOC $r \uparrow$}} 
& \multicolumn{2}{c}{\textbf{Kendall $\tau \uparrow$}} 
& \multicolumn{2}{c}{\textbf{Succ--Fail Diff. $\uparrow$}} \\
\cmidrule(lr){2-3} \cmidrule(lr){4-5} \cmidrule(lr){6-7}
\textbf{Weights $(\lambda_{\text{pref}}, \lambda_{\text{prog}})$}
& \textbf{LIBERO (10)} & \textbf{LIBERO (90)}
& \textbf{LIBERO (10)} & \textbf{LIBERO (90)}
& \textbf{LIBERO (10)} & \textbf{LIBERO (90)} \\
\midrule
$(2,\,1)$
& 0.983 & 0.964
& 0.981 & 0.898
& 0.347 & 0.352 \\

$(1,\,2)$
& 0.986 & 0.946
& 0.982 & 0.875
& 0.423 & 0.367 \\

$(1, 1)$
& \textbf{0.994} & \textbf{0.976}
& \textbf{0.986} & \textbf{0.919}
& \textbf{0.483} & \textbf{0.455} \\
\bottomrule
\end{tabular}
\caption{\textbf{Loss weight ablation results.} 
Hyperparameter sweep over preference/progress loss weights $(\lambda_{\text{pref}}, \lambda_{\text{prog}})$ on LIBERO-10 and LIBERO-90. Each model is trained on LIBERO-(10, Spatial, Object, Goal) and LIBERO-90 is heldout for evaluation only.
}
\label{tab:exp:libero_loss_weight_sweep}
\end{table*}

\subsection{Preference pair sampling ablations} 
As shown in \Cref{tab:exp:rbm_ood_and_libero_ablations_policy_ranking}, we ablate our preference pair sampling strategies to isolate which parts of the data construction are responsible for \emph{generalization}. We evaluate on held-out \evaldatasetood{} and the out-of-distribution LIBERO-90 benchmark, neither of which is used for training. We remove one strategy at a time from \Cref{sec:augmentation} to measure its contribution to held-out ranking and reward separation.

Across both held-out benchmarks, removing any component hurts performance, with the largest drop coming from disabling \textbf{trajectory rewinding}, which most strongly reduces success--failure separation and degrades trajectory ranking. \textbf{Different-task negatives} primarily affect instruction grounding: dropping them yields a consistent (though smaller) degradation in OOD ranking. Finally, \textbf{suboptimal trajectory} pairs help calibrate rewards across mixed-quality behavior; removing them reduces ranking quality and narrows the margin between successful and failed trajectories. Overall, each preference construction contributes a complementary signal, and combining them yields the strongest held-out ranking and reward separation.

\subsection{\evaldatasetood\ Pre-Training Objective Ablations}
\label{see:app:reward_eval:additional_results}

As shown in \Cref{tab:exp:rbm_ablations}, incorporating trajectory-level preference supervision consistently improves reward-model behavior compared to training on progress labels alone. 
On average, adding preferences boosts reward alignment improves trajectory ranking, indicating that pairwise comparisons provide a strong signal for ordering failed, suboptimal, and successful behaviors. 
\method{} achieves the best average Kendall score and the largest success–failure final reward difference, while maintaining high alignment. Compared to the \textit{+Preference} variant, \method{} additionally incorporates failure trajectories during training, thereby further improving discrimination between successful and unsuccessful behaviors.
Notably, \method\ yields consistent gains across all OOD datasets, suggesting that jointly leveraging dense progress targets, preference supervision, and failure data produces a reward function that both tracks task completion and better discriminates overall trajectory quality in OOD settings.

Moreover, \Cref{tab:exp:success_head_ablation} shows that removing the success head leads to a small drop in reward generalization, indicating that success prediction provides a useful training signal beyond its deployment-time use.

\begin{table}[h]
    \centering
    \scriptsize
    \setlength{\tabcolsep}{3.5pt}
    \renewcommand{\arraystretch}{1.1}
    \resizebox{\columnwidth}{!}{%
    \begin{tabular}{l ccc ccc}
    \toprule
    & \multicolumn{3}{c}{LIBERO-90} & \multicolumn{3}{c}{\evaldatasetood{}} \\
    \cmidrule(lr){2-4} \cmidrule(lr){5-7}
    Model
    & VOC $r \uparrow$ & Kend. $\tau \uparrow$ & S-F $\uparrow$
    & VOC $r \uparrow$ & Kend. $\tau \uparrow$ & S-F $\uparrow$ \\
    \midrule
    \textbf{\method{} (-succ)} & 0.97 & 0.89 & 0.40 & \textbf{0.94} & 0.60 & 0.27 \\
    \textbf{\method{}} & \textbf{0.98} & \textbf{0.92} & \textbf{0.46} & \textbf{0.94} & \textbf{0.64} & \textbf{0.32} \\
    \bottomrule
    \end{tabular}%
    }
    \caption{\textbf{Success prediction head ablation results.} Reward alignment, trajectory ranking, and final reward difference between successful and failure trajectories on \evaldatasetood{} and LIBERO-90.}
    \label{tab:exp:success_head_ablation}
\end{table}

Additionally, we report detailed evaluation results for each dataset in \evaldatasetood\ in \Cref{tab:exp:reward:pearson_ood} (reward alignment) and \Cref{tab:exp:trajectory_ranking_ood} (trajectory ranking) for all baselines.

\subsection{Training loss weight ablations}
\Cref{tab:exp:libero_loss_weight_sweep} sweeps the relative weighting between the preference and progress losses. We find that the best performance is achieved when the two objectives are weighted uniformly, with $(\lambda_{\text{pref}}, \lambda_{\text{prog}})=(1,\,1)$ outperforming settings that upweight either preference or progress on LIBERO-90 across alignment, ranking, and success--failure separation.

%% file: sections/appendix/policy_learning_exps.tex
\section{Policy Learning Experiment Details}
\label{sec:app:policy_learning}

\subsection{RL with Ablated Reward Models}
\textbf{Experiment Setup.}
For each ablated reward model described in \Cref{sec:exp:reward_model_ablations}, we train an RL policy from scratch using SAC~\citep{haarnoja2017soft}. We evaluate performance on two tasks from the LIBERO-90 suite: Task 28 (\emph{close the top drawer}) and Task 33 (\emph{close the microwave}). Both reward estimation and RL training use only the external camera view. The policy observations consist of a DINO-v2-small~\citep{oquab2024dinov2learningrobustvisual}-featurized external image concatenated with proprioceptive state inputs. During training, we perform 25 evaluation episodes every 5000 training steps and report the average success rate $\pm$ standard deviation. For each reward model, we train 5 random seeds. The SAC hyperparameters are provided in \Cref{tab:appdx:rl_ablation_sac_config}.

\begin{table}[h]
\centering
\small
\renewcommand{\arraystretch}{1.2}
\begin{tabular}{l c}
\toprule
\textbf{Hyperparameter} & \textbf{Value} \\
\midrule
Batch size & 128 \\
Target update rate $\tau$ & 0.005 \\
Discount factor $\gamma$ & $0.99$ \\
Target update interval & 1 \\
Number of critics & 5 \\
Critics sampled per update & 2 \\
Pooled critic features & True \\
Actor updates per train step & 1 \\
Critic updates per actor update & 1 \\
Actor learning rate & $1\times10^{-5}$ \\
Critic learning rate & $1\times10^{-5}$ \\
Entropy coeff learning rate & $3\times10^{-4}$ \\
Target entropy & 0 \\
Learning starts & 5000 steps \\
\bottomrule
\end{tabular}
\caption{\textbf{RL with Ablated Reward Models.} SAC hyperparameters.}
\label{tab:appdx:rl_ablation_sac_config}
\end{table}

\subsection{Automatic Online RL}
\label{sec:app:policy_learning:automatic-rl}

\paragraph{Environment Setup}
We setup a DROID~\citep{khazatsky2024droid}-style Franka Panda environment on a cluttered tabletop as depicted in \Cref{fig:exp:dsrl_fig}.
Following DROID, we have a Robotiq gripper, an exterior Zed camera mounted to the left of the robot arm, and a Zed wrist camera.
\begin{itemize}
    \item \textbf{Single Stage}: The task is to put the bowl onto the table, where the difficulty lies in the bowl starting in a tall dish rack which physically blocks the robot gripper if it approaches the bowl from too low of an angle. The clutter also makes it difficult for $\pi_0$ to perform this task with high success rates zero-shot (20\% initial success rate), making it well-suited for dense reward RL. The task instruction given to $\pi_0$ is ``put the bowl on the table.''
    \item \textbf{Multi-Stage}: The first task is to put the corn in the pot, and the second is to then put the lid on the corn, emulating a ``steam corn'' task.
    The pot being in the dish rack confuses $\pi_0$, so the base policy tends to either miss putting the corn in the pot or collide with the dish rack and get stuck.
    The exact instructions for $\pi_0$ are ``put the corn in the pot located in the dish rack'' and ``put the lid on the pot located in the dish rack.''
\end{itemize}

\paragraph{RL Algorithm} We train an RL policy from scratch with Diffusion Steering (DSRL)~\citep{wagenmaker2025steering} to steer a frozen $\pi_0$ policy pre-trained on the DROID~\citep{khazatsky2024droid} dataset.
The DSRL policy trains an SAC~\citep{haarnoja2017soft}-style algorithm that operates over the \emph{noise space} of the $\pi_0$ flow matching head.
Specifically, the inputs to the RL policy are:
\begin{itemize}
    \item DINO-v2-small~\citep{oquab2024dinov2learningrobustvisual}-featurized wrist image (384-dim)
    \item $\pi_0$ VLM hidden embedding (2048-dim)
    \item Proprioceptive joint and gripper positions (8-dim)
    \item For multi-stage only: the index of the current stage the policy is in (1-dim), concatenated with the proprioception.
\end{itemize}

The output is an action chunk of length 8, where each action is of dimension 32 ($\pi_0$'s flow matching head noise dimension) and the action bounds are $[-2.0, 2.0]$ for single-stage and $[-1.5, 1.5]$ for multi-stage.
The original DSRL paper proposed an MLP policy that outputs a single output noise action, which is copied $10$ times ($\pi_0$-DROID has a default action length of 10).
Unlike the original DSRL paper, we parameterize the policy with a transformer backbone so that it can output a unique noise embedding for each action in the 8-length action chunk---we found this action chunked version to perform more meaningful exploration and thus learn quicker. We copy the first 2 noise actions of the action chunk to fill the remaining $2$ noise timesteps for $\pi_0$-DROID, but we only execute 8 actions out of 10.

Similarly, the Q function takes as input an $8$-length action chunk but otherwise follows a standard Q-function formulation.
The network architectures follow the same one used in the Offline RL experiments, whose details are in \Cref{tab:transformer_architecture}.
SAC algorithm hyperparameters are detailed below.

\begin{table}[h]
\centering
\small
\renewcommand{\arraystretch}{1.2}
\begin{tabular}{l c}
\toprule
\textbf{Hyperparameter} & \textbf{Value} \\
\midrule
Batch size & 64 \\
Target update rate $\tau$ & 0.005 \\
Discount factor $\gamma$ & $0.995$ \\
Target update interval & 1 \\
Number of critics & 4 \\
Critics sampled per update & 2 \\
Pooled critic features & True \\
Actor updates per train step & 10 (single), 12 (multi) \\
Critic updates per actor update & 4 (single), 2 (multi) \\
Actor learning rate & $5\times10^{-5}$ \\
Critic learning rate & $1\times10^{-4}$ \\
Entropy coeff learning rate & $1\times10^{-4}$ \\
Target entropy & 0 \\
Learning starts & 1200 steps \\
Training steps & 10000 steps \\
\bottomrule
\end{tabular}
\caption{\textbf{Automatic Online RL.} DSRL SAC hyperparameters.}
\label{tab:appdx:dsrl_sac_config}
\end{table}

We run RL for 10000 environment steps, corresponding to about 40 minutes of total real-world experiment time.
We train after each episode rather than after each step, and we design an \emph{asynchronous} reward relabeling pipeline that relabels rewards in the replay buffer using reward models after they've been added, to prevent reward-related latency.

\textbf{Success Detection.} Success detection and episode termination come from the reward models. 
With \method, we use the model's success prediction probability and threshold it so that if the last timestep's success detection probability is $> 0.6$, the episode is marked as a success and terminated.
With RoboReward, we use its discrete reward predictions where we mark the episode as successful if it predicts 5/5 reward. If no success is detected by 240 timesteps, the episode terminates and resets.

For the multi-stage experiment, we run for 200 timesteps per stage. If the first stage (corn in pot) fails after 200 timesteps, we reset. 
If the reward model detects success (as in single-stage) in the first stage, we advance to the second stage, where the episode timeout is reset back to 200 timesteps.
If there's a failure in the second stage, we still reset the entire scene back to before the first stage was completed.
Each stage advance simply updates the language instruction for $\pi_0$ and the task index for the DSRL policy/Q function.

\textbf{Environment Reward.} For the single-stage setup, the base reward for the task is $-1/0$, where $-1$ is given at every step except success, where $0$ is given. This base reward is added to the reward predictions $\in [0, 1]$, thus bounding the per-step reward to be $[-1, 1]$, where the only possible positive reward signal comes at the end of a successful trajectory.

In the multi-stage setup, the base reward when the policy is in the first stage is $-2$ at each timestep, added to the reward predictions from the reward model. When the policy enters the second stage, the base reward is $-1/0$ just like in the single-stage setup. 
This simple reward design always encourages the policy to advance to the next stage, even under possibly suboptimal rewards.

\paragraph{Results}
Our single-stage results in \Cref{fig:exp:dsrl_fig} show a 55\% improvement in performance of \method\ over RoboReward after 10k online RL steps.
A key failure mode of RoboReward is its tendency to predict high rewards even when the robot is executing the wrong task.
In cluttered environments, RoboReward frequently assigns a maximum reward (5/5) when the robot manipulates objects that are not the target bowl.
This leads to a large number of false positives, as quantified in \Cref{tab:reward_fp_tp}.

\begin{table}[h]
    \centering
    \begin{tabular}{lcc}
        \toprule
        Method & True Positives & False Positives \\
        \midrule
        RoboReward & 6 & 45 \\
        \method & 18 & 0 \\
        \bottomrule
    \end{tabular}
    \caption{\textbf{Automatic Online RL reward relabeling.} True positives (TP) and false positives (FP) for reward predictions during online RL.}
    \label{tab:reward_fp_tp}
\end{table}

Finally, in our multi-stage results in \Cref{fig:exp:dsrl_fig}, RoboReward does not improve the $\pi_0$ policy at all. 
This failure to improve stems from how the $\pi_0$ policy tends to fail the first subtask, putting the corn in the pot, by dropping it \emph{near} but not in the pot. 
RoboReward often falsely assigns a 5/5 score to these failed states, allowing the policy to move onto the next stage.
As a result, during evaluation, the RoboReward-trained DSRL policy often drops the corn into various parts of the dish rack rather than into the pot.
\method{}, on the other hand, rarely assigns these states as successful, enabling the policy to learn more.

\subsection{Combining Noisy and Expert Trajectories via Offline
RL}
\label{sec:app:policy_learning:offline rl}

\paragraph{Procedure and Data Collection}
We study how different reward formulations affect downstream policy learning when training on
a mixture of expert and noisy trajectories using offline reinforcement learning. Specifically,
we compare three reward settings: (i) a sparse terminal reward, (ii) rewards predicted by
RoboReward, and (iii) rewards predicted by \method.

We evaluate offline RL on the SO-101 robot platform under two settings across two manipulation tasks:
\emph{Put the bread in the oven} and 
\emph{Put the red bowl on the blue plate}.
\textbf{Setting 1} is a clean single-task setting that evaluates \emph{Put the bread in the oven}; 
\textbf{setting 2} is a cluttered multi-task setting that evaluates \emph{Put the red bowl on the blue plate}. 
We visualize these settings in \Cref{fig:exp:offline_rl}.

For each task, we collect a mixed-expertise offline dataset containing both expert demonstrations and failed executions. In \textbf{Setting 1}, we collect 30 successful trajectories and 45 failed trajectories for \emph{Put the bread in the oven}. 
In \textbf{Setting 2}, we collect 20 successful trajectories and 15 failed trajectories each for 3 separate tasks:  \emph{Put the marker in the pen cup}, and \emph{Put the red cup on the purple coaster}, and the evaluation task \emph{Put the red bowl on the blue plate}. This second setting is intentionally more difficult as it includes more clutter, more object/goal randomization, multi-task distractors, and fewer successful demonstrations.
With offline RL, transitions from other tasks, whether the data is suboptimal or expert, should be useful for the \emph{Put the red bowl on the blue plate} task as they share a similar cluttered scene.

Observations include images from both an external camera and a wrist-mounted camera, as well as proprioceptive states. During offline RL training, we use images from both camera views, while reward estimation is performed using the external camera view only.

\paragraph{Offline RL Setup and IQL Objectives}
We follow an offline RL training setup similar to ReWiND~\citep{zhang2025rewind}.
For all methods, we train policies using Implicit Q-Learning (IQL)~\cite{kostrikov2022offline}
with identical actor, critic, and value network architectures, optimization hyperparameters,
and datasets. IQL learns a value function $V_\psi(s)$, an ensemble of action-value functions
$\{Q_{\theta_i}(s,a)\}_{i=1}^N$, and a policy $\pi_\phi(a \mid s)$ from an offline dataset
$\mathcal{D} = \{(s,a,r,s')\}$. Each critic is trained by minimizing the temporal-difference
loss $\mathcal{L}_Q(\theta_i) = \mathbb{E}_{\mathcal{D}}\!\left[(Q_{\theta_i}(s,a) -
(r + \gamma V_\psi(s')))^2\right]$. The value network is trained via expectile regression
toward the critic ensemble using $\mathcal{L}_V(\psi) =
\mathbb{E}_{\mathcal{D}}\!\left[\rho_\tau(\min_i Q_{\theta_i}(s,a) - V_\psi(s))\right]$,
where $\rho_\tau(u) = |\tau - \mathbb{I}(u < 0)|u^2$. The policy is learned via
advantage-weighted regression with objective
$\mathcal{L}_\pi(\phi) =
\mathbb{E}_{\mathcal{D}}\!\left[\exp(A(s,a)/\beta)\log \pi_\phi(a\mid s)\right]$,
where the advantage is defined as $A(s,a) = \min_i Q_{\theta_i}(s,a) - V_\psi(s)$ and
$\beta$ is the advantage temperature. 

To account for differences in reward sparsity and temporal structure, we sweep the discount
factor $\gamma \in \{0.9, 0.95, 0.99\}$ for each reward setting. All other IQL hyperparameters
and actor--critic architecture details are held fixed across methods and summarized in
\Cref{tab:iql_hyperparameters,tab:transformer_architecture}. 
For each method, we select the best-performing checkpoint based on validation performance.

\paragraph{Reward Specification}
RoboReward provides a sparse, trajectory-level progress signal. To enable its use in IQL,
we convert RoboReward outputs into a per-frame dense reward by querying the model on partial
trajectory prefixes $o_{1:t}$ and assigning the resulting prediction as the reward at timestep
$t$. In contrast, our reward model directly produces dense, temporally aligned rewards over
video sequences, which we use without additional post-processing. Aside from this difference
in reward construction, all other aspects of offline RL training are kept identical across
methods.

\begin{table}[h]
\centering
\small
\renewcommand{\arraystretch}{1.2}
\begin{tabular}{l c}
\toprule
\textbf{Hyperparameter} & \textbf{Value} \\
\midrule
Batch size & 256 \\
Target update rate $\tau$ & 0.005 \\
Discount factor $\gamma$ & $\{0.9, 0.95, 0.99\}$ \\
Target update interval & 1 \\
Advantage temperature & 2 \\
Expectile & 0.7 \\
Advantage clipping & 100.0 \\
Policy extraction method & AWR~\citep{awr} \\
Number of critics & 5 \\
Critics sampled per update & 2 \\
Pooled critic features & True \\
Updates per train step & 1 \\
Actor learning rate & $3\times10^{-4}$ \\
Critic learning rate & $3\times10^{-4}$ \\
Value network learning rate & $3\times10^{-4}$ \\
Weight decay & 0.0 \\
\bottomrule
\end{tabular}
\caption{\textbf{SO-101 offline RL experiment.} Hyperparameters used for Implicit Q-Learning (IQL) in offline policy training.}
\label{tab:iql_hyperparameters}
\end{table}

\begin{table}[h]
\centering
\small
\renewcommand{\arraystretch}{1.2}
\begin{tabular}{l c c}
\toprule
\textbf{Component} & \textbf{Actor} & \textbf{Critic} \\
\midrule
\multicolumn{3}{l}{\textbf{Transformer Encoder}} \\
Model dimension ($d_{\text{model}}$) & 256 & 256 \\
Number of heads & 8 & 8 \\
Encoder layers & 6 & 6 \\
Transformer dropout & 0.0 & 0.0 \\
Transformer activation & GELU & GELU \\
Layer normalization & False & True \\
Pooling strategy & -- & First token \\
\midrule
\multicolumn{3}{l}{\textbf{Feature Processing MLP}} \\
Hidden dimensions & [512, 512] & [768, 512] \\
Activation & ReLU & ReLU \\
Dropout & 0.0 & 0.0 \\
\midrule
\multicolumn{3}{l}{\textbf{Output Head}} \\
Output hidden dims & None & None \\
Action squashing & Tanh & -- \\
Deterministic policy & False & -- \\
Log-std init & 0 & -- \\
Log-std range & $[-20, 2]$ & -- \\
\bottomrule
\end{tabular}
\caption{\textbf{SO-101 offline RL model architectures.} Transformer-based actor and critic architectures used for offline IQL training.
Both networks share a lightweight Transformer encoder, followed by task-specific MLP heads.}
\label{tab:transformer_architecture}
\end{table}

\subsection{Data Filtering \& Retrieval}
\label{sec:app:policy_learning:data-filtering}
\begin{figure}[H]
    \centering
    \includegraphics[width=\linewidth]{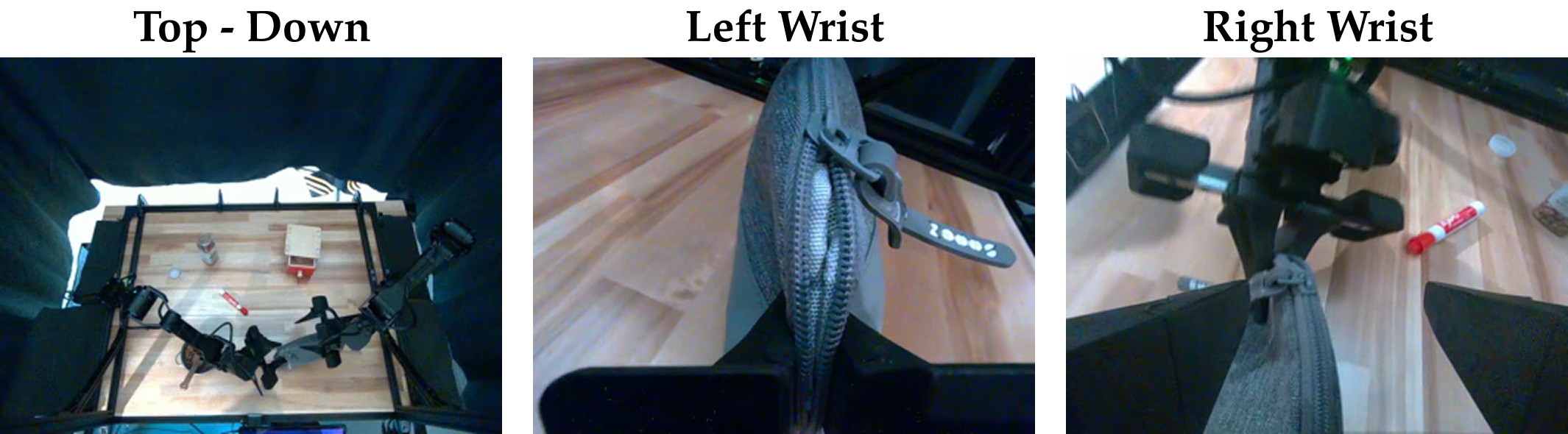}
    \caption{\textbf{Data filtering \& retrieval scene configuration} from all cameras.} 
    \label{fig:appdx:play_dataset_views}
\end{figure}

\paragraph{Procedure and Discussion}
We have 50 trajectories in our play dataset, where each trajectory consists of five tasks executed in random order: \emph{uncap the red pen}, \emph{open the bottle}, \emph{open the red drawer}, \emph{stir the pot}, and \emph{unzip the pencil case}, collected using the Trossen Stationary AI bimanual setup. The dataset includes three camera views: one top-down view and one wrist-mounted camera for each end effector. Example images from all camera views are shown in \Cref{fig:appdx:play_dataset_views}. Following prior work \citep{hong2025hand, memmel2025strap}, we first segment each trajectory into subtrajectories based on end-effector velocities and changes in gripper states. In practice, we found that explicitly incorporating gripper state changes produces cleaner segmentation of different tasks. After segmentation, the play dataset yields a total of 615 subtrajectories. If subtrajectories were uniformly distributed across tasks, this would correspond to 123 relevant segments per task; to ensure conservative evaluation, we retrieve 100 subtrajectories per task.

We compare our method against three baselines: RoboReward, pre-trained SigLIP~\citep{zhai2023siglip}, a vision–language model trained with contrastive image–language supervision, and a retrieval-specific baseline, STRAP~\citep{memmel2025strap}. 
We chose to include STRAP as it is one of the few viable approaches for zero-shot data filtering and retrieval; alternative methods either require policy rollouts \citep{agia2025cupid}, larger amounts of task-specific data \citep{xie2025data}, or training additional task-specific retrieval modules \citep{zhang2026scizor, hejna2025robotdatacurationmutual, du2023behavior, lin2024flowretrieval}.

Given a task language instruction, we perform retrieval using different reward models as follows. For RoboReward and \method{}-Prog, we first compute per-timestep reward predictions for each subtrajectory conditioned on the instruction. We then calculate the value-order correlation of each subtrajectory using the predicted rewards and select the top 100 subtrajectories with the highest VOC scores. When using the preference-based variant of our method (\method{}-Pref), we construct pairwise comparisons between all subtrajectories, aggregate the results into a win matrix, and rank subtrajectories according to estimated pairwise preferences. The top 100-ranked subtrajectories are then selected. For SigLIP, we compute the average vision–language similarity between each subtrajectory and the instruction and select the top 100 subtrajectories with the highest average similarity scores. We use wrist-camera images for retrieval in all methods, as they yield the highest-quality retrievals for all methods.

For $\pi_{0.5}$ policy fine-tuning via LoRA \cite{liu2024tail, liang2022transformer}, we directly use the retrieved 100 subtrajectories together with the corresponding task instruction. During fine-tuning, we include observations from all camera views, as well as states and actions. 
Results reported in \Cref{fig:exp:retrieval_rate}(b) use a strict success metric: a trial is counted as successful only if the robot fully completes the task. This criterion partially explains the low success rates observed for the baselines. Qualitative inspection of learned behaviors reveals clear differences between retrieval methods. Policies trained on SigLIP-retrieved data often learn to approach task-relevant objects (e.g., reaching the drawer) but fail to complete the task. In contrast, policies trained on RoboReward-retrieved data frequently exhibit random or unstable behaviors, consistent with a higher proportion of unrelated subtrajectories in the retrieved set.

\paragraph{Additional Experiments}

We additionally evaluate retrieval quality in a controlled setting where ground-truth trajectory labels are available. Instead of using the play dataset, we consider the Trossen subset of \evaldatasetood, which contains six tasks with known numbers of successful, suboptimal, and failed trajectories. We perform retrieval for two representative tasks—\emph{open the red drawer} and \emph{unzip the pencil case}—each of which contains three successful, two suboptimal, and two failure trajectories. Using the same retrieval procedures described above, we retrieve five trajectories per task and assess their quality. Results are shown in \Cref{tab:appdx:retrieval_quality}. \method{} consistently retrieves higher-quality trajectories, while baselines either fail to retrieve task-relevant trajectories or select failure cases. These results highlight \method{}’s ability to both distinguish between tasks and differentiate levels of execution quality.

\begin{table}[t]
\centering
\setlength{\tabcolsep}{5pt}
\renewcommand{\arraystretch}{1.1}
\begin{adjustbox}{max width=\linewidth}
\begin{tabular}{llcccc}
\toprule
\textbf{Task} & \textbf{Method} & \textbf{\#Succ} & \textbf{\#Subopt} & \textbf{\#Fail} & \textbf{\#Unrel} \\
\midrule
\multirow{5}{*}{Open the red drawer}
& SigLIP           & 2 & 2 & 1 & 0 \\
& RoboReward       & 2 & 1 & 2 & 0 \\
& STRAP            & 2 & 2 & 1 & 0 \\
& \method{}--Prog  & \textbf{3} & \textbf{2} & 0 & 0 \\
& \method{}--Pref  & \textbf{3} & \textbf{2} & 0 & 0 \\
\midrule
\multirow{5}{*}{Unzip the pencil case}
& SigLIP           & 2 & 1 & 2 & 0 \\
& RoboReward       & 0 & 0 & 2 & 3 \\
& STRAP            & 2 & 2 & 1 & 0 \\
& \method{}--Prog  & \textbf{3} & \textbf{2} & 0 & 0 \\
& \method{}--Pref  & 2 & 2 & 1 & 0 \\
\bottomrule
\end{tabular}
\end{adjustbox}
\caption{\textbf{Trajectory retrieval analysis.} Top-5 retrieval quality summary per task. Counts indicate how many of the top-5 retrieved trajectories are labeled as success (Succ), suboptimal (Subopt), failure (Fail), or unrelated (Unrel).}
\label{tab:appdx:retrieval_quality}
\end{table}

\subsection{Failure Detection}
\label{sec:app:policy_learning:ood-failure-detection}

\begin{figure*}[t]
    \centering
    \hfill
    \begin{subfigure}[t]{0.274\linewidth}
        \centering
        \includegraphics[width=\linewidth]{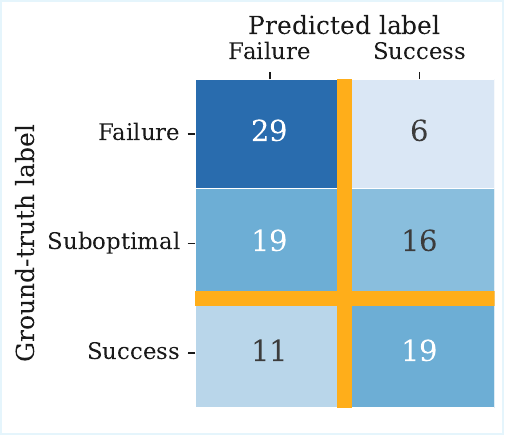}
        \caption{RoboReward-4B}
        \label{fig:ood_confusion_roboreward}
    \end{subfigure}
    \hfill
    \begin{subfigure}[t]{0.274\linewidth}
        \centering
        \includegraphics[width=\linewidth]{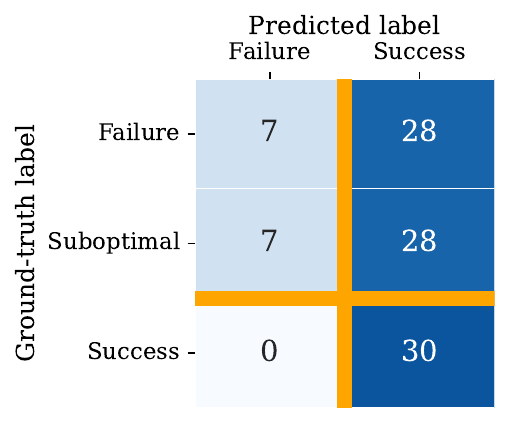}
        \caption{GPT-5-mini}
        \label{fig:ood_confusion_gpt5mini}
    \end{subfigure}
    \hfill
    \begin{subfigure}[t]{0.32\linewidth}
        \centering
        \includegraphics[width=\linewidth]{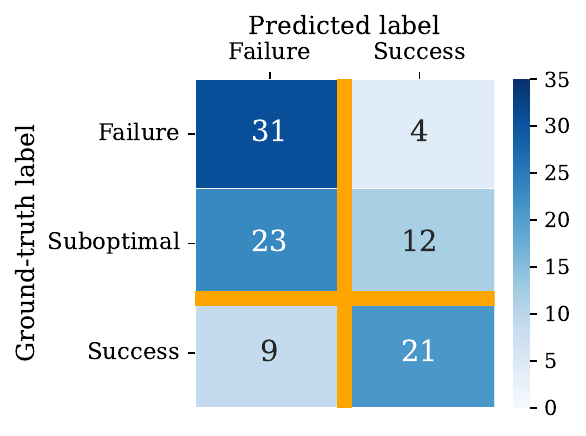}
        \caption{\method (Ours)}
        \label{fig:ood_confusion_ours}
    \end{subfigure}
    \caption{\textbf{Failure Detection OOD confusion matrices} with ternary ground truth and binary prediction.
Rows indicate ground-truth execution outcomes (\emph{failure}, \emph{suboptimal}, \emph{success}), while columns indicate binary predictions (\emph{predicted failure} vs.\ \emph{predicted success}).
\emph{Suboptimal} trajectories correspond to executions that make partial progress but do not complete the task. Suboptimal trajectories are treated as failures in our quantitative evaluation, as emphasized with the horizontal orange divider in the above confusion matrices.
 Color intensity reflects the number of trajectories.}
    \label{fig:ood_confusion_3x2}
\end{figure*}

For failure detection, we evaluate our model on {\mituniv} Franka dataset. For quantitative analysis in Table~\ref{tab:ood_failure_detection_full} and Figure~\ref{fig:ood_confusion_3x2}, we mark a trajectory as success if success probability is higher than or equal to 0.5, and failure based on our progress monitoring strategy explained in this section.

\paragraph{Baselines}
In addition to RoboReward-4B, VLAC, and GPT-5-mini, we compare against an uncertainty-based baseline, following ~\citep{gu2025safe, huang2023look}.
\emph{Token-Uncertainty} baseline estimates predictive uncertainty by computing entropy over generated language tokens. We compute token uncertainty on evaluation trajectories using the $\pi_0$-FAST DROID policy~\citep{pertsch2025fast}.

\paragraph{Discussion}
Figure~\ref{fig:ood_confusion_3x2} visualizes out-of-distribution failure detection using an asymmetric $3{\times}2$ confusion matrix: ground-truth labels include \emph{failure}, \emph{suboptimal}, and \emph{success}, while predictions are binary (\emph{failure} vs.\ \emph{success}).
This view separates two practically distinct error types: (i) missed failures/suboptimal executions (mass in the right column for the top two rows), and (ii) false alarms on successful executions (mass in the left column for the bottom row).
Across methods, \method\ allocates more mass to correctly flagging both \emph{failure} and \emph{suboptimal} trajectories as failures, while maintaining relatively few false alarms on \emph{success} trajectories, consistent with its higher average F1 in Table~\ref{tab:ood_failure_detection}.
In contrast, GPT-5-mini concentrates mass in the \emph{predicted success} column for \emph{failure} and \emph{suboptimal} rows, indicating a conservative bias that avoids false positives but misses many non-success outcomes, whereas RoboReward-4B is intermediate with improved failure sensitivity but still more confusion between \emph{suboptimal} and \emph{success} than \method.

\begin{table*}[!bt]
\vspace{6pt}
\centering
\small
\begin{adjustbox}{max width=\linewidth}
\begin{tabular}{l c c >{\columncolor{lightgray}}c c c >{\columncolor{lightgray}}c c c >{\columncolor{lightgray}}c c c >{\columncolor{lightgray}}c c c >{\columncolor{lightgray}}c }
\toprule
\multirow{2}{*}{Task} 
& \multicolumn{3}{c}{\textbf{Token-Unc.}} 
& \multicolumn{3}{c}{\textbf{VLAC}} 
& \multicolumn{3}{c}{\textbf{GPT-5-mini}} 
& \multicolumn{3}{c}
{\textbf{\roborewardcolor{RoboReward}-4B}} 
& \multicolumn{3}{c}{\textbf{\robometercolor{\method}}} \\
\cmidrule(lr){2-4} \cmidrule(lr){5-7} \cmidrule(lr){8-10}
& \textbf{TPR} & \textbf{TNR} & \textbf{F1} 
& \textbf{TPR} & \textbf{TNR} & \textbf{F1} 
& \textbf{TPR} & \textbf{TNR} & \textbf{F1} 
& \textbf{TPR} & \textbf{TNR} & \textbf{F1} 
& \textbf{TPR} & \textbf{TNR} & \textbf{F1}  \\
\midrule
move banana  & 0.00 & 1.00 & 0.53 & 0.89 & 0.35 & 0.45 & 0.31 & 1.00 & 0.48 & 1.00 & 0.62 & 0.91 & 1.00 & 0.75 & \textbf{0.94} \\
move mouse   & 0.00 & 1.00 & 0.50 & 1.00 & 0.00 & 0.00 & 0.80 & 1.00 & 0.89 & 0.80 & 0.75 & 0.80 & 1.00 & 0.75 & \textbf{0.91} \\
pour pebble  & 0.00 & 1.00 & 0.32 & 0.88 & 0.00 & 0.00 & 0.14 & 1.00 & 0.25 & 0.57 & 1.00 & 0.73 & 0.71 & 1.00 & \textbf{0.83} \\
fold towel   & 0.00 & 1.00 & 0.58 & 0.95 & 0.10 & 0.16 & 0.15 & 1.00 & 0.27 & 0.31 & 0.67 & 0.40 & 0.54 & 0.56 & \textbf{0.58} \\
pull tissue  & 0.00 & 1.00 & 0.43 & 1.00 & 0.00 & 0.00 & 0.00 & 1.00 & 0.00 & 0.55 & 0.00 & 0.57 & 0.73 & 0.50 & \textbf{0.76} \\
put spoon    & 0.00 & 1.00 & 0.22 & 1.00 & 0.00 & 0.00 & 0.14 & 1.00 & 0.25 & 0.57 & 1.00 & \textbf{0.73} & 0.57 & 1.00 & \textbf{0.73} \\
stir pot     & 0.00 & 1.00 & 0.47 & 0.94 & 0.00 & 0.00 & 0.09 & 1.00 & 0.17 & 0.91 & 1.00 & \textbf{0.95} & 0.82 & 1.00 & 0.90 \\
\midrule
\textbf{Average} 
& 0.00 & 1.00 & 0.48 & 0.95 & 0.10 & 0.16 & 0.20 & 1.00 & 0.33 & 0.69 & 0.63 & 0.74 & 0.77 & 0.70 & \textbf{0.81} \\
\bottomrule
\end{tabular}
\end{adjustbox}
\caption{
\textbf{Failure detection performance}.
We report true positive rate (TPR; correctly detecting failures), true negative rate (TNR; correctly identifying successful executions), and F1 score. 
}
\label{tab:ood_failure_detection_full}
\end{table*}

\paragraph{Failure Detection with Progress Monitoring}
Table~\ref{tab:ood_failure_detection_full} compares our model against baselines on three evaluation metrics:  true positive rate (TPR; correctly detecting failures), true negative rate (TNR; correctly identifying successful executions), and F1 score. \method{} consistently outperforms baselines, resulting in higher average F1 score. Token-Uncertainty and GPT-5-mini baselines attain perfect TNR but extremely low TPR, indicating a strong bias toward predicting success and missing failures. VLAC exhibits the opposite behavior: it frequently flags trajectories as failures, achieving high true positive rates but low true negative rates due to many false positives on successful executions, which results in lower F1 scores.
RoboReward-4B lies between these extremes, with improved failure sensitivity over GPT-5-mini but more confusion between \emph{suboptimal} and \emph{success} trajectories than \method.

\begin{figure*}[t]
    \centering
    \captionsetup{font=small}

    \begin{subfigure}[t]{0.49\textwidth}
        \centering
        \includegraphics[width=\linewidth]{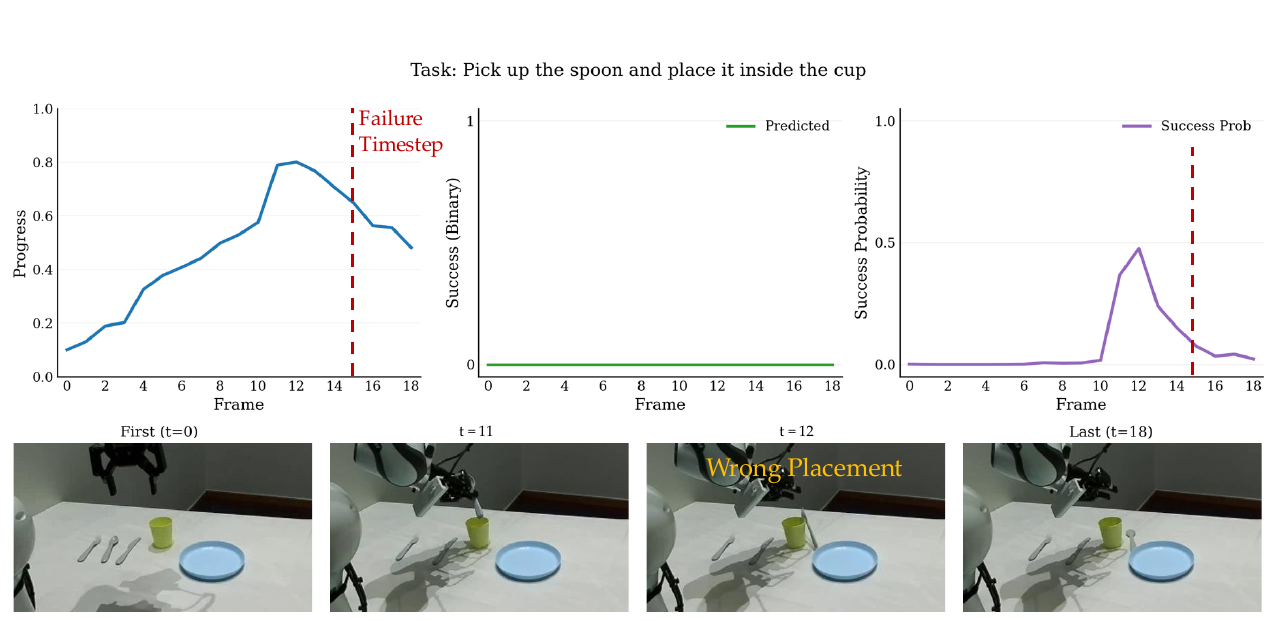}
        \caption{Dropped object during placement.}
    \end{subfigure}
    \hfill
    \begin{subfigure}[t]{0.49\textwidth}
        \centering
        \includegraphics[width=\linewidth]{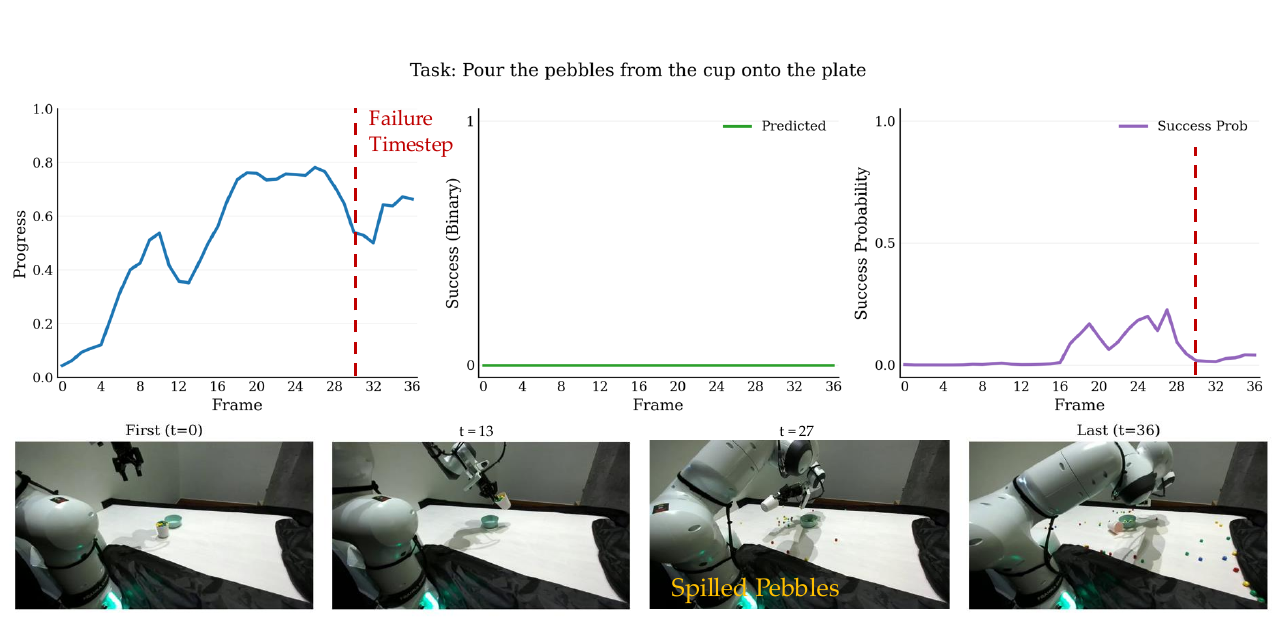}
        \caption{Spilled contents during pouring.}
    \end{subfigure}

    \caption{\textbf{Irreversible failures.}
    Terminal events such as drops or spills cause a sharp regression in predicted task progress, which our model reliably flags as failures shortly after the event.}
    \label{fig:failure_irreversible}
\end{figure*}

We provide qualitative examples of failure detection using progress monitoring across different failure categories in Figures~\ref{fig:failure_irreversible}--\ref{fig:failure_insufficient}.
Specifically, we compute the Pearson correlation between progress values and time over a sliding window, and flag a failure at the first timestep where this correlation becomes lower than a threshold. This captures both irreversible failures, where rewards sharply decrease after an error (e.g., object drops), and insufficient-progress failures, where rewards stagnate or regress over time. If no such failure is detected or if our model predicts success, the trajectory is classified as successful.
Our failure evaluation dataset include both \emph{irreversible failures} (e.g., object drops or spills) and \emph{insufficient-progress failures}, where the robot stalls, oscillates, or terminates execution before completing the task.
We additionally highlight \emph{semantic failures}, where the robot executes a physically plausible behavior that violates the task instruction. All qualitative examples use window size of 5 besides Figure~\ref{fig:failure_irreversible} (b) which uses window size of 9. We use correlation threshold of -0.5 in all examples.

\begin{figure*}[t]
    \centering
    \captionsetup{font=small}

    \begin{subfigure}[t]{0.49\textwidth}
        \centering
        \includegraphics[width=\linewidth]{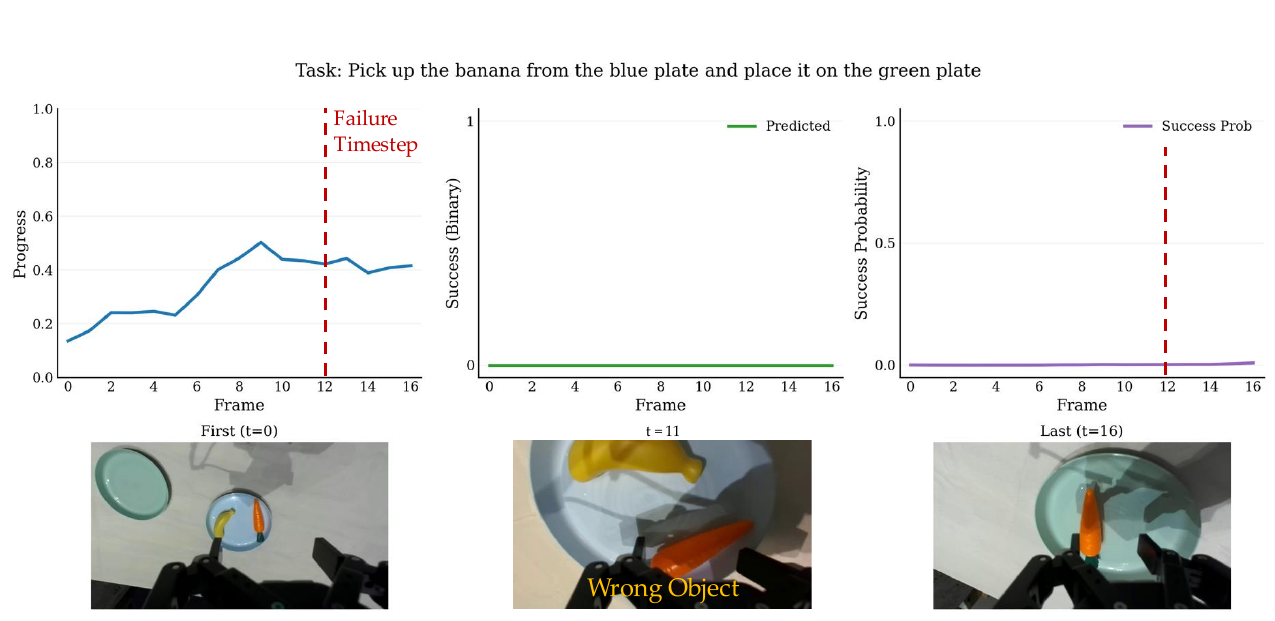}
        \caption{Wrong object grasped despite correct motion pattern.}
    \end{subfigure}
    \hfill
    \begin{subfigure}[t]{0.49\textwidth}
        \centering
        \includegraphics[width=\linewidth]{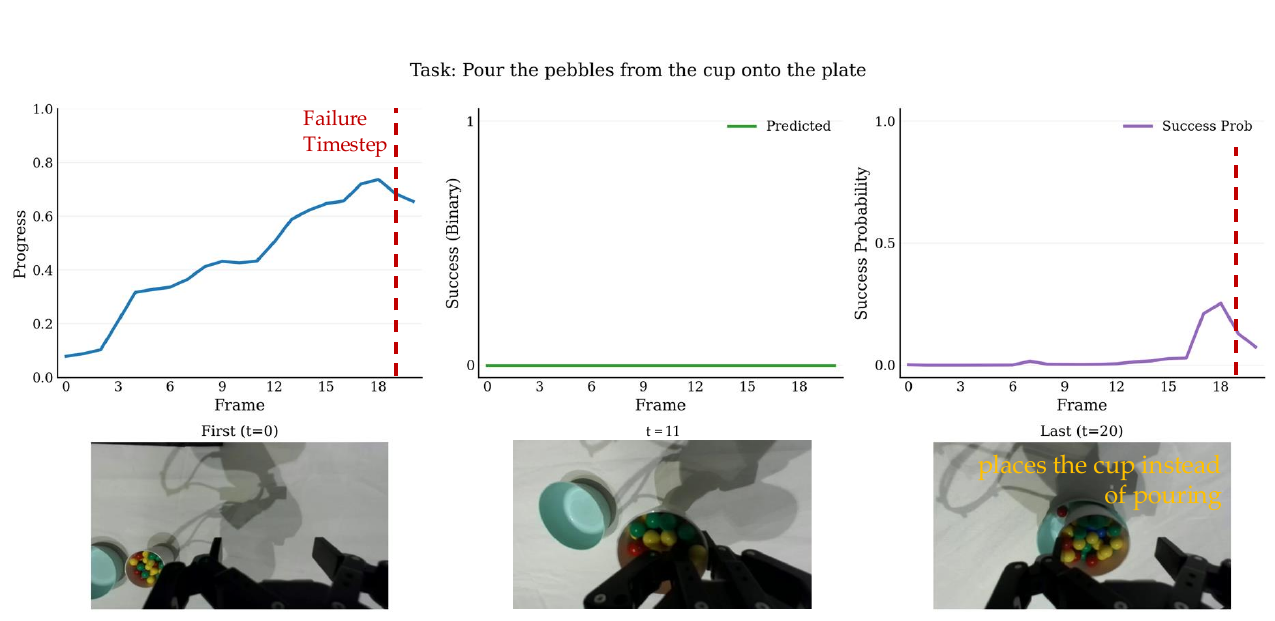}
        \caption{Instruction-inconsistent action (placing the cup instead of pouring).}
    \end{subfigure}

    \caption{\textbf{Semantic failures.}
    The robot executes smooth and physically plausible trajectories but violates the task instruction, resulting in persistently low predicted progress and failure detection without an abrupt terminal event.}
    \label{fig:failure_semantic}
\end{figure*}

Across all cases, our reward model identifies failures by detecting regressions or stagnation in predicted task progress.
For irreversible failures (see Figure~\ref{fig:failure_irreversible}), the predicted progress increases initially but drops sharply after the terminal event (e.g., dropping or spilling), leading to early failure detection.
In semantic failures (see Figure~\ref{fig:failure_semantic}), progress remains consistently low despite smooth execution, reflecting instruction-level mismatch.
For insufficient-progress failures (see Figure~\ref{fig:failure_insufficient}), the progress signal plateaus or oscillates without converging to success, allowing the model to flag failure even in the absence of abrupt terminal events.

\section{Low-rank Fine-tuning \method-4B}
\label{sec:appdx:robofac_lora_finetune}

\textbf{RoboFAC dataset.} We fine-tune and evaluate on RoboFAC, a video-based dataset designed for robotic failure analysis and correction. 
RoboFAC spans 16 tasks across 53 scenes and provides 78K video QA pairs annotated over 10,722 trajectories. The dataset contains both failures and successes: 9,440 failure trajectories and 1,282 successful trajectories. Failures are collected in both simulation and the real world, with 8,960 simulated failures and 480 real-world failures; additionally, the dataset includes 1{,}160 simulated successes and 122 real-world successes.

\textbf{LoRA fine-tuning Robometer-4B.}
We keep our pre-trained Robometer-4B reward model backbone fixed and train LLM LoRA adapters (76M parameters) while also fine-tuning the MLP prediction heads (progress, preference, and success). We train for 500 steps with a batch size of 8 on a single NVIDIA RTX A6000 for approximately 8 hours of wall clock time.

\textbf{Baselines.}
To isolate the benefit of initializing from our pre-trained \method-4B checkpoint versus starting from scratch, we include two Qwen3-VL baselines fine-tuned on the same dataset. 
First, we perform LoRA fine-tuning starting from \texttt{Qwen/Qwen3-VL-4B-Instruct} and train the prediction heads from random initialization. We also fully fine-tune all parameters of \texttt{Qwen/Qwen3-VL-4B-Instruct}), again training the prediction heads from random initialization. 

\textbf{Results.}
\Cref{tab:robofac_finetune_results} reports offline reward evaluation on RoboFAC. In the zero-shot setting, \method-4B already achieves strong correlation with ground-truth progress (VOC $r{=}0.652$, Kendall $\tau{=}0.436$), indicating meaningful transfer without any RoboFAC supervision. After adaptation, \method-4B provides a substantially better initialization than training the base VLM from scratch.
Full fine-tuning from our checkpoint reaches VOC $r{=}0.884$ and Kendall $\tau{=}0.802$, outperforming the strongest from-scratch baseline by $+21.6\%$ in VOC and nearly an order of magnitude in Kendall $\tau$. Notably, LoRA on \method-4B attains essentially the same performance, indicating that the gains primarily come from our pre-trained initialization rather than full-parameter adaptation.

\section{\method\ with Model-Based RL}
\label{sec:appdx:modelbasedrl}
\begin{figure}[ht]
    \centering
    \includegraphics[width=\linewidth]{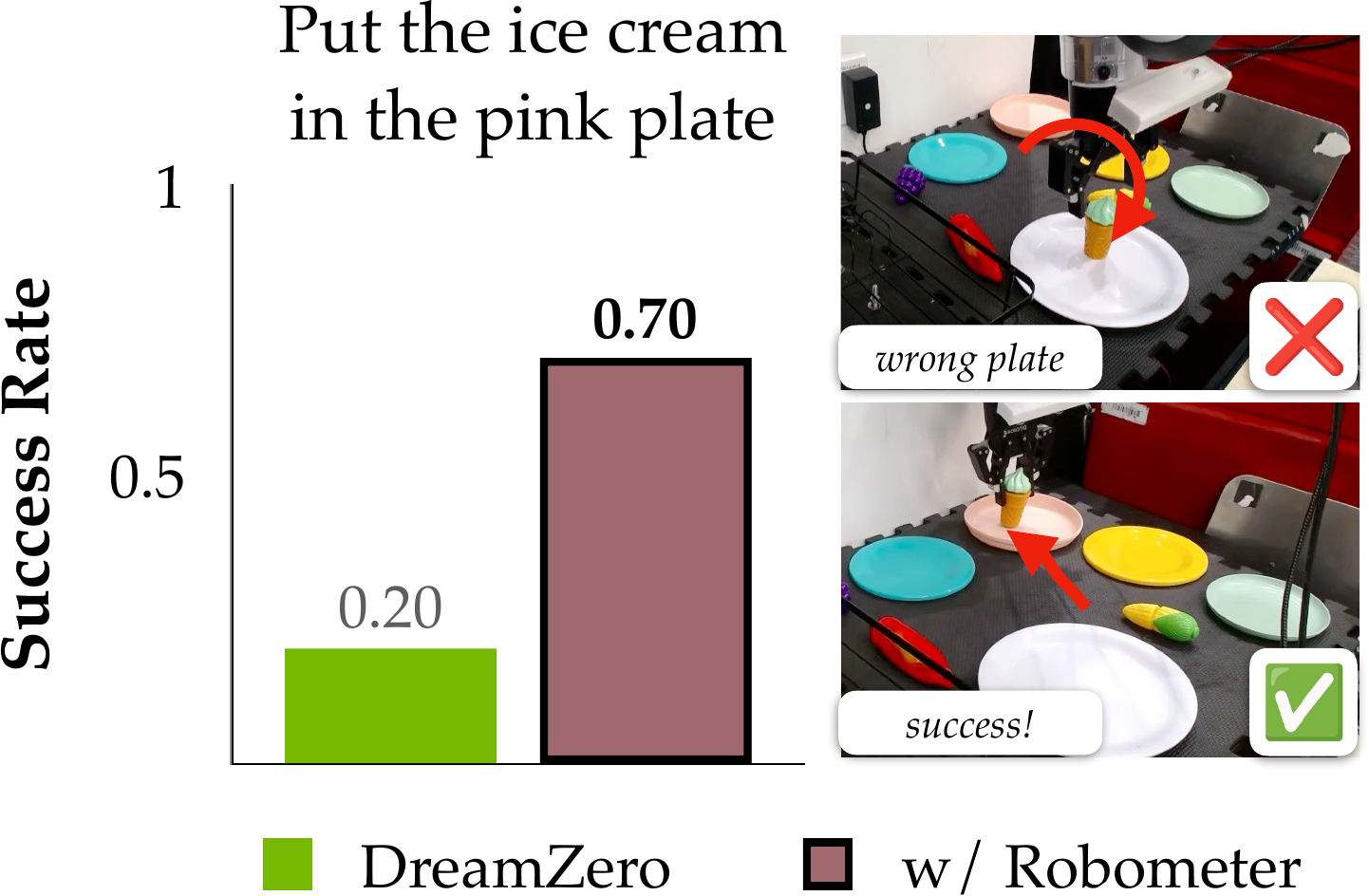}
    \caption{\textbf{Model-Based RL} with \method\ integrated into DreamZero~\citep{ye2026dreamzero}. 
    In this cluttered scene, \method\ improves DreamZero's performance from 20\% success rate to 70\%.
    }
    \label{fig:appdx:dreamzero_exp}
\end{figure}

\textbf{Setup.} For this proof-of-concept experiment, we integrate \method\ into a model-based algorithm by using it to rank candidate trajectories sampled from the world model.

We adopt the DreamZero~\citep{ye2026dreamzero} world model checkpoint trained on DROID to use in an identical DROID setup to the online RL experiments, albeit with a different task.

Given 3 camera observations, proprioceptive states, and a language instruction, DreamZero simultaneously generates candidate future observation sequences (multiple video frames spanning 1.6 seconds) with associated action chunks of length 24.

We compare DreamZero with and without \method\ ranking candidate trajectories. 
DreamZero without \method\ directly generates a single candidate future observation sequence from which we extract the action chunk. 
With \method, we generate 6 candidate observation sequences at each inference step, rank them with the progress output from \method, and then execute the action sequence extracted from the highest-ranked generated observation sequence.

In total, generating a single action chunk of length 24, including the world model forward pass for the 6 candidate observation sequences, takes approximately 28 seconds on 1 H200, resulting in a single trajectory taking around 3 minutes to execute including robot execution time.
The vast majority of this inference time is from the world model; \method's forward pass only takes around 0.6-1 seconds.

\textbf{Results.}
We evaluate on a ``put the ice cream in the pink plate'' task in an extremely cluttered scene with many possible receptacle plates of different colors.
DreamZero typically places the ice cream cone in the wrong plate, while integrating \method\ corrects for this mistake.
Overall, results in \Cref{fig:appdx:dreamzero_exp} demonstrate a \textbf{$3.5\times$} improvement in success rate evaluated over 10 trials with \method.

%% file: refs.bib
@IEEEtranBSTCTL{max8authors,
  CTLuse_forced_etal       = "yes",
  CTLmax_names_forced_etal = "8",
  CTLnames_show_etal       = "8"
}

@string{CVPR = "IEEE Conference on Computer Vision and Pattern Recognition"}

@string{ICCV = "International Conference on Computer Vision"}

@string{NeurIPS = "Neural Information Processing Systems"}

@string{ICLR = "International Conference on Learning Representations"}

@string{ICML = "International Conference on Machine Learning"}

@string{AAAI = "Association for the Advancement of Artificial Intelligence"}

@string{ICRA = "International Conference on Robotics and Automation"}

@string{CoRL = "Conference on Robot Learning"}

@inproceedings{yang2024rank,
  author    = {Yang, Daniel and Tjia, Davin and Berg, Jacob and Damen, Dima and Agrawal, Pulkit and Gupta, Abhishek},
  title     = {Rank2Reward: Learning Shaped Reward Functions from Passive Video},
  booktitle   = {International Conference on Robotics and Automation (ICRA)},
  year      = {2024},
}

@inproceedings{roboclip,
    author = {Sumedh Anand Sontakke and Jesse Zhang and Séb Arnold and Karl Pertsch and Erdem Biyik and Dorsa Sadigh and Chelsea Finn and Laurent Itti},
    title = {RoboCLIP: One Demonstration is Enough to Learn Robot Policies},
    booktitle = {Advances in Neural Information Processing Systems (NeurIPS)},
    year = {2023},
}

@inproceedings{ma2024generative,
    author    = {Ma, Yecheng Jason and Hejna, Joey and Wahid, Ayzaan and Fu, Chuyuan and Shah, Dhruv and Liang, Jacky and Xu, Zhuo and Kirmani, Sean and Xu, Peng and Driess, Danny and Xiao, Ted and Tompson, Jonathan and Bastani, Osbert and Jayaraman, Dinesh and Yu, Wenhao and Zhang, Tingnan and Sadigh, Dorsa and Xia, Fei},
    title     = {Vision Language Models are In-Context Value Learners},
    booktitle= {International Conference on Learning Representations (ICLR)},
    year      = {2025}
}

@inproceedings{ma2023liv,
    title={LIV: Language-Image Representations and Rewards for Robotic Control},
    author={Ma, Yecheng Jason and Liang, William and Som, Vaidehi and Kumar, Vikash and Zhang, Amy and Bastani, Osbert and Jayaraman, Dinesh},
    booktitle={International Conference on Machine Learning (ICML)},
    year={2023}
}

@inproceedings{ma2023vipuniversalvisualreward,
      title={VIP: Towards Universal Visual Reward and Representation via Value-Implicit Pre-Training}, 
      author={Yecheng Jason Ma and Shagun Sodhani and Dinesh Jayaraman and Osbert Bastani and Vikash Kumar and Amy Zhang},
      year={2023},
    booktitle= {International Conference on Learning Representations (ICLR)},
}

@inproceedings{yu2021metaworldbenchmarkevaluationmultitask,
      title={Meta-World: A Benchmark and Evaluation for Multi-Task and Meta Reinforcement Learning}, 
      author={Tianhe Yu and Deirdre Quillen and Zhanpeng He and Ryan Julian and Avnish Narayan and Hayden Shively and Adithya Bellathur and Karol Hausman and Chelsea Finn and Sergey Levine},
      year={2019},
      booktitle={Conference on Robot Learning (CoRL)},
}

@inproceedings{brohan2022rt1,
      title={RT-1: Robotics Transformer for Real-World Control at Scale}, 
      author={Anthony Brohan and Noah Brown and Justice Carbajal and Yevgen Chebotar and Joseph Dabis and Chelsea Finn and Keerthana Gopalakrishnan and Karol Hausman and Alex Herzog and Jasmine Hsu and Julian Ibarz and Brian Ichter and Alex Irpan and Tomas Jackson and Sally Jesmonth and Nikhil J Joshi and Ryan Julian and Dmitry Kalashnikov and Yuheng Kuang and Isabel Leal and Kuang-Huei Lee and Sergey Levine and Yao Lu and Utsav Malla and Deeksha Manjunath and Igor Mordatch and Ofir Nachum and Carolina Parada and Jodilyn Peralta and Emily Perez and Karl Pertsch and Jornell Quiambao and Kanishka Rao and Michael Ryoo and Grecia Salazar and Pannag Sanketi and Kevin Sayed and Jaspiar Singh and Sumedh Sontakke and Austin Stone and Clayton Tan and Huong Tran and Vincent Vanhoucke and Steve Vega and Quan Vuong and Fei Xia and Ted Xiao and Peng Xu and Sichun Xu and Tianhe Yu and Brianna Zitkovich},
      year={2023},
      booktitle={Robotics: Science and Systems (RSS)},
}

@article{black2024pi0visionlanguageactionflowmodel,
      title={$\pi_0$: A Vision-Language-Action Flow Model for General Robot Control}, 
      author={Kevin Black and Noah Brown and Danny Driess and Adnan Esmail and Michael Equi and Chelsea Finn and Niccolo Fusai and Lachy Groom and Karol Hausman and Brian Ichter and Szymon Jakubczak and Tim Jones and Liyiming Ke and Sergey Levine and Adrian Li-Bell and Mohith Mothukuri and Suraj Nair and Karl Pertsch and Lucy Xiaoyang Shi and James Tanner and Quan Vuong and Anna Walling and Haohuan Wang and Ury Zhilinsky},
      year={2024},
      journal={arXiv preprint arxiv:2410.24164},
}

@inproceedings{hamster2025,
    author = {Yi Li and Yuquan Deng and Jesse Zhang and Joel Jang and Marius Memmel and Caelan Reed Garrett and Fabio Ramos and Dieter Fox and Anqi Li and Abhishek Gupta and Ankit Goyal},
    title = {HAMSTER: Hierarchical Action Models for Open-World Robot Manipulation},
    booktitle={International Conference on Learning Representations (ICLR)},
    year={2025}
}

@inproceedings{korkmaz2025mile,
 title={MILE: Model-based Intervention Learning},
 author={Yigit Korkmaz and Erdem Bıyık},
 booktitle={International Conference on Robotics and Automation (ICRA)},
 year={2025}
}

@inproceedings{kwon2023reward,
 title={Reward Design with Language Models},
 author={Kwon, Minae and Xie, Sang Michael and Bullard, Kalesha and Sadigh, Dorsa},
 booktitle={International Conference on Learning Representations (ICLR)},
 year={2023}
}

@inproceedings{ma2023eureka,
    title   = {Eureka: Human-Level Reward Design via Coding Large Language Models},
    author  = {Yecheng Jason Ma and William Liang and Guanzhi Wang and De-An Huang and Osbert Bastani and Dinesh Jayaraman and Yuke Zhu and Linxi Fan and Anima Anandkumar},
    year    = {2024},
    booktitle={International Conference on Learning Representations (ICLR)},
}

@inproceedings{yu2023language,
    title={Language to Rewards for Robotic Skill Synthesis},
    author={Yu, Wenhao and Gileadi, Nimrod and Fu, Chuyuan and Kirmani, Sean and Lee, Kuang-Huei and Gonzalez Arenas, Montse and Lewis Chiang, Hao-Tien and Erez, Tom and Hasenclever, Leonard and Humplik, Jan and Ichter, Brian and Xiao, Ted and Xu, Peng and Zeng, Andy and Zhang, Tingnan and Heess, Nicolas and Sadigh, Dorsa and Tan, Jie and Tassa, Yuval and Xia, Fei},
    year={2023},
    booktitle={Conference on Robot Learning (CoRL)},
}

@inproceedings{mahmoudieh2022zero,
    title = {Zero-Shot Reward Specification via Grounded Natural Language},
    author = {Mahmoudieh, Parsa and Pathak, Deepak and Darrell, Trevor},
    booktitle = {International Conference on Machine Learning (ICML)},
    year = {2022},
}

@inproceedings{ng2000IRL,
author = {Ng, Andrew Y. and Russell, Stuart J.},
title = {Algorithms for Inverse Reinforcement Learning},
year = {2000},
booktitle = {International Conference on Machine Learning (ICML)},
}

@inproceedings{abbeell2004IRL,
author = {Abbeel, Pieter and Ng, Andrew Y.},
title = {Apprenticeship Learning via Inverse Reinforcement Learning},
year = {2004},
booktitle = {International Conference on Machine Learning (ICML)},
}

@inproceedings{ziebart2008maxentirl,
author = {Ziebart, Brian D. and Maas, Andrew and Bagnell, J. Andrew and Dey, Anind K.},
title = {Maximum Entropy Inverse Reinforcement Learning},
year = {2008},
booktitle = {AAAI Conference on Artificial Intelligence},
}

@inproceedings{finn2016gcl,
author = {Finn, Chelsea and Levine, Sergey and Abbeel, Pieter},
title = {Guided Cost Learning: Deep Inverse Optimal Control via Policy Optimization},
year = {2016},
booktitle = {International Conference on Machine Learning (ICML)},
}

@inproceedings{ho2016generative,
  title={Generative adversarial imitation learning},
  author={Ho, Jonathan and Ermon, Stefano},
  booktitle={Advances in Neural Information Processing Systems (NeurIPS)},
  year={2016}
}

@inproceedings{fu2017learning,
  title={Learning robust rewards with adversarial inverse reinforcement learning},
  author={Fu, Justin and Luo, Katie and Levine, Sergey},
  booktitle={International Conference on Learning Representations (ICLR)},
  year={2018}
}

@inproceedings{fu2018VICE,
	author = {Fu, Justin and Singh, Avi and Ghosh, Dibya and Yang, Larry and Levine, Sergey},
	booktitle = {Advances in Neural Information Processing Systems (NeurIPS)},
	title = {Variational Inverse Control with Events: A General Framework for Data-Driven Reward Definition},
	year = {2018}
 }

@inproceedings{sadigh2017active,
 author = {Sadigh, Dorsa and Dragan, Anca D. and Sastry, S. Shankar and Seshia, Sanjit A.},
 title = {Active Preference-Based Learning of Reward Functions},
 booktitle = {Robotics: Science and Systems (RSS)},
 year = {2017},
}

@inproceedings{myers2021learning,
title={Learning Multimodal Rewards from Rankings},
author={Vivek Myers and Erdem Biyik and Nima Anari and Dorsa Sadigh},
booktitle={Conference on Robot Learning (CoRL)},
year={2021},
}

@inproceedings{lee2021pebble,
title={PEBBLE: Feedback-Efficient Interactive Reinforcement Learning via Relabeling Experience and Unsupervised Pre-training}, 
author={Kimin Lee and Laura Smith and Pieter Abbeel},
year={2021},
booktitle={International Conference on Machine Learning (ICML)}
}

@inproceedings{yang2024trajectory,
 title={Trajectory Improvement and Reward Learning from Comparative Language Feedback},
 author={Zhaojing Yang and Miru Jun and Jeremy Tien and Stuart J. Russell and Anca Dragan and Erdem Bıyık},
 booktitle={Conference on Robot Learning (CoRL)},
 year={2024}
}

@inproceedings{bajcsy2018learning,
  title={Learning from physical human corrections, one feature at a time},
  author={Bajcsy, Andrea and Losey, Dylan P and O'Malley, Marcia K and Dragan, Anca D},
  booktitle={International Conference on Human-Robot Interaction (HRI)},
  year={2018}
}

@inproceedings{hejna2022fewshot,
 title={Few-Shot Preference Learning for Human-in-the-Loop RL},
 author={Hejna, Joey and Sadigh, Dorsa},
 booktitle={Conference on Robot Learning (CoRL)},
 year={2022}
}

@inproceedings{biyik2020active,
 title={Active Preference-Based Gaussian Process Regression for Reward Learning},
 author={Biyik, Erdem and Huynh, Nicolas and Kochenderfer, Mykel J. and Sadigh, Dorsa},
 booktitle={Robotics: Science and Systems (RSS)},
 year={2020},
}

@inproceedings{christiano2017rlhf,
author = {Christiano, Paul F. and Leike, Jan and Brown, Tom B. and Martic, Miljan and Legg, Shane and Amodei, Dario},
title = {Deep reinforcement learning from human preferences},
year = {2017},
booktitle = {Advances in Neural Information Processing Systems (NeurIPS)},
}

@InProceedings{wangrlvlmf2024,
          title = 	 {RL-VLM-F: Reinforcement Learning from Vision Language Foundation Model Feedback},
          author =       {Wang, Yufei and Sun, Zhanyi and Zhang, Jesse and Xian, Zhou and Biyik, Erdem and Held, David and Erickson, Zackory},
          booktitle = 	 {International Conference on Machine Learning (ICML)},
          year = 	 {2024}
        }

@article{geminiteam2024geminifamilyhighlycapable,
      title={Gemini: A Family of Highly Capable Multimodal Models}, 
      author={Gemini Team},
      year={2024},
      journal={arXiv preprint arXiv:2312.11805},
}

@inproceedings{
kostrikov2022offline,
title={Offline Reinforcement Learning with Implicit Q-Learning},
author={Ilya Kostrikov and Ashvin Nair and Sergey Levine},
booktitle={International Conference on Learning Representations (ICLR)},
year={2022},
}

@inproceedings{awr,
  author = {Peng, Xue Bin and Kumar, Aviral and Zhang, Grace and Levine, Sergey},
  title = {Advantage-Weighted Regression: Simple and Scalable Off-Policy Reinforcement Learning},
  booktitle = {arXiv preprint arXiv:1910.00177},
  year = {2019},
}

@inproceedings{open_x_embodiment_rt_x_2023,
title={Open {X-E}mbodiment: Robotic Learning Datasets and {RT-X} Models},
author = {Open X-Embodiment Collaboration and Abby O'Neill and Abdul Rehman and Abhinav Gupta and Abhiram Maddukuri and Abhishek Gupta and Abhishek Padalkar and Abraham Lee and Acorn Pooley and Agrim Gupta and Ajay Mandlekar and Ajinkya Jain and Albert Tung and Alex Bewley and Alex Herzog and Alex Irpan and Alexander Khazatsky and Anant Rai and Anchit Gupta and Andrew Wang and Andrey Kolobov and Anikait Singh and Animesh Garg and Aniruddha Kembhavi and Annie Xie and Anthony Brohan and Antonin Raffin and Archit Sharma and Arefeh Yavary and Arhan Jain and Ashwin Balakrishna and Ayzaan Wahid and Ben Burgess-Limerick and Beomjoon Kim and Bernhard Schölkopf and Blake Wulfe and Brian Ichter and Cewu Lu and Charles Xu and Charlotte Le and Chelsea Finn and Chen Wang and Chenfeng Xu and Cheng Chi and Chenguang Huang and Christine Chan and Christopher Agia and Chuer Pan and Chuyuan Fu and Coline Devin and Danfei Xu and Daniel Morton and Danny Driess and Daphne Chen and Deepak Pathak and Dhruv Shah and Dieter Büchler and Dinesh Jayaraman and Dmitry Kalashnikov and Dorsa Sadigh and Edward Johns and Ethan Foster and Fangchen Liu and Federico Ceola and Fei Xia and Feiyu Zhao and Felipe Vieira Frujeri and Freek Stulp and Gaoyue Zhou and Gaurav S. Sukhatme and Gautam Salhotra and Ge Yan and Gilbert Feng and Giulio Schiavi and Glen Berseth and Gregory Kahn and Guangwen Yang and Guanzhi Wang and Hao Su and Hao-Shu Fang and Haochen Shi and Henghui Bao and Heni Ben Amor and Henrik I Christensen and Hiroki Furuta and Homanga Bharadhwaj and Homer Walke and Hongjie Fang and Huy Ha and Igor Mordatch and Ilija Radosavovic and Isabel Leal and Jacky Liang and Jad Abou-Chakra and Jaehyung Kim and Jaimyn Drake and Jan Peters and Jan Schneider and Jasmine Hsu and Jay Vakil and Jeannette Bohg and Jeffrey Bingham and Jeffrey Wu and Jensen Gao and Jiaheng Hu and Jiajun Wu and Jialin Wu and Jiankai Sun and Jianlan Luo and Jiayuan Gu and Jie Tan and Jihoon Oh and Jimmy Wu and Jingpei Lu and Jingyun Yang and Jitendra Malik and João Silvério and Joey Hejna and Jonathan Booher and Jonathan Tompson and Jonathan Yang and Jordi Salvador and Joseph J. Lim and Junhyek Han and Kaiyuan Wang and Kanishka Rao and Karl Pertsch and Karol Hausman and Keegan Go and Keerthana Gopalakrishnan and Ken Goldberg and Kendra Byrne and Kenneth Oslund and Kento Kawaharazuka and Kevin Black and Kevin Lin and Kevin Zhang and Kiana Ehsani and Kiran Lekkala and Kirsty Ellis and Krishan Rana and Krishnan Srinivasan and Kuan Fang and Kunal Pratap Singh and Kuo-Hao Zeng and Kyle Hatch and Kyle Hsu and Laurent Itti and Lawrence Yunliang Chen and Lerrel Pinto and Li Fei-Fei and Liam Tan and Linxi "Jim" Fan and Lionel Ott and Lisa Lee and Luca Weihs and Magnum Chen and Marion Lepert and Marius Memmel and Masayoshi Tomizuka and Masha Itkina and Mateo Guaman Castro and Max Spero and Maximilian Du and Michael Ahn and Michael C. Yip and Mingtong Zhang and Mingyu Ding and Minho Heo and Mohan Kumar Srirama and Mohit Sharma and Moo Jin Kim and Naoaki Kanazawa and Nicklas Hansen and Nicolas Heess and Nikhil J Joshi and Niko Suenderhauf and Ning Liu and Norman Di Palo and Nur Muhammad Mahi Shafiullah and Oier Mees and Oliver Kroemer and Osbert Bastani and Pannag R Sanketi and Patrick "Tree" Miller and Patrick Yin and Paul Wohlhart and Peng Xu and Peter David Fagan and Peter Mitrano and Pierre Sermanet and Pieter Abbeel and Priya Sundaresan and Qiuyu Chen and Quan Vuong and Rafael Rafailov and Ran Tian and Ria Doshi and Roberto Mart{'i}n-Mart{'i}n and Rohan Baijal and Rosario Scalise and Rose Hendrix and Roy Lin and Runjia Qian and Ruohan Zhang and Russell Mendonca and Rutav Shah and Ryan Hoque and Ryan Julian and Samuel Bustamante and Sean Kirmani and Sergey Levine and Shan Lin and Sherry Moore and Shikhar Bahl and Shivin Dass and Shubham Sonawani and Shubham Tulsiani and Shuran Song and Sichun Xu and Siddhant Haldar and Siddharth Karamcheti and Simeon Adebola and Simon Guist and Soroush Nasiriany and Stefan Schaal and Stefan Welker and Stephen Tian and Subramanian Ramamoorthy and Sudeep Dasari and Suneel Belkhale and Sungjae Park and Suraj Nair and Suvir Mirchandani and Takayuki Osa and Tanmay Gupta and Tatsuya Harada and Tatsuya Matsushima and Ted Xiao and Thomas Kollar and Tianhe Yu and Tianli Ding and Todor Davchev and Tony Z. Zhao and Travis Armstrong and Trevor Darrell and Trinity Chung and Vidhi Jain and Vikash Kumar and Vincent Vanhoucke and Wei Zhan and Wenxuan Zhou and Wolfram Burgard and Xi Chen and Xiangyu Chen and Xiaolong Wang and Xinghao Zhu and Xinyang Geng and Xiyuan Liu and Xu Liangwei and Xuanlin Li and Yansong Pang and Yao Lu and Yecheng Jason Ma and Yejin Kim and Yevgen Chebotar and Yifan Zhou and Yifeng Zhu and Yilin Wu and Ying Xu and Yixuan Wang and Yonatan Bisk and Yongqiang Dou and Yoonyoung Cho and Youngwoon Lee and Yuchen Cui and Yue Cao and Yueh-Hua Wu and Yujin Tang and Yuke Zhu and Yunchu Zhang and Yunfan Jiang and Yunshuang Li and Yunzhu Li and Yusuke Iwasawa and Yutaka Matsuo and Zehan Ma and Zhuo Xu and Zichen Jeff Cui and Zichen Zhang and Zipeng Fu and Zipeng Lin},
booktitle={International Conference on Robotics and Automation (ICRA)},
year = {2024},
}

@InProceedings{haarnoja2017soft,
  title = 	 {Soft Actor-Critic: Off-Policy Maximum Entropy Deep Reinforcement Learning with a Stochastic Actor},
  author =       {Haarnoja, Tuomas and Zhou, Aurick and Abbeel, Pieter and Levine, Sergey},
  booktitle = 	 {International Conference on Machine Learning (ICML)},
  year = 	 {2018},

}

@inproceedings{
zhou2024autonomous,
title={Autonomous Improvement of Instruction Following Skills via Foundation Models},
author={Zhiyuan Zhou and Pranav Atreya and Abraham Lee and Homer Rich Walke and Oier Mees and Sergey Levine},
booktitle={Conference on Robot Learning (CoRL)},
year={2024},
}

@article{oquab2024dinov2learningrobustvisual,
      title={DINOv2: Learning Robust Visual Features without Supervision}, 
      author={Maxime Oquab and Timothée Darcet and Théo Moutakanni and Huy Vo and Marc Szafraniec and Vasil Khalidov and Pierre Fernandez and Daniel Haziza and Francisco Massa and Alaaeldin El-Nouby and Mahmoud Assran and Nicolas Ballas and Wojciech Galuba and Russell Howes and Po-Yao Huang and Shang-Wen Li and Ishan Misra and Michael Rabbat and Vasu Sharma and Gabriel Synnaeve and Hu Xu and Hervé Jegou and Julien Mairal and Patrick Labatut and Armand Joulin and Piotr Bojanowski},
      year={2024},
      journal={arXiv preprint arXiv:2304.07193},
}

@inproceedings{reimers-2019-sentence-bert,
  title = "Sentence-BERT: Sentence Embeddings using Siamese BERT-Networks",
  author = "Reimers, Nils and Gurevych, Iryna",
  booktitle = "Empirical Methods in Natural Language Processing (EMNLP)",
  year = "2019",
}

@ARTICLE{Damen2022EpicKitchens,
           title={Rescaling Egocentric Vision: Collection, Pipeline and Challenges for EPIC-KITCHENS-100},
           author={Damen, Dima and Doughty, Hazel and Farinella, Giovanni Maria and Furnari, Antonino 
           and Ma, Jian and Kazakos, Evangelos and Moltisanti, Davide and Munro, Jonathan 
           and Perrett, Toby and Price, Will and Wray, Michael},
           journal   = {International Journal of Computer Vision (IJCV)},
           year      = {2022},
           volume = {130},
           pages = {33–55},
}

@inproceedings{
jang2021bcz,
title={{BC}-Z: Zero-Shot Task Generalization with Robotic Imitation Learning},
author={Eric Jang and Alex Irpan and Mohi Khansari and Daniel Kappler and Frederik Ebert and Corey Lynch and Sergey Levine and Chelsea Finn},
booktitle={Conference on Robot Learning (CoRL)},
year={2021},
}

@misc{dass2023jacoplay,
  author = {Dass, Shivin and Yapeter, Jullian and Zhang, Jesse and Zhang, Jiahui
            and Pertsch, Karl and Nikolaidis, Stefanos and Lim, Joseph J.},
  title = {CLVR Jaco Play Dataset},
  version = {1.0.0},
  year = {2023}
}

@inproceedings{
walke2023bridgedata,
title={BridgeData V2: A Dataset for Robot Learning at Scale},
author={Homer Rich Walke and Kevin Black and Tony Z. Zhao and Quan Vuong and Chongyi Zheng and Philippe Hansen-Estruch and Andre Wang He and Vivek Myers and Moo Jin Kim and Max Du and Abraham Lee and Kuan Fang and Chelsea Finn and Sergey Levine},
booktitle={Conference on Robot Learning (CoRL)},
year={2023},
}

@misc{fanuc_manipulation2023,
  title={Fanuc Manipulation: A Dataset for Learning-based Manipulation with FANUC Mate 200iD Robot},
  author={Zhu, Xinghao and Tian, Ran and Xu, Chenfeng and Ding, Mingyu and Zhan, Wei and Tomizuka, Masayoshi},
  year={2023}
}

@inproceedings{belkhale2023hydra,
 title={HYDRA: Hybrid Robot Actions for Imitation Learning},
 author={Belkhale, Suneel and Cui, Yuchen and Sadigh, Dorsa},
booktitle={Conference on Robot Learning (CoRL)},
 year={2023}
}

@misc{ucsd_kitchens,
  author = {Ge Yan and Kris Wu and Xiaolong Wang},
  title = {UCSD kitchens Dataset},
  year = {2023},
  month = {August}
}

@article{austinbuds,
  title={Bottom-Up Skill Discovery From Unsegmented Demonstrations for Long-Horizon Robot Manipulation},
  author={Zhu, Yifeng and Stone, Peter and Zhu, Yuke},
  journal={IEEE Robotics and Automation Letters (RA-L)},
  year={2022},
}

@inproceedings{
liu2024tail,
title={{TAIL}: Task-specific Adapters for Imitation Learning with Large Pretrained Models},
author={Zuxin Liu and Jesse Zhang and Kavosh Asadi and Yao Liu and Ding Zhao and Shoham Sabach and Rasool Fakoor},
booktitle={International Conference on Learning Representations (ICLR)},
year={2024},
}

@inproceedings{
kim2024openvla,
title={Open{VLA}: An Open-Source Vision-Language-Action Model},
author={Moo Jin Kim and Karl Pertsch and Siddharth Karamcheti and Ted Xiao and Ashwin Balakrishna and Suraj Nair and Rafael Rafailov and Ethan P Foster and Pannag R Sanketi and Quan Vuong and Thomas Kollar and Benjamin Burchfiel and Russ Tedrake and Dorsa Sadigh and Sergey Levine and Percy Liang and Chelsea Finn},
booktitle={Conference on Robot Learning (CoRL)},
year={2024},
}

@inproceedings{fan2022minedojo,
  title = {MineDojo: Building Open-Ended Embodied Agents with Internet-Scale Knowledge},
  author = {Linxi Fan and Guanzhi Wang and Yunfan Jiang and Ajay Mandlekar and Yuncong Yang and Haoyi Zhu and Andrew Tang and De-An Huang and Yuke Zhu and Anima Anandkumar},
  booktitle = {Conference on Neural Information Processing Systems Datasets and Benchmarks Track},
  year = {2022},
}

@inproceedings{DECKARD2023,
        title = {Do Embodied Agents Dream of Pixelated Sheep?:
                Embodied Decision Making using Language Guided
                World Modelling},
        author = {Kolby Nottingham and Prithviraj Ammanabrolu and
                Alane Suhr and Yejin Choi and Hannaneh Hajishirzi and
                Sameer Singh and Roy Fox},
  booktitle = {International Conference on Machine Learning (ICML)},
        year = {2023},
}

@article{nam2023liftunsupervisedreinforcementlearning,
      title={LiFT: Unsupervised Reinforcement Learning with Foundation Models as Teachers}, 
      author={Taewook Nam and Juyong Lee and Jesse Zhang and Sung Ju Hwang and Joseph J. Lim and Karl Pertsch},
      year={2023},
      journal={arXiv preprint arXiv:2312.08958},
}

@inproceedings{
rocamonde2024visionlanguage,
title={Vision-Language Models are Zero-Shot Reward Models for Reinforcement Learning},
author={Juan Rocamonde and Victoriano Montesinos and Elvis Nava and Ethan Perez and David Lindner},
booktitle={International Conference on Learning Representations (ICLR)},
year={2024},
}

@inproceedings{cui2022can,
  title={Can foundation models perform zero-shot task specification for robot manipulation?},
  author={Cui, Yuchen and Niekum, Scott and Gupta, Abhinav and Kumar, Vikash and Rajeswaran, Aravind},
  booktitle={Learning for Dynamics and Control Conference (L4DC)},
  year={2022},
}

@INPROCEEDINGS{venkataraman2025rlmvlmoffline,
  author={Venkataraman, Sreyas and Wang, Yufei and Wang, Ziyu and Ravie, Navin Sriram and Erickson, Zackory and Held, David},
  booktitle={IEEE/RSJ International Conference on Intelligent Robots and Systems (IROS)}, 
  title={Real-World Offline Reinforcement Learning from Vision Language Model Feedback}, 
  year={2025},
}

@inproceedings{kim2025reds,
    title={Subtask-Aware Visual Reward Learning from Segmented Demonstrations},
    author={Changyeon Kim and Minho Heo and Doohyun Lee and Honglak Lee and Jinwoo Shin and Joseph J. Lim and Kimin Lee},
    year={2025},
    booktitle={International Conference on Learning Representations (ICLR)},
}

@inproceedings{
hung2025victor,
title={{VIC}toR: Learning Hierarchical Vision-Instruction Correlation Rewards for Long-horizon Manipulation},
author={Kuo-Han Hung and Pang-Chi Lo and Jia-Fong Yeh and Han-Yuan Hsu and Yi-Ting Chen and Winston H. Hsu},
booktitle={International Conference on Learning Representations (ICLR)},
year={2025},
}

@inproceedings{wagenmaker2025steering,
  author    = {Wagenmaker, Andrew and Nakamoto, Mitsuhiko and Zhang, Yunchu and Park, Seohong and Yagoub, Waleed and Nagabandi, Anusha and Gupta, Abhishek and Levine, Sergey},
  title     = {Steering Your Diffusion Policy with Latent Space Reinforcement Learning},
  booktitle   = {Conference on Robot Learning (CoRL)},
  year      = {2025},
}

@article{khazatsky2024droid,
    title   = {DROID: A Large-Scale In-The-Wild Robot Manipulation Dataset},
    author  = {Alexander Khazatsky and Karl Pertsch and Suraj Nair and Ashwin Balakrishna and Sudeep Dasari and Siddharth Karamcheti and Soroush Nasiriany and Mohan Kumar Srirama and Lawrence Yunliang Chen and Kirsty Ellis and Peter David Fagan and Joey Hejna and Masha Itkina and Marion Lepert and Yecheng Jason Ma and Patrick Tree Miller and Jimmy Wu and Suneel Belkhale and Shivin Dass and Huy Ha and Arhan Jain and Abraham Lee and Youngwoon Lee and Marius Memmel and Sungjae Park and Ilija Radosavovic and Kaiyuan Wang and Albert Zhan and Kevin Black and Cheng Chi and Kyle Beltran Hatch and Shan Lin and Jingpei Lu and Jean Mercat and Abdul Rehman and Pannag R Sanketi and Archit Sharma and Cody Simpson and Quan Vuong and Homer Rich Walke and Blake Wulfe and Ted Xiao and Jonathan Heewon Yang and Arefeh Yavary and Tony Z. Zhao and Christopher Agia and Rohan Baijal and Mateo Guaman Castro and Daphne Chen and Qiuyu Chen and Trinity Chung and Jaimyn Drake and Ethan Paul Foster and Jensen Gao and Vitor Guizilini and David Antonio Herrera and Minho Heo and Kyle Hsu and Jiaheng Hu and Muhammad Zubair Irshad and Donovon Jackson and Charlotte Le and Yunshuang Li and Kevin Lin and Roy Lin and Zehan Ma and Abhiram Maddukuri and Suvir Mirchandani and Daniel Morton and Tony Nguyen and Abigail O'Neill and Rosario Scalise and Derick Seale and Victor Son and Stephen Tian and Emi Tran and Andrew E. Wang and Yilin Wu and Annie Xie and Jingyun Yang and Patrick Yin and Yunchu Zhang and Osbert Bastani and Glen Berseth and Jeannette Bohg and Ken Goldberg and Abhinav Gupta and Abhishek Gupta and Dinesh Jayaraman and Joseph J Lim and Jitendra Malik and Roberto Martín-Martín and Subramanian Ramamoorthy and Dorsa Sadigh and Shuran Song and Jiajun Wu and Michael C. Yip and Yuke Zhu and Thomas Kollar and Sergey Levine and Chelsea Finn},
    year    = {2024},
    journal={arXiv preprint arXiv:2403.12945},
}

@inproceedings{
  zhang2025rewind,
  title={Re{W}i{ND}: Language-Guided Rewards Teach Robot Policies without New Demonstrations},
  author={Jiahui Zhang and Yusen Luo and Abrar Anwar and Sumedh Anand Sontakke and Joseph J Lim and Jesse Thomason and Erdem Biyik and Jesse Zhang},
  booktitle={Conference on Robot Learning (CoRL)},
  year={2025},
}

@inproceedings{furl, 
title={{FuRL}: Visual-Language Models as Fuzzy Rewards for Reinforcement Learning}, 
author={Yuwei Fu and Haichao Zhang and  Di Wu and  Wei Xu and Benoit Boulet}, 
booktitle={Internatonal Conference on Machine Learning (ICML)}, 
year={2024}
}

@article{adeniji2023languagerewardmodulationpretraining,
  title = {Language Reward Modulation for Pretraining Reinforcement Learning},
  author = {Adeniji, Ademi and Xie, Amber and Sferrazza, Carmelo and Seo, Younggyo and James, Stephen and Abbeel, Pieter},
  year = {2024},
  journal={arXiv preprint arXiv:2308.12270},
}

@article{liu2023libero,
  title={LIBERO: Benchmarking Knowledge Transfer for Lifelong Robot Learning},
  author={Liu, Bo and Zhu, Yifeng and Gao, Chongkai and Feng, Yihao and Liu, Qiang and Zhu, Yuke and Stone, Peter},
  journal={arXiv preprint arXiv:2306.03310},
  year={2023}
}

@article{bai2025qwen2,
  title={Qwen2. 5-vl technical report},
  author={Bai, Shuai and Chen, Keqin and Liu, Xuejing and Wang, Jialin and Ge, Wenbin and Song, Sibo and Dang, Kai and Wang, Peng and Wang, Shijie and Tang, Jun and others},
  journal={arXiv preprint arXiv:2502.13923},
  year={2025}
}

@inproceedings{
    xie2024textreward,
    title={Text2Reward: Reward Shaping with Language Models for Reinforcement Learning},
    author={Tianbao Xie and Siheng Zhao and Chen Henry Wu and Yitao Liu and Qian Luo and Victor Zhong and Yanchao Yang and Tao Yu},
    booktitle={International Conference on Learning Representations (ICLR)},
    year={2024},
}

@inproceedings{du2023vision,
  title = 	 {Vision-Language Models as Success Detectors},
  author =       {Du, Yuqing and Konyushkova, Ksenia and Denil, Misha and Raju, Akhil and Landon, Jessica and Hill, Felix and de Freitas, Nando and Cabi, Serkan},
  booktitle = 	 {Conference on Lifelong Learning Agents},
  year = 	 {2023},
}

@inproceedings{guan2024tasksuccessenoughinvestigating,
      title={Task Success is not Enough: Investigating the Use of Video-Language Models as Behavior Critics for Catching Undesirable Agent Behaviors}, 
      author={Lin Guan and Yifan Zhou and Denis Liu and Yantian Zha and Heni Ben Amor and Subbarao Kambhampati},
      year={2024},
      booktitle={Conference on Language Modeling (COLM)},
}

@article{zhai2025VLAC,
  title={A Vision-Language-Action-Critic Model for Robotic Real-World Reinforcement Learning},
  author={Zhai, Shaopeng and Zhang, Qi and Zhang, Tianyi and Huang, Fuxian and Zhang, Haoran and Zhou, Ming and Zhang, Shengzhe and Liu, Litao and Lin, Sixu and Pang, Jiangmiao},
  journal={arXiv preprint arXiv:2509.15937},
  year={2025}
}

@article{tan2025robodopamine,
    title={Robo-Dopamine: General Process Reward Modeling for High-Precision Robotic Manipulation}, 
    author={Tan, Huajie and Chen, Sixiang and Xu, Yijie and Wang, Zixiao and Ji, Yuheng and Chi, Cheng and Lyu, Yaoxu and Zhao, Zhongxia and Chen, Xiansheng and Co, Peterson and Xie, Shaoxuan and Yao, Guocai and Wang, Pengwei and Wang, Zhongyuan and Zhang, Shanghang},
    journal={arXiv preprint arXiv:2512.23703},
    year={2025}
}

@article{Qwen3-VL,
      title={Qwen3-VL Technical Report}, 
      author={Shuai Bai and Yuxuan Cai and Ruizhe Chen and Keqin Chen and Xionghui Chen and Zesen Cheng and Lianghao Deng and Wei Ding and Chang Gao and Chunjiang Ge and Wenbin Ge and Zhifang Guo and Qidong Huang and Jie Huang and Fei Huang and Binyuan Hui and Shutong Jiang and Zhaohai Li and Mingsheng Li and Mei Li and Kaixin Li and Zicheng Lin and Junyang Lin and Xuejing Liu and Jiawei Liu and Chenglong Liu and Yang Liu and Dayiheng Liu and Shixuan Liu and Dunjie Lu and Ruilin Luo and Chenxu Lv and Rui Men and Lingchen Meng and Xuancheng Ren and Xingzhang Ren and Sibo Song and Yuchong Sun and Jun Tang and Jianhong Tu and Jianqiang Wan and Peng Wang and Pengfei Wang and Qiuyue Wang and Yuxuan Wang and Tianbao Xie and Yiheng Xu and Haiyang Xu and Jin Xu and Zhibo Yang and Mingkun Yang and Jianxin Yang and An Yang and Bowen Yu and Fei Zhang and Hang Zhang and Xi Zhang and Bo Zheng and Humen Zhong and Jingren Zhou and Fan Zhou and Jing Zhou and Yuanzhi Zhu and Ke Zhu},
	  journal={arXiv preprint arXiv:2511.21631},
      year={2025}
}

@article{chen2025sarm,
  title         = {SARM: Stage-Aware Reward Modeling for Long Horizon Robot Manipulation},
  author        = {Chen, Qianzhong and Yu, Justin and Schwager, Mac and Abbeel, Pieter and Shentu, Fred and Wu, Philipp},
  year          = {2025},
  journal       = {arXiv preprint arXiv:2509.25358},
}

@inproceedings{
ghasemipour2025selfimproving,
title={Self-Improving Embodied Foundation Models},
author={Seyed Kamyar Seyed Ghasemipour and Ayzaan Wahid and Jonathan Tompson and Pannag R Sanketi and Igor Mordatch},
booktitle={Advances in Neural Information Processing Systems (NeurIPS)},
year={2025},
}

@article{intelligence2025pi06vla,
  title   = {$\pi^{*}_{0.6}$: A VLA That Learns From Experience},
  author  = {Physical Intelligence and Ali Amin and Raichelle Aniceto and Ashwin Balakrishna and Kevin Black and Ken Conley and Grace Connors and James Darpinian and Karan Dhabalia and Jared DiCarlo and Danny Driess and Michael Equi and Adnan Esmail and Yunhao Fang and Chelsea Finn and Catherine Glossop and Thomas Godden and Ivan Goryachev and Lachy Groom and Hunter Hancock and Karol Hausman and Gashon Hussein and Brian Ichter and Szymon Jakubczak and Rowan Jen and Tim Jones and Ben Katz and Liyiming Ke and Chandra Kuchi and Marinda Lamb and Devin LeBlanc and Sergey Levine and Adrian Li-Bell and Yao Lu and Vishnu Mano and Mohith Mothukuri and Suraj Nair and Karl Pertsch and Allen Z. Ren and Charvi Sharma and Lucy Xiaoyang Shi and Laura Smith and Jost Tobias Springenberg and Kyle Stachowicz and Will Stoeckle and Alex Swerdlow and James Tanner and Marcel Torne and Quan Vuong and Anna Walling and Haohuan Wang and Blake Williams and Sukwon Yoo and Lili Yu and Ury Zhilinsky and Zhiyuan Zhou},
  journal = {arXiv:2511.14759},
  year    = {2025}
}

@article{
lee2026roboreward,
title={RoboReward: General-Purpose Vision-Language Reward Models for Robotics},
author={Tony Lee and Andrew Wagenmaker and Karl Pertsch and Percy Liang and Sergey Levine and Chelsea Finn},
journal = {arXiv preprint arXiv:2601.00675},
year    = {2026},
}

@article{lynch2023interactive,
  title={Interactive language: Talking to robots in real time},
  author={Lynch, Corey and Wahid, Ayzaan and Tompson, Jonathan and Ding, Tianli and Betker, James and Baruch, Robert and Armstrong, Travis and Florence, Pete},
  journal={IEEE Robotics and Automation Letters (RA-L)},
  year={2023},
}

@article{bharadhwaj2023roboagent,
    title={RoboAgent: Generalization and Efficiency in Robot Manipulation via Semantic Augmentations and Action Chunking},
    author={Homanga Bharadhwaj and Jay Vakil and Mohit Sharma and Abhinav Gupta and Shubham Tulsiani and Vikash Kumar},
    year={2023},
    journal = {arXiv preprint arXiv:2309.01918},
}

@inproceedings{heo2023furniturebench,
  title={FurnitureBench: Reproducible Real-World Benchmark for Long-Horizon Complex Manipulation},
  author={Minho Heo and Youngwoon Lee and Doohyun Lee and Joseph J. Lim},
  booktitle={Robotics: Science and Systems (RSS)},
  year={2023}
}

@inproceedings{shah2023mutex,
    title={{MUTEX}: Learning Unified Policies from Multimodal Task Specifications},
    author={Rutav Shah and Roberto Mart{\'\i}n-Mart{\'\i}n and Yuke Zhu},
    booktitle={Conference on Robot Learning (CoRL)},
    year={2023},
}

@article{luo2023multistage,
    author    = {Jianlan Luo and Charles Xu and Xinyang Geng and Gilbert Feng and Kuan Fang and Liam Tan and Stefan Schaal and Sergey Levine},
    title     = {Multi-Stage Cable Routing through Hierarchical Imitation Learning},
    journal = {arXiv preprint arXiv:2307.08927},
    year      = {2023},
}

@article{Radosavovic2023,
    title={Robot Learning with Sensorimotor Pre-training},
    author={Ilija Radosavovic and Baifeng Shi and Letian Fu and Ken Goldberg and Trevor Darrell and Jitendra Malik},
    year={2023},
    journal = {arXiv preprint arXiv:2306.10007},
}

@inproceedings{zhou2023train,
  author={Zhou, Gaoyue and Dean, Victoria and Srirama, Mohan Kumar and Rajeswaran, Aravind and Pari, Jyothish and Hatch, Kyle and Jain, Aryan and Yu, Tianhe and Abbeel, Pieter and Pinto, Lerrel and Finn, Chelsea and Gupta, Abhinav},
  booktitle={International Conference on Robotics and Automation (ICRA)}, 
  title={Train Offline, Test Online: A Real Robot Learning Benchmark}, 
  year={2023},
 }

@inproceedings{saxena2023multiresolution,
title={Multi-Resolution Sensing for Real-Time Control with Vision-Language Models},
author={Saumya Saxena and Mohit Sharma and Oliver Kroemer},
booktitle={Conference on Robot Learning (CoRL)},
year={2023},
}

@InProceedings{Radosavovic2022,
  title = {Real-World Robot Learning with Masked Visual Pre-training},
  author = {Ilija Radosavovic and Tete Xiao and Stephen James and Pieter Abbeel and Jitendra Malik and Trevor Darrell},
  booktitle = {Conference on Robot Learning (CoRL)},
  year = {2022}
}

@inproceedings{fu2024mobile,
  author    = {Fu, Zipeng and Zhao, Tony Z. and Finn, Chelsea},
  title     = {Mobile ALOHA: Learning Bimanual Mobile Manipulation with Low-Cost Whole-Body Teleoperation},
  booktitle = {{Conference on Robot Learning (CoRL)}},
  year      = {2024},
}

@inproceedings{vogel_edan_2020,
        title = {EDAN - an EMG-Controlled Daily Assistant to Help People with Physical Disabilities},
        booktitle = {IEEE/RSJ International Conference on Intelligent Robots and Systems (IROS)},
        author = {Vogel, Jörn and Hagengruber, Annette and Iskandar, Maged and Quere, Gabriel and Leipscher, Ulrike and Bustamante, Samuel and Dietrich, Alexander and Hoeppner, Hannes and Leidner, Daniel and Albu-Schäffer, Alin},
        year = {2020}
}

@inproceedings{quere_shared_2020,
        title = {Shared {Control} {Templates} for {Assistive} {Robotics}},
        booktitle = {International Conference on Robotics and Automation (ICRA)},
        author = {Quere, Gabriel and Hagengruber, Annette and Iskandar, Maged and Bustamante, Samuel and Leidner, Daniel and Stulp, Freek and Vogel, Joern},
        year = {2020},
}

@Article{OsaLSMO2022,
  author  = {Takayuki Osa},
  journal = {The International Journal of Robotics Research},
  title   = {Motion Planning by Learning the Solution Manifold in Trajectory Optimization},
  year    = {2022},
}

@inproceedings{haldar2023watch,
  title={Watch and match: Supercharging imitation with regularized optimal transport},
  author={Haldar, Siddhant and Mathur, Vaibhav and Yarats, Denis and Pinto, Lerrel},
  booktitle={Conference on Robot Learning (CoRL)},
  year={2023},
}

@article{xie2025human2robot,
  title   = {Human2Robot: Learning Robot Actions from Paired Human-Robot Videos},
  author  = {Xie, Sicheng and Cao, Haidong and Weng, Zejia and Xing, Zhen and Chen, Haoran and Shen, Shiwei and Leng, Jiaqi and Wu, Zuxuan and Jiang, Yu-Gang},
  journal = {arXiv preprint arXiv:2502.16587},
  year    = {2025},
}

@article{budzianowski2025opengvl,
  title   = {OpenGVL: Benchmarking Visual Temporal Progress for Data Curation},
  author  = {Budzianowski, Paweł and Wiśnios, Emilia and Góral, Gracjan and Kulakov, Igor and Petrenko, Viktor and Walas, Krzysztof},
  journal = {arXiv preprint arXiv:2509.17321},
  year    = {2025}
}

@inproceedings{
muslimani2025towards,
title={Towards Improving Reward Design in {RL}: A Reward Alignment Metric for {RL} Practitioners},
author={Calarina Muslimani and Kerrick Johnstonbaugh and Suyog Chandramouli and Serena Booth and W. Bradley Knox and Matthew E. Taylor},
booktitle={Reinforcement Learning Conference (RLC)},
year={2025},
}

@article{hwang2025maskedirl,
  title   = {Masked IRL: LLM-Guided Reward Disambiguation from Demonstrations and Language},
  author  = {Hwang, Minyoung and Forsey-Smerek, Alexandra and Dennler, Nathaniel and Bobu, Andreea},
  journal = {arXiv preprint arXiv:2511.14565},
  year    = {2025},
}

@inproceedings{bobu2021feature,
  author={Bobu, Andreea and Wiggert, Marius and Tomlin, Claire and Dragan, Anca D.},
  booktitle={ACM/IEEE International Conference on Human-Robot Interaction (HRI)}, 
  title={Feature Expansive Reward Learning: Rethinking Human Input}, 
  year={2021},
}

@inproceedings{kwok2025robomonkey,
  title = 	 {RoboMonkey: Scaling Test-Time Sampling and Verification for Vision-Language-Action Models},
  author =       {Kwok, Jacky and Agia, Christopher and Sinha, Rohan and Foutter, Matt and Li, Shulu and Stoica, Ion and Mirhoseini, Azalia and Pavone, Marco},
  booktitle = 	 {Conference on Robot Learning (CoRL)},
  year = 	 {2025},
}

@inproceedings{fu2024robot,
title={Robot Policy Learning with Temporal Optimal Transport Reward},
author={Yuwei Fu and Haichao Zhang and Di Wu and Wei Xu and Benoit Boulet},
booktitle={Conference on Neural Information Processing Systems (NeurIPS)},
year={2024}
}

@article{huang2025adapowerspecializingworldfoundation,
      title={AdaPower: Specializing World Foundation Models for Predictive Manipulation}, 
      author={Yuhang Huang and Shilong Zou and Jiazhao Zhang and Xinwang Liu and Ruizhen Hu and Kai Xu},
      year={2025},
      journal={arXiv preprint arXiv:2512.035358}
}

@inproceedings{hu2022lora,
title={Lo{RA}: Low-Rank Adaptation of Large Language Models},
author={Edward J Hu and Yelong Shen and Phillip Wallis and Zeyuan Allen-Zhu and Yuanzhi Li and Shean Wang and Lu Wang and Weizhu Chen},
booktitle={International Conference on Learning Representations (ICLR)},
year={2022}
}

@inproceedings{gu2025safe,
title={{SAFE}: Multitask Failure Detection for Vision-Language-Action Models},
author={Qiao Gu and Yuanliang Ju and Shengxiang Sun and Igor Gilitschenski and Haruki Nishimura and Masha Itkina and Florian Shkurti},
booktitle={Conference on Neural Information Processing Systems (NeurIPS)},
year={2025}
}

@article{zhang2026progresslmprogressreasoningvisionlanguage,
      title={PROGRESSLM: Towards Progress Reasoning in Vision-Language Models}, 
      author={Jianshu Zhang and Chengxuan Qian and Haosen Sun and Haoran Lu and Dingcheng Wang and Letian Xue and Han Liu},
      year={2026},
      journal={arXiv preprint arXiv:2601.15224},
}

@misc{contributors2024agibotworldrepo,
  title={AgiBot World Colosseum},
  author={AgiBot World Colosseum contributors},
  year={2024}
}

@article{jiang2025galaxea,
  title={Galaxea open-world dataset and g0 dual-system vla model},
  author={Jiang, Tao and Yuan, Tianyuan and Liu, Yicheng and Lu, Chenhao and Cui, Jianning and Liu, Xiao and Cheng, Shuiqi and Gao, Jiyang and Xu, Huazhe and Zhao, Hang},
  journal={arXiv preprint arXiv:2509.00576},
  year={2025}
}

@article{lee2025molmoact,
  title={Molmoact: Action reasoning models that can reason in space},
  author={Lee, Jason and Duan, Jiafei and Fang, Haoquan and Deng, Yuquan and Liu, Shuo and Li, Boyang and Fang, Bohan and Zhang, Jieyu and Wang, Yi Ru and Lee, Sangho and others},
  journal={arXiv preprint arXiv:2508.07917},
  year={2025}
}

@article{zhao2025humanoid,
  title={Humanoid everyday: A comprehensive robotic dataset for open-world humanoid manipulation},
  author={Zhao, Zhenyu and Jing, Hongyi and Liu, Xiawei and Mao, Jiageng and Jha, Abha and Yang, Hanwen and Xue, Rong and Zakharor, Sergey and Guizilini, Vitor and Wang, Yue},
  journal={arXiv preprint arXiv:2510.08807},
  year={2025}
}

@article{hwang2025motif,
  title={Motif: Motion instruction fine-tuning},
  author={Hwang, Minyoung and Hejna, Joey and Sadigh, Dorsa and Bisk, Yonatan},
  journal={IEEE Robotics and Automation Letters (RA-L)},
  year={2025},
}

@article{fang2023rh20t,
  title={Rh20t: A comprehensive robotic dataset for learning diverse skills in one-shot},
  author={Fang, Hao-Shu and Fang, Hongjie and Tang, Zhenyu and Liu, Jirong and Wang, Chenxi and Wang, Junbo and Zhu, Haoyi and Lu, Cewu},
  journal={arXiv preprint arXiv:2307.00595},
  year={2023}
}

@article{qiu2025humanoid,
  title={Humanoid Policy\~{} Human Policy},
  author={Qiu, Ri-Zhao and Yang, Shiqi and Cheng, Xuxin and Chawla, Chaitanya and Li, Jialong and He, Tairan and Yan, Ge and Yoon, David J and Hoque, Ryan and Paulsen, Lars and others},
  journal={arXiv preprint arXiv:2503.13441},
  year={2025}
}

@inproceedings{atreya2025roboarena,
  title = {RoboArena: Distributed Real-World Evaluation of Generalist Robot Policies},
  author = {Atreya, Pranav and Pertsch, Karl and Lee, Tony and Kim, Moo Jin and Jain, Arhan and Kuramshin, Artur and Eppner, Clemens and Neary, Cyrus and Hu, Edward and Ramos, Fabio and others},
  booktitle = {Conference on Robot Learning (CoRL)},
  year = {2025}
}

@article{lin2025failsafe,
  title={Failsafe: Reasoning and recovery from failures in vision-language-action models},
  author={Lin, Zijun and Duan, Jiafei and Fang, Haoquan and Fox, Dieter and Krishna, Ranjay and Tan, Cheston and Wen, Bihan},
  journal={arXiv preprint arXiv:2510.01642},
  year={2025}
}

@inproceedings{dai2025racer,
  title={Racer: Rich language-guided failure recovery policies for imitation learning},
  author={Dai, Yinpei and Lee, Jayjun and Fazeli, Nima and Chai, Joyce},
  booktitle={International Conference on Robotics and Automation (ICRA)},
  year={2025},
}

@article{zhou2025autoeval,
  title={Autoeval: Autonomous evaluation of generalist robot manipulation policies in the real world},
  author={Zhou, Zhiyuan and Atreya, Pranav and Tan, You Liang and Pertsch, Karl and Levine, Sergey},
  journal={arXiv preprint arXiv:2503.24278},
  year={2025}
}

@inproceedings{inceoglu2020fino,
  title={FINO-Net: A Deep Multimodal Sensor Fusion Framework for Manipulation Failure Detection},
  author={Inceoglu, Arda and Aksoy, Eren Erdal and Ak, Abdullah Cihan and Sariel, Sanem},
  booktitle={IEEE/RSJ International Conference on Intelligent Robots and Systems (IROS)},
  year={2021},
}

@inproceedings{taomaniskill3,
  title={ManiSkill3: GPU Parallelized Robotics Simulation and Rendering for Generalizable Embodied AI},
  author={Stone Tao and Fanbo Xiang and Arth Shukla and Yuzhe Qin and Xander Hinrichsen and Xiaodi Yuan and Chen Bao and Xinsong Lin and Yulin Liu and Tse-kai Chan and Yuan Gao and Xuanlin Li and Tongzhou Mu and Nan Xiao and Arnav Gurha and Viswesh Nagaswamy Rajesh and Yong Woo Choi and Yen-Ru Chen and Zhiao Huang and Roberto Calandra and Rui Chen and Shan Luo and Hao Su},
  booktitle = {Robotics: Science and Systems (RSS)},
  year={2025},
}

@article{james2020rlbench,
  title={Rlbench: The robot learning benchmark \& learning environment},
  author={James, Stephen and Ma, Zicong and Arrojo, David Rovick and Davison, Andrew J},
  journal={IEEE Robotics and Automation Letters (RA-L)},
  year={2020},
}

@inproceedings{jain2025sailor,
  title={A Smooth Sea Never Made a Skilled {SAILOR}: Robust Imitation via Learning to Search},
  author={Arnav Kumar Jain and Vibhakar Mohta and Subin Kim and Atiksh Bhardwaj and Juntao Ren and Yunhai Feng and Sanjiban Choudhury and Gokul Swamy},
  booktitle={Conference on Neural Information Processing Systems (NeurIPS)},
  year={2025},
}

@inproceedings{leiImageGAIL2018,
  author={Tai, Lei and Zhang, Jingwei and Liu, Ming and Burgard, Wolfram},
  booktitle={International Conference on Robotics and Automation (ICRA)}, 
  title={Socially Compliant Navigation Through Raw Depth Inputs with Generative Adversarial Imitation Learning}, 
  year={2018},
}

@article{Laming1984TheRO,
  title={The relativity of ‘absolute’ judgements},
  author={D. R. J. Laming},
  journal={British Journal of Mathematical and Statistical Psychology},
  year={1984},
  volume={37},
  pages={152-183},
}

@article{Stewart2005AbsoluteIB,
  title={Absolute identification by relative judgment.},
  author={Neil Stewart and Gordon D. A. Brown and Nick Chater},
  journal={Psychological review},
  year={2005},
  volume={112 4},
  pages={881-911},
}

@article{sharif2016relativememory,
author = {Marissa A. Sharif and Daniel M. Oppenheimer},
title ={The Effect of Relative Encoding on Memory-Based Judgments},
journal = {Psychological Science},
volume = {27},
number = {8},
pages = {1136-1145},
year = {2016},
}

@inproceedings{bellemare2017distributional,
  title={A distributional perspective on reinforcement learning},
  author={Bellemare, Marc G and Dabney, Will and Munos, R{\'e}mi},
  booktitle={International Conference on Machine Learning (ICML)},
  year={2017}
}

@inproceedings{zhai2023siglip,
    title={Sigmoid Loss for Language Image Pre-Training}, 
    author={Xiaohua Zhai and Basil Mustafa and Alexander Kolesnikov and Lucas Beyer},
    year={2023},
    booktitle={International Conference on Computer Vision (ICCV)},
}

@article{intelligence2025pi05,
  title={$\pi_{0.5}$: a Vision-Language-Action Model with Open-World Generalization},
  author={Intelligence, Physical and Black, Kevin and Brown, Noah and Darpinian, James and Dhabalia, Karan and Driess, Danny and Esmail, Adnan and Equi, Michael and Finn, Chelsea and Fusai, Niccolo and others},
  journal={arXiv preprint arXiv:2504.16054},
  year={2025}
}

@article{kostrikov2021offline,
  title={Offline reinforcement learning with implicit q-learning},
  author={Kostrikov, Ilya and Nair, Ashvin and Levine, Sergey},
  journal={arXiv preprint arXiv:2110.06169},
  year={2021}
}

@article{lynch2019play,
  title   = {Learning Latent Plans from Play},
  author  = {Lynch, Corey and Khansari, Mohi and Xiao, Ted and Kumar, Vikash
             and Tompson, Jonathan and Levine, Sergey and Sermanet, Pierre},
  journal = {Conference on Robot Learning (CoRL)},
  year    = {2019},
}

@inproceedings{xu2025detect,
  title={Can We Detect Failures Without Failure Data? {U}ncertainty-Aware Runtime Failure Detection for Imitation Learning Policies},
  author={Xu, Chen and Nguyen, Tony Khuong and Dixon, Emma and Rodriguez, Christopher and Miller, Patrick and Lee, Robert and Shah, Paarth and Ambrus, Rares and Nishimura, Haruki and Itkina, Masha},
  booktitle={Robotics: Science and Systems (RSS)},
  year={2025}
}

@inproceedings{agia2025unpacking,
    title        = {Unpacking Failure Modes of Generative Policies: Runtime Monitoring of Consistency and Progress},
    author       = {Agia, Christopher and Sinha, Rohan and Yang, Jingyun and Cao, Ziang and Antonova, Rika and Pavone, Marco and Bohg, Jeannette},
    year         = {2025},
    booktitle    = {Conference on Robot Learning (CoRL)},
}

@article{egodex,
      title={EgoDex: Learning Dexterous Manipulation from Large-Scale Egocentric Video}, 
      author={Ryan Hoque and Peide Huang and David J. Yoon and Mouli Sivapurapu and Jian Zhang},
      year={2025},
      journal={arXiv preprint arXiv:2505.11709}
}

@article{Bradley1952RankAO,
  title={Rank Analysis of Incomplete Block Designs: I. The Method of Paired Comparisons},
  author={Ralph Allan Bradley and Milton E. Terry},
  journal={Biometrika},
  year={1952},
  volume={39},
  pages={324},
}

@inproceedings{chen2024internvl,
    title={Internvl: Scaling up vision foundation models and aligning for generic visual-linguistic tasks},
    author={Chen, Zhe and Wu, Jiannan and Wang, Wenhai and Su, Weijie and Chen, Guo and Xing, Sen and Zhong, Muyan and Zhang, Qinglong and Zhu, Xizhou and Lu, Lewei and others},
    booktitle={Conference on Computer Vision and Pattern Recognition (CVPR)},
    year={2024}
}

@inproceedings{hong2025hand,
    title={HAND Me the Data: Fast Robot Adaptation via Hand Path Retrieval}, 
    author={Matthew Hong and Anthony Liang and Kevin Kim and Harshitha Rajaprakash and Jesse Thomason and Erdem Bıyık and Jesse Zhang},
    booktitle={International Conference on Robotics and Automation (ICRA)},
    year={2026}
}

@inproceedings{memmel2025strap,
    title={STRAP: Robot Sub-Trajectory Retrieval for Augmented Policy Learning},
    author={Memmel, Marius and Berg, Jacob and Chen, Bingqing and Gupta, Abhishek and Francis, Jonathan},
    booktitle={International Conference on Learning Representations (ICLR)},
    year={2025}
}

@inproceedings{zhang2026scizor,
  title={SCIZOR: Self-Supervised Data Curation for Large-Scale Imitation Learning},
  author={Zhang, Yu and Xie, Yuqi and Liu, Huihan and Shah, Rutav and Wan, Michael and Fan, Linxi and Zhu, Yuke},
  booktitle={International Conference on Robotics and Automation (ICRA)},
  year={2026}
}

@inproceedings{lin2024flowretrieval,
  title={FlowRetrieval: Flow-Guided Data Retrieval for Few-Shot Imitation Learning},
  author={Lin, Li-Heng and Cui, Yuchen and Xie, Amber and Hua, Tianyu and Sadigh, Dorsa},
  booktitle={Conference on Robot Learning (CoRL)},
  year={2024}
}

@inproceedings{du2023behavior,
  title={Behavior retrieval: Few-shot imitation learning by querying unlabeled datasets},
  author={Du, Maximilian and Nair, Suraj and Sadigh, Dorsa and Finn, Chelsea},
  booktitle={Robotics: Science and Systems (RSS)},
  year={2023}
}

@inproceedings{hejna2025robotdatacurationmutual,
    title={Robot Data Curation with Mutual Information Estimators}, 
    author={Joey Hejna and Suvir Mirchandani and Ashwin Balakrishna and Annie Xie and Ayzaan Wahid and Jonathan Tompson and Pannag Sanketi and Dhruv Shah and Coline Devin and Dorsa Sadigh},
    year={2025},
    booktitle={Robotics: Science and Systems (RSS)},
}

@inproceedings{xie2025data,
  title={Data Retrieval with Importance Weights for Few-Shot Imitation Learning},
  author={Xie, Amber and Chand, Rahul and Sadigh, Dorsa and Hejna, Joey},
  booktitle={Conference on Robot Learning (CoRL)},
  year={2025},
}

@inproceedings{agia2025cupid,
    title   = {CUPID: Curating Data your Robot Loves with Influence Functions},
    author  = {Agia, Christopher and Sinha, Rohan and Yang, Jingyun and Antonova, Rika and Pavone, Marco and Nishimura, Haruki and Itkina, Masha and Bohg, Jeannette},
    year    = {2025},
    booktitle={Conference on Robot Learning (CoRL)},
}

@inproceedings{mitra2024one,
  title={Which one? leveraging context between objects and multiple views for language grounding},
  author={Mitra, Chancharik and Anwar, Abrar and Corona, Rodolfo and Klein, Dan and Darrell, Trevor and Thomason, Jesse},
  booktitle={Conference of the North American Chapter of the Association for Computational Linguistics (NAACL)},
  year={2024}
}

@inproceedings{liang2022transformer,
  title={Transformer adapters for robot learning},
  author={Liang, Anthony and Singh, Ishika and Pertsch, Karl and Thomason, Jesse},
  booktitle={CoRL 2022 Workshop on Pre-training Robot Learning},
  year={2022}
}

@article{monroe2017colors,
  title={Colors in context: A pragmatic neural model for grounded language understanding},
  author={Monroe, Will and Hawkins, Robert XD and Goodman, Noah and Potts, Christopher},
  journal={Transactions of the Association for Computational Linguistics (TACL)},
  year={2017}
}

@article{bao2022learning,
  title={Learning to mediate disparities towards pragmatic communication},
  author={Bao, Yuwei and Ghosh, Sayan and Chai, Joyce},
  journal={arXiv preprint arXiv:2203.13685},
  year={2022}
}

@article{lu2025robofac,
  title={RoboFAC: A Comprehensive Framework for Robotic Failure Analysis and Correction},
  author={Weifeng Lu and Minghao Ye and Zewei Ye and Ruihan Tao and Shuo Yang and Bo Zhao},
  journal={arXiv preprint arXiv:2505.12224},
  year={2025}
}

@inproceedings{tian2026position,
  title = {Position: Good Embodied Reward Models Need Bad Behavior Data},
  author = {Tian, Ran and Wu, Yilin and Bajcsy, Andrea},
  booktitle = {International Conference on Machine Learning (ICML)},
  year = {2026},
}

@article{chen2026topreward,
  title        = {TOPReward: Token Probabilities as Hidden Zero-Shot Rewards for Robotics},
  author       = {Chen, Shirui and Harrison, Cole and Lee, Ying-Chun and Yang, Angela Jin and Ren, Zhongzheng and Ratliff, Lillian J. and Duan, Jiafei and Fox, Dieter and Krishna, Ranjay},
  journal      = {arXiv preprint arXiv:2602.19313},
  year         = {2026}
}

@article{huang2023look,
  title={Look before you leap: An exploratory study of uncertainty measurement for large language models},
  author={Huang, Yuheng and Song, Jiayang and Wang, Zhijie and Zhao, Shengming and Chen, Huaming and Juefei-Xu, Felix and Ma, Lei},
  journal={arXiv preprint arXiv:2307.10236},
  year={2023}
}

@inproceedings{pertsch2025fast,
  title={Fast: Efficient action tokenization for vision-language-action models},
  author={Pertsch, Karl and Stachowicz, Kyle and Ichter, Brian and Driess, Danny and Nair, Suraj and Vuong, Quan and Mees, Oier and Finn, Chelsea and Levine, Sergey},
  booktitle={Robotics: Science and Systems (RSS)},
  year={2025}
}

@article{ye2026dreamzero,
      title={World Action Models are Zero-shot Policies},
      author={Seonghyeon Ye and Yunhao Ge and Kaiyuan Zheng and Shenyuan Gao and Sihyun Yu and George Kurian and Suneel Indupuru and You Liang Tan and Chuning Zhu and Jiannan Xiang and Ayaan Malik and Kyungmin Lee and William Liang and Nadun Ranawaka and Jiasheng Gu and Yinzhen Xu and Guanzhi Wang and Fengyuan Hu and Avnish Narayan and Johan Bjorck and Jing Wang and Gwanghyun Kim and Dantong Niu and Ruijie Zheng and Yuqi Xie and Jimmy Wu and Qi Wang and Ryan Julian and Danfei Xu and Yilun Du and Yevgen Chebotar and Scott Reed and Jan Kautz and Yuke Zhu and Linxi "Jim" Fan and Joel Jang},
      year={2026},
      journal={arXiv preprint arXiv:2602.15922},
}

@inproceedings{zhang2025peek,
      title={PEEK: Guiding and Minimal Image Representations for Zero-Shot Generalization of Robot Manipulation Policies}, 
      author={Jesse Zhang and Marius Memmel and Kevin Kim and Dieter Fox and Jesse Thomason and Fabio Ramos and Erdem Bıyık and Abhishek Gupta and Anqi Li},
      booktitle   = {International Conference on Robotics and Automation (ICRA)},
      year={2026},
}

@article{molmo2openweightsdata,
    title={Molmo2: Open Weights and Data for Vision-Language Models with Video Understanding and Grounding},
    author={Christopher Clark and Jieyu Zhang and Zixian Ma and Jae Sung Park and Mohammadreza Salehi and Rohun Tripathi and Sangho Lee and Zhongzheng Ren and Chris Dongjoo Kim and Yinuo Yang and Vincent Shao and Yue Yang and Weikai Huang and Ziqi Gao and Taira Anderson and Jianrui Zhang and Jitesh Jain and George Stoica and Winson Han and Ali Farhadi and Ranjay Krishna},
    year={2026},
    journal={arXiv preprint arXiv:2601.10611}
}

@inproceedings{farebrother2024stop,
    title={Stop Regressing: Training Value Functions via Classification for Scalable Deep {RL}},
    author={Jesse Farebrother and Jordi Orbay and Quan Vuong and Adrien Ali Taiga and Yevgen Chebotar and Ted Xiao and Alex Irpan and Sergey Levine and Pablo Samuel Castro and Aleksandra Faust and Aviral Kumar and Rishabh Agarwal},
    booktitle={International Conference on Machine Learning (ICML)},
    year={2024}
}
